\documentclass[final,1p,times]{elsarticle}

\usepackage{amssymb,amsmath,amsthm}
\usepackage{amsmath}
\usepackage{txfonts}
\usepackage{latexsym,graphics}
\usepackage{graphicx}
\usepackage{subfigure}
\usepackage{mathrsfs}
\usepackage[flushleft]{caption2}
\usepackage{multirow}
\date{ }

\allowdisplaybreaks

\journal{Elsevier}

\begin{document}

\begin{frontmatter}



\title{ Wavelet Frame Based Image Restoration Using Sparsity, Nonlocal and Support Prior of Frame Coefficients \tnoteref{label1}}
\tnotetext[label1]{This work was supported by the Natural Science Foundation of China, Grant Nos. 11201054, 91330201, by the National Basic Research Program
(973 Program), Grant No. 2015CB856000, and by the Fundamental Research Funds for the Central Universities ZYGX2013Z005.}

 \author[label2]{Liangtian He }
 \ead{liangtian.he@qq.com}

 \author[label2,label3,label4]{Yilun Wang \corref{cor1}}
 \ead{yilun.wang@gmail.com}

 \cortext[cor1]{Corresponding author}

 \address[label2]{School of Mathematical Sciences,\\
 University of Electronic Science and Technology of China, \\
 Chengdu, Sichuan, 611731, P.R. China}

\address[label3]{Center for Information in BioMedicine \\
University of Electronic Science and Technology of China, \\
 Chengdu 610054, China}

 \address[label4]{Center for Applied Mathematics, Cornell Unviersity \\
  Ithaca, NY, 14850 USA}

\begin{abstract}
The wavelet frame systems have been widely investigated and applied for image restoration and many other image processing problems over the past decades, attributing to their good capability of sparsely approximating piece-wise smooth functions such as images. Most wavelet frame based models exploit the $l_1$ norm of frame coefficients for a sparsity constraint in the past. The authors in \cite{ZhangY2013, Dong2013} proposed an $l_0$ minimization model, where the $l_0$ norm of wavelet frame coefficients is penalized instead, and have demonstrated that significant improvements can be achieved compared to the commonly used $l_1$ minimization model. Very recently, the authors in \cite{Chen2015} proposed $l_0$-$l_2$ minimization model, where the nonlocal prior of frame coefficients is incorporated. This model proved to outperform the single $l_0$ minimization based model in terms of better recovered image quality. In this paper, we propose a truncated $l_0$-$l_2$ minimization model which combines sparsity, nonlocal and support prior of the frame coefficients. The extensive experiments have shown that the recovery results from the proposed regularization method performs better than existing state-of-the-art wavelet frame based methods, in terms of  edge enhancement and texture preserving performance.
\end{abstract}


\end{frontmatter}


\section{Introduction}
Image restoration, including image denoising, deblurring, inpainting, etc., is one of the most important fields in imaging sciences. Its main purpose is to enhance the
quality of an observed image that is corrupted in various ways during the process of imaging, acquisition and communication, and enable us to obtain a better and visually improved image. Image restoration tasks are often formulated as solving a linear inverse problem:
\begin{equation}\label{image restoration}
f = Au + \epsilon
\end{equation}
where $f$ is the observed corrupted image, $\epsilon$ denotes the additive white Gaussian noise with variance $\sigma^2$, the matrix $A$ is a linear operator. Different image restoration problems corresponding to different types of $A$, e.g., an identity operator for image denoising, a projection operator for inpainting, or a convolution operator for deconvolution, etc. Most image recovery tasks are ill-posed inverse linear problems, which makes solving (\ref{image restoration}) becomes nontrivial. Therefore, proper regularization techniques should be exploited to regularize the recovery process. Among them, variational models and wavelet frame based methods
are widely adopted.

The trend of variational models for image processing started from the popular Rudin-Osher-Fatemi (ROF) model which penalizes the total variation (TV) norm of the image \cite{ROF 1992}. The ROF model is effective for recovering images that are piece-wise constant, such as binary images. However, it is well known that TV regularization often suffers from so-called stair-case effect. In order to overcome this drawback, many other types of variational models have been further proposed and we refer the readers to \cite{Chambolle 1997}, \cite{Wang 2008}, \cite{Bredies 2010} and the references therein for more details.

In recent years, the sparsity-based prior based on wavelet frame has been playing a very important role in the development of effective image recovery models. The key idea behind
the wavelet frame based image restoration models is that the interested image is compressible in this transform domain. In other words, most important information of the interested image
can be preserved by using only few frame coefficients. Therefore, the regularized process can be chosen by minimizing the functional that promotes the sparsity of the underlying
solution in the transform domain. One commonly utilized regularizer term for the wavelet frame based models is the $l_1$ norm of transform coefficients. The connection of wavelet frame based approaches with variational and PDE based approaches is also studied in \cite{Cai2012}. Such connections explain the reason why wavelet frame based approaches are often superior to some of the variational based models. It is because the multiresolution structure and redundancy property of wavelet frames allow to adaptively select proper differential operators according to the order of the singularity of the underlying solutions for different regions of a given image.

There are several different wavelet frame based models in the literature including the synthesis based approaches \cite{Daubechies2007, Fadili2005, Fadili2009, Figueiredo2003, Figueiredo2005}, the analysis based approach \cite{Cai2009a, Elad2005, Starck2005} and the balanced based approach \cite{Chan2003, Cai2008, Cai2010}. These approaches are generally different due to  the redundancy of wavelet frame systems. In other words, the mapping from image $u$ to its wavelet frame coefficients is not one-to-one, i.e., the representation of $u$ in the wavelet frame domain is not unique. However, what these models share in common is that they mostly penalize the $l_1$ norm of the wavelet frame coefficients one way or another for sparse representation. It is well known that the $l_1$ norm based approaches are capable of obtaining sparsest solution if the operator $A$ satisfies certain conditions according to compressed sensing theories developed by Candes and Donoho \cite{Cand¨¨s2008, Chartrand2009, Chartrand2008}. For image restoration tasks, unfortunately, the conditions are not necessarily satisfied. Therefore, the $l_1$ norm based models often achieves suboptimal performance. Recently, Zhang et.al \cite{ZhangY2013} proposed to penalized the $l_0$ ``norm" of wavelet frame coefficients instead, and they developed an algorithm called penalty decomposition (PD) to solve the analysis approach with $l_0$ minimization. However, due to the non-convexity of $l_0$ ``norm", the computational cost of PD method is a little bit high. Then, Dong et.al \cite{Dong2013} developed a more efficient algorithm for this $l_0$ minimization model, called mean doubly augmented lagrangian (MDAL) method. Numerical experiments in \cite{Dong2013} demonstrated that this analysis based approach based on $l_0$ minimization can obtain higher quality recovery than those counterparts based on $l_1$ minimization.

In recent years, a class of nonlocal image recovery methods have drawn much attention. The nonlocal approaches are built on the observation that image structures of small regions tend to repeat themselves in spatial domain, which is suitable for exploiting the redundancy information in natural images. They have shown to be very effective for texture recovery. There are two types of nonlocal schemes developed to exploit such a nonlocal prior. One is the so-called nonlocal means proposed by Buades et.al \cite{Buades2005} for image denoising and has been extended to solve other inverse problems in image processing tasks; see e.g., \cite{Kindermann2005, Zhangxiaoqun2010, Deledalle2009}. Another is the patch-based method, where the nonlocal idea is combined with patch-dictionary methods, see \cite{Dabov2007, Dongweisheng2011, Dongweisheng2013a, Mairal2009, Dongweisheng2013b} for more details. The performance of nonlocal based methods is impressive for image restoration. However, since a large number of image patches should be clustered and sparsely coded during the iteration, the computational burden is considerably high.

Very recently, Quan et.al \cite{Quan2014} and Cai et.al \cite{Cai2014} proposed a data-driven local or nonlocal wavelet frame for image restoration. However, the computation costs of these methods are even higher than some patch-based approaches, e.g., the MATLAB implementation of the algorithm in \cite{Quan2014} takes about 9 minute for a natural image with size $256\times 256$, on a normal laptop. Chen et.al \cite{Chen2015} proposed an $l_0$-$l_2$ minimization model to balance the computational time and recovery quality. In their proposed model, nonlocal prior of the frame coefficients is incorporated in the variational model, in  terms of the $l_2$ norm. It plays an important role of estimating the frame coefficients that contain the textures and finer details of images. Numerical experiments have demonstrated the effectiveness of the added $l_2$ regularization term.

Sharp edges are essentially helpful to make a recovered image visually clear. In our previous work \cite{He2014}, we proposed a wavelet frame based truncated $l_1$ minimization model for image inpainting problem, where the support prior of frame coefficients is respected. This model  can generate a recovery image with much better enhanced edges than the counterpart based on single $l_1$ minimization or $l_0$ minimization.
Aiming at having the best of both edge enhanced and texture preserved approaches, this paper proposed a truncated $l_0$-$l_2$ scheme which will allow to simultaneously exploit three image priors: (i) sparsity prior of local intensity variations; (ii) self-repetition prior of local image structures in spatial domain; (iii) the nonzero/support prior of frame coefficients in the domain of wavelet frame transform. The main research motivation is to better preserve sharp edges. The main differences between our proposed  method and most existing wavelet frame based image restoration approaches are briefly summarized as follows:

\begin{itemize}
\item In contrast to wavelet frame based $l_1$ or $l_0$ minimization model \cite{Cai2009a, Dong2013}, where only sparsity prior is utilized,
 the added $l_2$ regularizer term  exploits the self-recurrence prior of local image structures in spatial domain [5], i.e., the nonlocal prior is also incorporated. Therefore, the textures and tiny details can be preserved well in the restored image.

\item Different with the truncated $l_1$ minimization model proposed for image inpainting problem in our previous work \cite{He2014}. The new truncated $l_0$ minimization model is proposed and the corresponding efficient algorithm is developed here. In addition, the nonlocal prior of frame coefficients is added for the purpose of better  preserving important image details.

\item Different with the $l_0$-$l_2$ minimization model proposed in \cite{Chen2015}, where both sparsity prior and nonlocal prior are exploited, the support prior of frame coefficients is also respected in our developed truncated $l_0$-$l_2$ minimization model, leading to significantly better edge preserving performance.

\end{itemize}

The rest of this paper is organized as follows. In the next section, we first give a brief review of wavelet tight frames. Then we further revisit the current wavelet frame based image recovery models using either the single $l_1$ or $l_0$ regularization term \cite{Cai2009a, Dong2013}, and the combined $l_0$-$l_2$ minimization model \cite{Chen2015}. In section 3, we propose a new wavelet frame based image recovery model utilizing the truncated $l_0$-$l_2$ regularizer and develop a corresponding efficient algorithm in the spirit of iterative support detection. Section 4 is devoted to the experimental evaluation of the proposed image restoration method. We compare it with several state-of-art methods  such as the split bregman method for solving the wavelet frame based $l_1$ minimization model \cite{Cai2009a}, the MDAL method for solving the $l_0$ regularization model \cite{Dong2013}, the algorithm proposed in \cite{Chen2015} for $l_0$-$l_2$ regularizer, and the famous Iterative Decoupled Deblurring BM3D (IDD-BM3D) algorithm \cite{Dabov2008, Danielyan2012}. Section 5 is devoted to the conclusion of this paper and discussions on some possible future work.

\section{Preliminaries and Previous works}

\subsection{Tight wavelet frames and wavelet frame based image processing methods}
In this section, we briefly introduce some preliminaries of tight wavelet frames, and then revisit some of the current typical wavelet frame based image restoration models and the corresponding efficient algorithms. We refer the interested readers to \cite{Ron1997, Daubechies2003, Shen2010, Dong2010} for further detailed introduction of wavelet frame and its applications.

Tight wavelet frame are widely utilized in image processing. One wavelet frame for $L_2(\mathbb{R})$ is a system generated by the shifts and dilations of a finite set of generators $\Psi = \{ \Psi_1, \Psi_1, \ldots, \Psi_n \} \subset L_2(\mathbb{R})$:
\[
X(\Psi) = \{ \Psi_{l,j,k}, 1\leq l, j\in \mathbb{Z}, k\in \mathbb{Z} \}
\]
where $\Psi_{l,j,k} = 2^{j/2}\Psi_l(2^j \cdot -k)$. Such set $X(\Psi)$ is called tight frame of $L_2(\mathbb{R})$ if
\[
f = \sum_{\psi\in \Psi}<f,\psi>\psi,   \forall f \in L_2(\mathbb{R}).
\]
The construction of framelets can be obtained according to the unitary extension principle (UEP).  We refer the readers to \cite{Dong2010, Cai2009b} for more details. Following the common experiment implementations, the linear B-spline framelet is used by considering the balance of the quality and time. The linear B-spline framelet has two generators and the associated masks $\{ h_0, h_1, h_2 \}$ are
\[
h_0 = \frac{1}{4}[1,2,1]; h_1=\frac{\sqrt{2}}{4}[1,0,-1]; h_2=\frac{1}{4}[-1,2,-1].
\]
Given the 1D tight wavelet frame, the framelets for $L_2(\mathbb{R}^2)$ can be easily constructed by using tensors products of 1D framelets.

In the discrete setting, we will use $ W \in \mathbb{R}^{m \times n} $ with $ m \geq n $ to denote the transform matrix of framelet decomposition and use $ W^T$ to denote the fast reconstruction. Then according to the unitary extension principle we have $ W^TW = I $. The matrix $W$ is called the analysis (decomposition) operator, and its transpose $W^T$ is called the synthesis (reconstruction) operator. The $L$-level framelet decomposition of $u$ will be further denoted as:
\[
Wu = (\ldots,W_{l,j}u,\ldots) \qquad  \mathrm{for} \quad 0\leq 1 \leq L-1,j \in \mathcal{I}
\]
where $ \mathcal{I} $ denotes the index set of the framelet bands and $ W_{l,j}u \in \mathbb{R}^n $ is the wavelet frame coefficients of $u$ in bands $j$ at level $l$. The frame coefficients $W_{l,j}u$ can be constructed from the masks associated with the framelets. We consider the $L$-Level undecimal wavelet tight frame system without the down-sampling and up-sampling operators  here. Let $h_0$ denote the mask associated with the scaling function and $\{h_1, h_2, \ldots, h_n\}$ denote the masks associated with other framelets. Denote
\begin{equation} \label{eq: frame construction}
h_j^{(l)} = \underbrace{h_0*h_0*\cdots h_0}_{l-1}*h_j
\end{equation}
where $*$ denotes the discrete convolution operator. Then $W_{l,j}$ corresponds to the Toeplitz-plus-Hankel matrix that represents the convolution operator $h_j^{(l)}$ under Neumann boundary condition.

\subsection{The single $l_1$ and $l_0$ minimization model}
Due to the redundancy of the wavelet frame systems ($WW^T \neq I$), there are several different wavelet frame based models. These models  mostly penalize the $l_1$ norm of the wavelet frame coefficients one way or another for sparsity constraint. Detailed description of these different models can be found in \cite{Dong2010, Shen2011}. Numerical experiments in \cite{Dong2010} demonstrated that the quality of the restored images by these models is approximately comparable. In this paper, we only consider the following analysis based approach:
\begin{equation} \label{eq:single l1}
\min_u \frac{1}{2}||Au-f||_2^2 + ||\lambda \cdot Wu||_{1,p}
\end{equation}
where $p = 1$ or $p = 2$ corresponds to anisotropic $ {\ell}_1 $ norm and isotropic $ {\ell}_1 $ norm, respectively.
The generalized $ {\ell}_1 $-norm here is defined as
\begin{equation}
||\lambda \cdot Wu||_{1,p} = ||\sum_{l=0}^{L-1}\left(\sum_{j \in \mathcal{I}}\lambda_{l,j}|W_{l,j}u|^p\right)^{1/p}||_1
\end{equation}
where $| \cdot |^p $ and $(\cdot)^{\frac{1}{p}} $ are entrywise operations. If letting $ \alpha = Wu $ and substitute it into (\ref{eq:single l1}), we can get the rewritten form of (\ref{eq:single l1}) as follows
\begin{equation}
\min_{u,\alpha} \frac{1}{2}||Au-f||_2^2 + ||\lambda \cdot \alpha||_{1,p} \qquad  s.t.\quad \alpha = Wu.
\end{equation}

As mentioned above, most of the frame based models exploit the $l_1$-norm of frame coefficients as the sparsity regularizer, and  can be efficiently solved via lots of off-the-shelf methods, such as the famous split bregman method or alternating direction multipliers method \cite{Cai2009a, Goldtein2009, Esser2009, Boyd2010}. Recently,  $l_p$ quasi-norm ($0\leq p\leq 1$) regularization was further investigated to recover the image with more sharped edges. The authors in \cite{ZhangY2013} proposed to use the $l_0$ ``norm" instead of the $l_1$ norm in the analysis model:
\begin{equation} \label{eq:single l0}
\min_u \frac{1}{2}||Au-f||_2^2 + \lambda_\textbf{i} || (Wu)_\textbf{i} ||_0
\end{equation}
where the multi-index $\textbf{i}$ is used here and $(Wu)_\textbf{i}$ (similar for $\lambda_\textbf{i}$ ) denotes the value of $Wu$ at a given pixel location within a certain level and band of wavelet frame transform. $\lambda_\textbf{i}$ is the positive regularization parameter. The $l_0$ ``norm" is defined to be the number of the non-zero elements of $Wu$.
Note that its proximity operator can be easily computed by the hard-thresholding operator.

An algorithm called PD method was proposed to solve the above $l_0$ minimization problem in \cite{ZhangY2013}. Recently, a more efficient algorithm, called MDAL method is developed for solving the same problem in literature \cite{Dong2013}. It can be seen as an extension of the augmented Lagrangian (DAL) method \cite{Rockafellar1976, Zhangxiaoqun2011} to handle the non-convex regularizer such as $l_0$ minimization.
Letting $ \alpha = Wu $ and substituting it into (\ref{eq:single l0}), ones can obtain a constrained form:
\begin{equation}
\min_u \frac{1}{2}||Au-f||_2^2 + \lambda_\textbf{i} || (\alpha)_\textbf{i} ||_0, \quad s.t. \quad \alpha = Wu
\end{equation}
Then, the DAL method can be formulated as
\begin{equation}\label{eq:DAL}
\left\{\begin{array}{lll}
u^{k+1} = \mathrm{arg} \min_u \frac{1}{2}||Au - f||_2^2 + \frac{\mu}{2}||Wu - \alpha^{k} + b^{k}||_2^2 + \frac{\gamma}{2}||u - u^{k}||_2^2 \\
\alpha^{k+1} = \mathrm{arg} \min_{\alpha} ||\lambda \cdot \alpha|| + \frac{\mu}{2}||\alpha - (Wu^{k+1} + b^{k})||_2^2 + \frac{\gamma}{2}||\alpha - \alpha^{k}||_2^2 \\
b^{k+1} = b^{k} + Wu^{k+1} - \alpha^{k+1}
\end{array}\right.
\end{equation}
where $\mu> 0$ is a penalty parameter, and the parameter $\gamma>0$ controls the regularity of the iterative sequence.

Although it seems to be reasonable to apply the DAL algorithm (\ref{eq:DAL}) to solve the $l_0$ minimization problem, the numerical experiments in \cite{Dong2013} demonstrate that the iteration sequence generated via (\ref{eq:DAL}) may be unstable or  at least the convergence speed is quite slow. Therefore, the authors in \cite{Dong2013} utilized the arithmetic means of the solution sequence, denoted by
\begin{equation} \label{eq:mean step}
\bar{u}^{k} = \frac{1}{k+1}\sum_{j=0}^{k} u^k; \quad  \bar{\alpha}^{k} = \frac{1}{k+1}\sum_{j=0}^{k} \alpha^k.
\end{equation}
as the final output instead of the sequence $(u^k, \alpha^k)$ itself. The authors called this algorithm mean doubly augmented Lagrangian (MDAL) method. Numerical experiments show that the sequence $(\bar{u}^k, \bar{\alpha}^{k})$ generated by the MDAL method is really convergent, and both the convergence speed and the quality of recovery are superior to the PD method \cite{ZhangY2013}.

\subsection{The combined $l_0$-$l_2$ minimization model}
Note that the wavelet frame coefficients can be obtained via the discrete convolution operator such as the definition in (\ref{eq: frame construction}), which implies the fact that they represent the results of different types of finite-difference operators defined in the neighborhood. Therefore, inspired by the previous work \cite{Dongweisheng2011, Dongweisheng2013a}, the authors in \cite{Chen2015} proposed a nonlocal estimation method of the frame coefficients.

The frame coefficient of image $u$ in band $j$ at level $l$, is denoted as $\alpha_{l,j} = W_{l,j}u$. Assume that the coefficient $\alpha_{l,j}$ is generated by convoluting the image $u$ with the operator $h_j^{(l)}$  defined in (\ref{eq: frame construction}). It is well known that the convolution is performed in the local image regions. Therefore, for any pixel indexes $\textbf{p}$ and $\textbf{q}$, the neighborhoods around the pixels $\textbf{p}$ and $\textbf{q}$ of the image are similar, and we have that
\begin{equation}\label{eq:similar equation}
\alpha_{l,j}(\textbf{p}) \approx \alpha_{l,j}(\textbf{q})
\end{equation}
The relationship in (\ref{eq:similar equation}) means that the similarity of two frame coefficients can be measured by the similarity of the corresponding image patches. Hence one frame coefficient corresponding to a given patch can be approximately estimated by the frame coefficients obtained by the similar patches.

Denote the image patch centered at the pixel $\textbf{p}$ as $\textbf{u}_\textbf{p}$, which is defined as $\textbf{u}_\textbf{p} = \{ u_\textbf{q}, \textbf{q}\in \mathcal{N}(\textbf{p}) \}$. Here $\mathcal{N}(\textbf{p})$ denotes a neighborhood at the pixel $\textbf{p}$. Then from the collection of $m$ similar patches $\{ \textbf{u}_{\textbf{p}_1}, \textbf{u}_{\textbf{p}_2}, \ldots, \textbf{u}_{\textbf{p}_m} \}$, the nonlocal estimation of frame coefficients $\alpha_{l,j}(\textbf{p})$ can be obtained by
\begin{equation} \label{beta equation}
\beta_{l,j}(\textbf{p}) = \sum_{i=1}^{m}\omega_{i}\alpha_{l,j}(\textbf{p}_{i})
\end{equation}
where $w_i$ is the weight based on patch similarity. Similarly to the nonlocal means method \cite{Buades2005}, the weight is inversely proportional to the distance between the patches $\textbf{u}_{\textbf{p}}$ and $\textbf{u}_{\textbf{p}_i}$
\begin{equation} \label{filtering and normalizing factor}
w_i = \frac{1}{C}\mathrm{exp}(-||\textbf{u}_{\textbf{p}} -  \textbf{u}_{\textbf{p}_i}||_2^2/h )
\end{equation}
where $h$ is the filtering parameter and $C$ is the normalizing factor.

Therefore, combining the sparse prior and nonlocal estimation of the frame coefficients, the authors in \cite{Chen2015} proposed the wavelet frame based $l_0$-$l_2$ minimization image restoration model
\begin{equation} \label{eq:l0l2}
\min_{u,\beta} \frac{1}{2}||Au-f||_2^2 + \lambda_\textbf{i} || (Wu)_\textbf{i} ||_0 + \frac{\nu}{2}||Wu - \beta||_2^2
\end{equation}
where $\beta = [\beta_{l,j}]_{0\leq l\leq L-1, j\in \mathcal{I}}$ denotes the nonlocal estimation of $Wu$.
In this combined $l_0$-$l_2$ model, the second regularization term plays the role of recovering both the smoothness of homogeneous regions and the sharpness of edges, the third regularization term related to the nonlocal estimator is utilized to preserve the coefficients that contain textures and finer details.
The variables $u$ and $\beta$ are updated alternatively. When the value of $\beta$ is fixed, the minimization problem (\ref{eq:l0l2}) can be rewritten as a constrained optimization problem as follows:
\begin{equation} \label{eq:l0l2 constraint}
\min_{u,\alpha} \frac{1}{2}||Au-f||_2^2 + \lambda_\textbf{i} || (Wu)_\textbf{i} ||_0 + \frac{\nu}{2}||\alpha - \beta||_2^2,  \quad s.t.  \quad \alpha = Wu
\end{equation}
Similar to the MDAL method applied to the single $l_0$ optimization problem, the DAL method applied to (\ref{eq:l0l2 constraint}) is reformulated as follows:
\begin{equation}\label{eq: DALL0L2}
\left\{\begin{array}{lll}
u^{k+1} = \mathrm{arg} \min_u \frac{1}{2}||Au - f||_2^2 + \frac{\mu}{2}||Wu - \alpha^{k} + b^{k}||_2^2 + \frac{\gamma}{2}||u - u^{k}||_2^2 \\
\alpha^{k+1} = \mathrm{arg} \min_{\alpha} ||\lambda \cdot \alpha|| + \frac{\nu}{2}||\alpha - \beta||_2^2 + \frac{\mu}{2}||\alpha - (Wu^{k+1} + b^{k})||_2^2 + \frac{\gamma}{2}||\alpha - \alpha^{k}||_2^2 \\
b^{k+1} = b^{k} + Wu^{k+1} - \alpha^{k+1}
\end{array}\right.
\end{equation}

It is easy to show that each of the subproblems of (\ref{eq: DALL0L2}) has a closed form solution
\begin{equation}\label{eq:DALSolution}
\left\{\begin{array}{lll}
u^{k+1} = (A^T + (\mu + \gamma)I)^{-1}(A^Tf + \gamma u^k + \mu W^T(\alpha^k - b^k)) \\
\alpha^{k+1} =  \mathcal{H}_{\lambda, \nu, \mu, \gamma}(\beta, Wu^{k+1}+b^k, \alpha^k)               \\
b^{k+1} = b^{k} + Wu^{k+1} - \alpha^{k+1}
\end{array}\right.
\end{equation}
where the operator $\mathcal{H}$ is a generalized component-wisely hard-threhsolding operator defined as follows:
\begin{equation}
(\mathcal{H}_{\lambda, \nu, \mu, \gamma}(x,y,z))_i = \left\{ \begin{array}{ll}
0,\quad  \quad \quad  \mathrm{if} \quad | \frac{\nu x_i + \mu y_i + \gamma z_i}{\nu + \mu + \gamma} | < \frac{2\lambda_i}{\nu + \mu + \gamma}        \\
\frac{\nu x_i + \mu y_i + \gamma z_i}{\nu + \mu + \gamma}, \quad \mathrm{otherwise} \end{array} \right.
\end{equation}
When the $u$ is fixed, the update of $\beta$ relies on the frame coefficients $\alpha=Wu$. If we use the current estimate $\alpha^k$ to approximate $\alpha$,  the estimation of $\beta$ at the $(k+1)$-th iteration can be updated via the following formula:
\begin{equation} \label{eq: beta}
\beta_{l,j}(\textbf{p}) = \sum_{i=1}^{m}\omega_i\alpha_{l,j}^k(\textbf{p}_i)
\end{equation}
As for the initialization of nonlocal estimation $\beta$, the authors first use the standard Tikhonov regularization method to obtain an initial image $\hat{u}$, then the similarity weight can be obtained according to $\hat{u}$, and finally the initial estimation $\beta^{0}$ is computed by
\begin{equation} \label{eq: beta initial}
\beta_{l,j}^{0}(\textbf{p}) = \sum_{i=1}^{m}\omega_iW_{l,j}\hat{u}(\textbf{p}_i)
\end{equation}
The arithmetic means of the sequence $(u^k, \alpha^k)$ are treated as the actual output of the algorithm, which is the same as the strategy (9).  So we call it as non-local mean doubly augmented Lagrangian method (Non-local MDAL) in the section of numerical experiments.

\section{Our Proposed Model and Algorithm}
\subsection{Genesis of the idea}

The main idea of our developed model in this paper comes from our proposed iterative support detection (ISD) for sparse signal recovery in compressive sensing \cite{Wang2010}. To make the paper self-contained, we first briefly revisit the idea of ISD. Compressive sensing reconstructs a underlying sparse
signal from a small set of linear projections. Let $ \bar{x} $ denote a $ k $-sparse signal and $ b = A\bar{x} $
represent a set of $ m $ linear projections of $ \bar{x} $.  The optimization problem
\begin{equation} \label{eq: l0}
(l_0) \qquad \min_{x}||x||_0 \quad s.t. \quad Ax = b.
\end{equation}
where $||x||_0$ is defined as the number of nonzero components of $x$, can exactly reconstruct $\bar{x}$ from $O(k)$ random projections. However, because $||x||_0$ is nonconvex and combinatorial, it is really a challenging task to solve this $l_0$ minimization optimization problem. A general alternative is the Basis Pursuit (BP) problem
\begin{equation}\label{eq: l1}
(\mathrm{BP}/l_1) \qquad \min_{x}||x||_1 \quad s.t. \quad Ax = b.
\end{equation}
Unlike the BP problem which is a one-stage convex relaxation method, ISD is  a multi-stage convex relaxation method, alternatively calling its two components: support detection and signal reconstruction. ISD starts from solving a standard BP problem. If the BP model returns a correct sparse solution, things are fine and ISD stops there. Otherwise,  from the incorrect reconstruction,
support detection will be performed to identify an index set $I$, which contain some elements of supp$ (\bar{x}) = \{i : \bar{x}_i \neq 0\} $. After the acquiring of detected support information, the corresponding expected nonzero elements will be truncated from the $\ell_1$ regularization term and the resulted model is as follows
\begin{equation}\label{eq: Truncated l1}
(\mathrm{Truncated \quad BP}/l_1) \qquad \min_{x}||x_T||_1  \quad s.t. \quad Ax = b.
\end{equation}
where $ T = I^C $ and $ ||x_T||_1 = \sum_{i \not\in I}|x_i| $. Those entries of supp($\bar{x}$) in $I$ will help (\ref{eq: Truncated l1}) reconstruct a better solution compared to (\ref{eq: l1}). From this better solution, support detection will be able to identify more entries of supp($\bar{x}$) and then yield an even better $I$. In this way, the two components of ISD work together to gradually recover supp($\bar{x}$) and improve the reconstruction result.

Intuitively, the truncated $l_0$ model is similar in the spirit of above truncated $l_1$ minimization model, after the acquiring of reliable support detection. The corresponding elements should be not forced to move closer to 0 and therefore we move it out from the regularization term. In fact, once the nonzero locations are identified, their influence is downweighted in order to allow the true nonzero components should not be shrunk. Therefore, the corresponding elements can also be truncated from the original $l_0$ regularization term in the same way. The resulted truncated $l_0$ model is
\begin{equation} \label{eq: Truncated l0}
(\mathrm{Truncated} \quad l_0) \qquad \min_{x}||x_T||_0 \quad s.t. \quad Ax = b.
\end{equation}
where $x_T$ is the truncated subvector of $x$, and $||x_T||_0$ is defined to be the number of the non-zero elements of $x_T$.

In most cases, the true signal $\bar{x}$ itself is not sparse or compressible. However, its representation under a certain basis, frame, or dictionary can be sparse or compressible. In such a case, assuming that $ \bar{y }= W\bar{x} $ is sparse or compressible for a certain linear transform $W$, one should minimize $||Wx||_1$ and $||(Wx)_T||_1$ as a substitute respectively, instead of minimizing $||x||_1$ and $||x||_T$.

From the inexact intermediate reconstruction, ISD tries to obtain a reliable support detection, which can be able to take advantages of the features and prior
information about the true signal $\bar{x}$. The authors in \cite{Wang2010} focus on the sparse or compressible signals with elements having a
fast decaying distribution of nonzeros. For these kinds of signals, they proposed to perform the support detection by thresholding the solution of (\ref{eq: Truncated l1}), and called the corresponding support detection method \textit{threshold}-ISD. We emphasize that the fast decaying property is a mild assumption because in fact most natural images satisfy this property in an appropriate basis, for instance, wavelets, curvelets and wavelet frames et al. In addition, it should be pointed out that the fast decaying property is not even necessary if the structure information of the underlying solution is exploited, and we refer the interested readers to \cite{Fan2014} for more details.

\subsection{Wavelet frame based truncated $l_0$-$l_2$ image restoration model}

In the above section, we have briefly reviewed some typical wavelet frame based image restoration models and their corresponding efficient algorithms. The single $l_1$ or $l_0$ minimization model utilizes the sparsity prior of local intensity variations, i.e., sparsity prior of the frame coefficients. The combined $l_0$-$l_2$ minimization model further exploits self-repetition prior of local image structures in spatial domain, i.e., the nonlocal prior of the frame coefficients. In this paper, the support prior of the frame coefficients will be self-learning via ISD and  be respected with the aim to better preserve the sharp edges, and we propose the truncated $l_0$-$l_2$ minimization model and develop a corresponding efficient algorithm, which allow the regularization model simultaneously to exploit three important image priors: sparsity, nonlocal and support prior of the frame coefficients in the transform domain of wavelet frame.
\begin{equation} \label{eq:new truncatedl0l2}
\min_{u,\beta} \frac{1}{2}||Au-f||_2^2 + (\lambda_\textbf{i})_T || ((Wu)_\textbf{i})_T ||_0 + \frac{\nu}{2}||Wu - \beta||_2^2
\end{equation}
where $((Wu)_\textbf{i})_T$ is the truncated version of $(Wu)_\textbf{i}$, and $||((Wu)_\textbf{i})_T||_0$ denotes the number of the non-zero elements of $((Wu)_\textbf{i})_T$.

\subsubsection{Component 1: Solving the truncated $l_0$-$l_2$ minimization problem}
The equivalent constraint optimization problem of (\ref{eq:new truncatedl0l2}) is
\begin{equation} \label{eq: Truncated l0l2 constraint}
\min_{u,\beta} \frac{1}{2}||Au-f||_2^2 + (\lambda_\textbf{i})_T || (\alpha_\textbf{i})_T ||_0 + \frac{\nu}{2}||\alpha - \beta||_2^2, \quad s.t. \quad \alpha = Wu
\end{equation}
When the $\beta$ is fixed, the DAL method applied to (\ref{eq: Truncated l0l2 constraint}) is formulated as
\begin{equation}\label{eq: DAL truncated l0l2}
\left\{\begin{array}{lll}
u^{k+1} = \mathrm{arg} \min_u \frac{1}{2}||Au - f||_2^2 + \frac{\mu}{2}||Wu - \alpha^{k} + b^{k}||_2^2 + \frac{\gamma}{2}||u - u^{k}||_2^2 \\
\alpha^{k+1} = \mathrm{arg} \min_{\alpha} ||(\lambda \cdot \alpha)_T|| + \frac{\nu}{2}||\alpha - \beta||_2^2 + \frac{\mu}{2}||\alpha - (Wu^{k+1} + b^{k})||_2^2 + \frac{\gamma}{2}||\alpha - \alpha^{k}||_2^2 \\
b^{k+1} = b^{k} + Wu^{k+1} - \alpha^{k+1}
\end{array}\right.
\end{equation}
Fortunately,  each subproblem of (\ref{eq: DAL truncated l0l2}) has a closed form solution. Specifically,
\begin{equation}
\alpha^{k+1} =  \mathcal{H}_{T, \lambda, \nu, \mu, \gamma}(\beta, Wu^{k+1}+b^k, \alpha^k)
\end{equation}
where the operator $\mathcal{H}$ is a generalized component-wisely selective hard-threhsolding operator defined as follows:
\begin{equation} \label{eq:selective operator}
(\mathcal{H}_{T, \lambda, \nu, \mu, \gamma}(x,y,z))_i = \left\{ \begin{array}{ll}
0,\quad  \quad \quad  \mathrm{if} \quad  i \in T  \mathrm{and} | \frac{\nu x_i + \mu y_i + \gamma z_i}{\nu + \mu + \gamma} | < \frac{2\lambda_i}{\nu + \mu + \gamma}        \\
\frac{\nu x_i + \mu y_i + \gamma z_i}{\nu + \mu + \gamma}, \quad \mathrm{otherwise} \end{array} \right.
\end{equation}

We can see  that the above computation of $\alpha$ is in fact a selective hard-threhsolding procedure. It is known that the edges of an image should correspond to the large nonzero coefficients in the domain of wavelet transform. Assuming that we can acquire the reliable large nonzero support set, some components will not be shrunk if we believe that they are really nonzero elements via this selective hard-threhsolding operator, leading to better edge-enhanced recovered image.
As a multi-stage refinement of the corresponding non-convex $l_0$-$l_2$ model, our proposed method is expected to achieve an even better performance as long as the detected support information is reliable enough. Experiments confirmed the effectiveness of our proposed method.

When the $u$ is fixed, the update of $\beta$ relies on the frame coefficients $\alpha=Wu$, which is the same as (\ref{eq: beta}) and (\ref{eq: beta initial}). In addition,  the arithmetic means of the solution sequence is used as the final output.

\subsubsection{Component 2: Truncation determination based on iterative support detection}
In this part, we develop an effective support detection method for images with a fast decaying distribution of coefficients under wavelet frame transform. The support detection is based on thresholding, for $s$-th stage, the support indexes are obtained as follows:
\begin{equation} \label{eq: support detection}
I^{(s+1)}:= \{ \textbf{i}: |(Wu)_{\textbf{i}}| > \epsilon^{(s)} \}
\end{equation}
where $s = 0,1,2,\ldots$ and $T = I^C$. It should be pointed out that support index sets are not necessarily increasing or nested, i.e., $I^{(s)} \subset I^{(s+1)}$ may not hold for all $s$. This is very important because the support index set we get from the recent intermediate solution may contains wrong detections by $\epsilon^{(s)}$ thresholding. Not requiring $I^{(s)}$ to be monotonic leaves the chance for support index set to remove previous wrong detections, and makes $I^{(s)}$ less sensitive
to $\epsilon^{(s)}$. This way $\epsilon^{(s)}$ is easier to choose. Similarly to the \textit{threhsold}-ISD strategy in \cite{Wang2010}, we set
\begin{equation}\label{eq:threshold-value setting}
\epsilon^{(s)} := \textrm{max}\{|(Wu)_{\textbf{i}}^{(s)}|\}/\rho^{(s+1)}.
\end{equation}
with $\rho > 0$. An excessively large $\rho$ results in too many false support detections and leads to low solution quality, while an excessively small $\rho$ tends to cause a large number of iterations. It will be quite effective with an appropriate $\rho$, though the proper range of $\rho$ might be case-dependent. Empirically, the performance of is not very sensitive to the choice of $\rho$.

\subsubsection{The algorithmic framework for our proposed model}
Now we summarize the overall algorithmic framework for our proposed truncated $l_0$-$l_2$ model.  \\
\begin{tabular}{l}
\hline
\textbf{Algorithm 1} Wavelet frame based truncated $l_0$-$l_2$ Image Restoration  \\
\hline
Given an observed image $f$ and the convolution operator $A$. \\
1. Initialization: \\
\qquad Set the initial support estimation as empty set, i.e., $I^{(1)} = \emptyset $.\\
2. \textbf{Outer loop} (support detection): iteration on $s=1,2,\ldots,S$ \\
\qquad (a) \quad Update the support estimation via \textit{threshold}-ISD method (\ref{eq: support detection}). \\
\qquad (b) \quad \textbf{Inner loop} (solving the truncated $l_0$-$l_2$ model (\ref{eq:new truncatedl0l2})) \\
\qquad \qquad \qquad\textbf{ While} the stopping condition is not satisfied, iterate on $k=1,2,\ldots,K$ \textbf{Do}   \\
\qquad \qquad \qquad \qquad (I)   Image estimate $  u^{k+1} = (A^T + (\mu + \gamma)I)^{-1}(A^Tf + \gamma u^k + \mu W^T(\alpha^k - b^k)).  $                                   \\
\qquad \qquad \qquad \qquad (II)  Compute $ \alpha^{k+1} =  \mathcal{H}_{T, \lambda, \nu, \mu, \gamma}(\beta, Wu^{k+1}+b^k, \alpha^k)$ via (\ref{eq:selective operator}).  \\
\qquad \qquad \qquad \qquad (III) Update  $   b^{k+1} = b^{k} + Wu^{k+1} - \alpha^{k+1} $.       \\
\qquad \qquad \qquad \qquad (IV) If $mod(k,2)=0$. Update  $  \beta_{l,j}(\textbf{p}) = \sum_{i=1}^{m}\omega_i\alpha_{l,j}^{k+1}(\textbf{p}_i) $ via (\ref{eq: beta}).       \\
\qquad \qquad \qquad\textbf{ End}   \\

\hline
\end{tabular}
\\

 It is easy to see that our proposed algorithm is a multi-satge procedure. Note that since $I^{(1)} = \emptyset$, the first iteration of outer loop, i.e., the first stage of Algorithm 1, reduces to the standard $l_0$-$l_2$ model. We emphasize that the computational cost time of our proposed algorithm is not necessary several times more than that of Non-local MDAL method, if the looser stopping tolerance (except the final stage) and warm-starting strategy are adopted. Empirically, the best recovered results can be achieved when the outer iteration number is 2 or 3 in most cases. In this paper, the looser stopping strategy is not adopted for the purpose of better presenting the intermediate recovery results. We mainly pay our attention on the recovery quality and our proposed algorithm performs much better than the corresponding single-stage method in this aspect. However, the image quality is improved enough to be worth the extra computational cost. In such cases,  Algorithm 1 is often 2-3 times longer than Non-local MDAL method in our experiments.

\section{Numerical Experiments}\label{sec:experiments}
In this section, we compare the proposed algorithm (shown as Algorithm 1) with the PBOS algorithm for the non-local TV model \cite{Zhangxiaoqun2010}, the Split Bregman algorithm for the $l_1$ minimization problem \cite{Cai2009a}, the MDAL method for the $l_0$ minimization model \cite{Dong2013} and the non-local MDAL method for the $l_0$-$l_2$ minimization problem \cite{Chen2015}. The specific image restoration task that we consider is image deconvolution here, though other kinds of tasks can be considered in a similar
way. In the following experiments, four standard natural images (see Figure 1), which consist of complex components in different scales and with different patterns, are used in our tests. The intensity of a pixel of these test gray-scale images ranges from $0$ to $255$. We compare the quality of the recoveries not only in terms of  quantitative measurements (i.e., the increasing of PSNR and SSIM values, defined in the following part), but also the visibly detailed improvements of the restored images. All the experiments were performed under Windows 7 and MATLAB v7.10.0 (R2010a) running on a desktop with an Intel(R) Core(TM) i7-4790 CPU (3.60GHz) and 32GB of memory.

\subsection{Experiments Settings and Choices of Parameters }
\begin{figure}[h]
  \centering
   \includegraphics[width=0.2\textwidth]{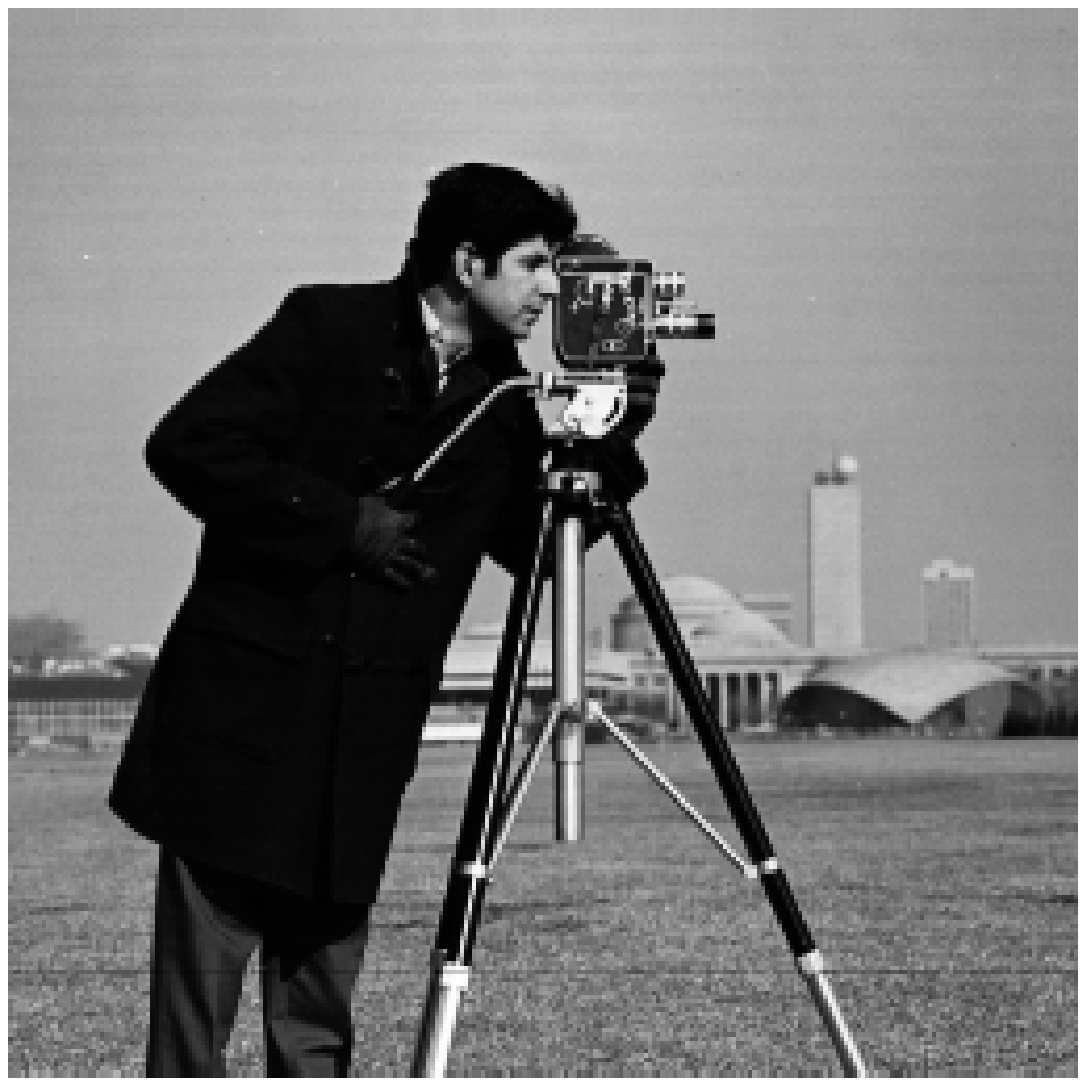}
   \includegraphics[width=0.2\textwidth]{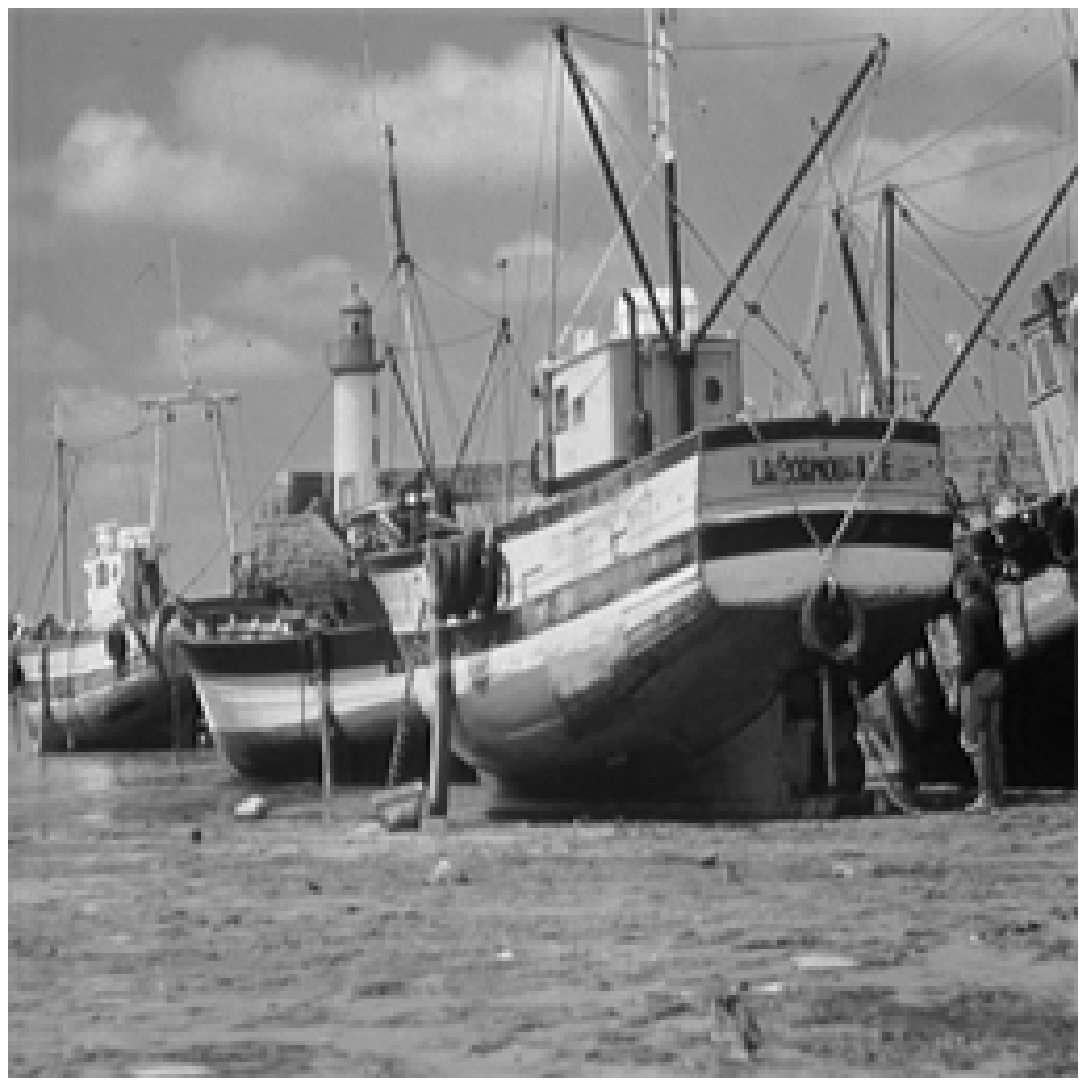}
   \includegraphics[width=0.3\textwidth]{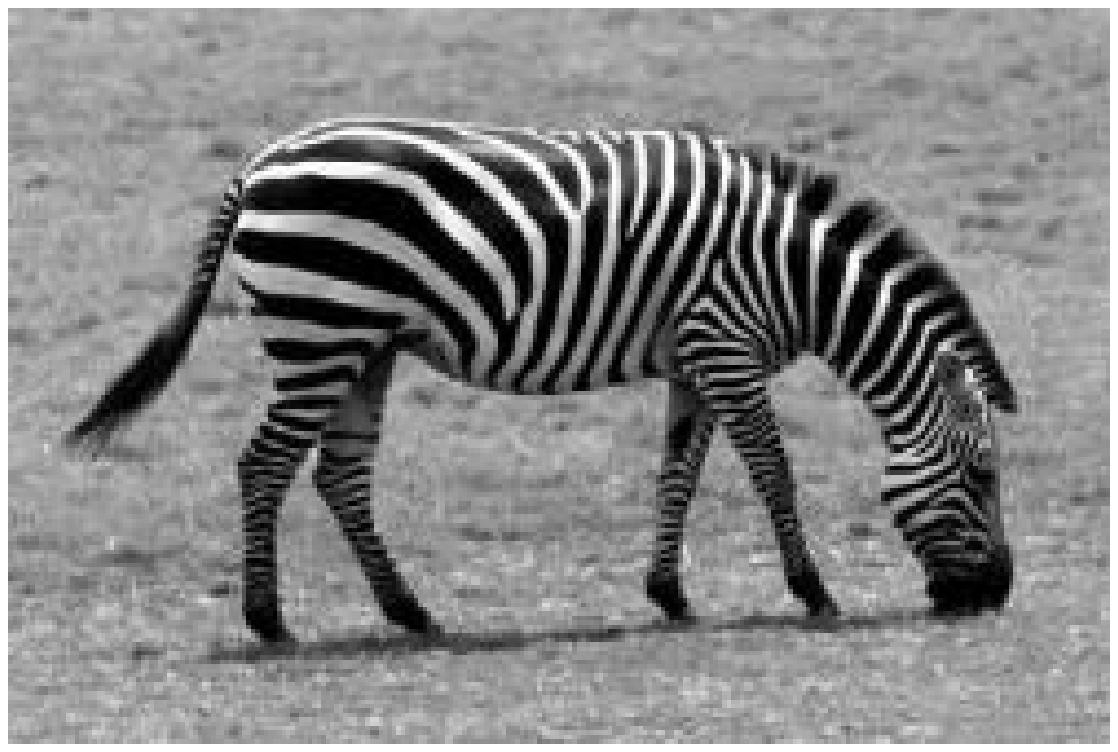}
   \includegraphics[width=0.28\textwidth]{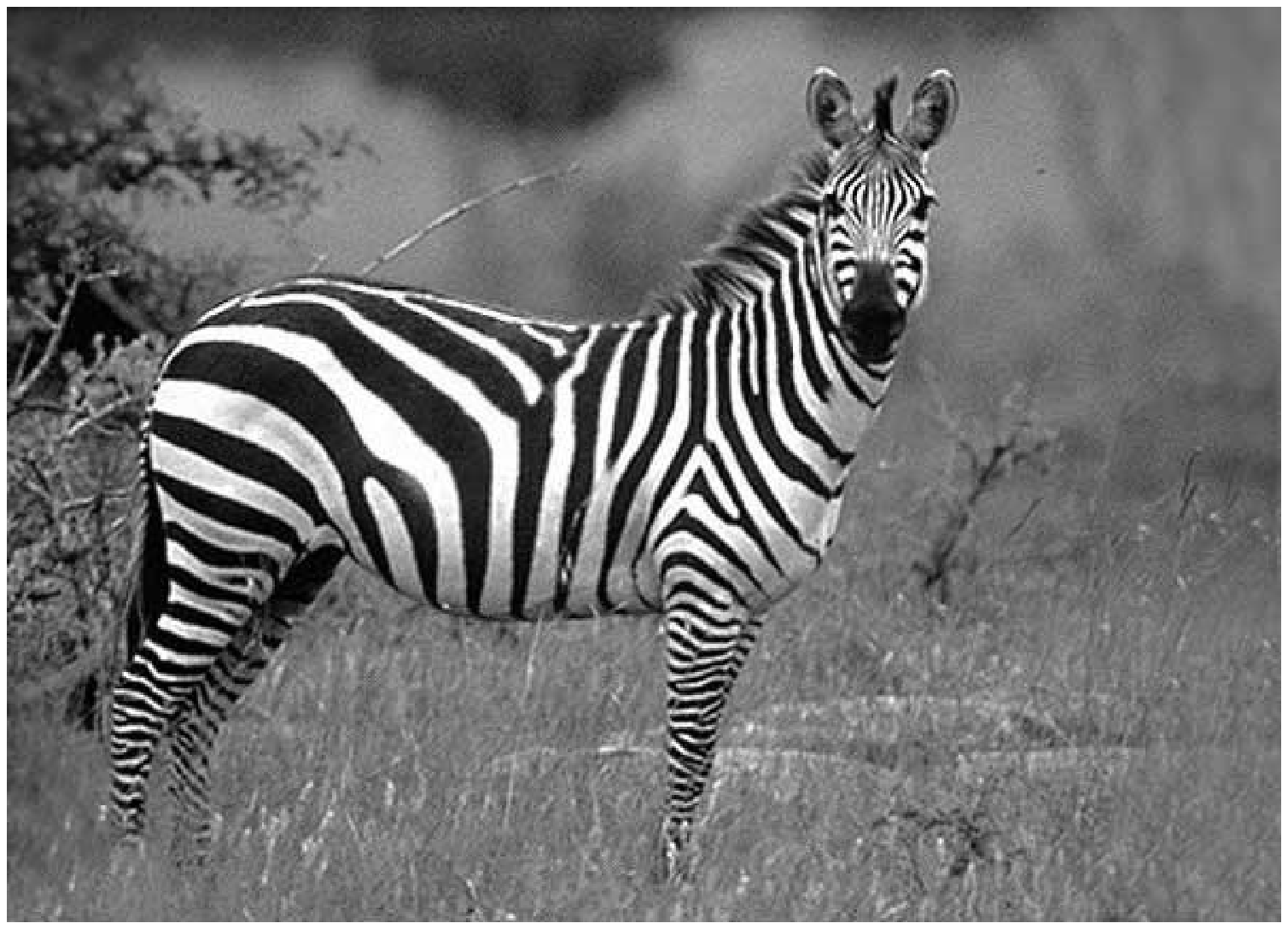}\\
  \caption{\small Original 4 test images (from the left to right: cameraman, boat, zebra1, zebra2, respectively). Sizes of them are: $256\times 256, 256\times 256, 250\times 167, 600\times 431$, respectively. }
\label{fig:testimages}
\end{figure}

The four blurred and noisy images in our tests are degraded as follows. Since the periodic boundary condition is used to generate the convolution operator in \cite{Cai2009a, Dong2013, Chen2015}, we use the same boundary condition to blur the test images. Four types of blurring kernels are used, type I: fspecial (¡¯motion¡¯,10,20), type II: fspecial (¡¯gaussian¡¯,25,1.6), type III: $9\times 9$ uniform and type IV: fspecial (¡¯motion¡¯,15,30). The blurred images are further corrupted by Gaussian noise with zero mean and standard deviation of $\sigma=3.0, 4.0$ in kernel type I, $\sigma=\sqrt{2}$ in type II, $\sigma=\sqrt{0.3}$ in type III, $\sigma=3.0, 4.0$ in type IV, respectively.

\begin{table*} \rotatebox{90}   { 
\centering
\begin{tabular} {|c|c|c|c|c|c|c|c|c|}
\hline 
  \multirow{2}{*}{Test image} & Blur type& \multicolumn{1}{|c|}{Non-local TV \cite{Zhangxiaoqun2010}} &  \multicolumn{1}{|c|}{Split Bregman \cite{Cai2009a}}   & \multicolumn{1}{|c|}{ MDAL \cite{Dong2013}} & \multicolumn{1}{|c|}{ Non-local MDAL \cite{Chen2015}} & \multicolumn{1}{|c|}{ Proposed} & \multicolumn{1}{|c|}{ Oracle}     \\
 \cline{3-8}
     &             /Noise level                & \multicolumn{1}{|c|}{PSNR/SSIM} & \multicolumn{1}{|c|}{PSNR/SSIM}    & \multicolumn{1}{|c|}{PSNR/SSIM} & \multicolumn{1}{|c|}{PSNR/SSIM} &    \multicolumn{1}{|c|}{PSNR/SSIM}    &    \multicolumn{1}{|c|}{PSNR/SSIM}    \\
 \hline
  \multirow{6}{*}{Cameraman}&  Type I / $\sigma=3.0$  &  27.48/0.8230   &  27.23/0.8414	& 28.38/0.8487  &  28.32/0.8568  & \textbf{28.85/0.8589} &  32.42/0.9228 \\
    \cline{2-8}
    & Type I / $\sigma=4.0$ &    26.67/0.7889   	&  26.38/0.8200   &  26.93/0.8248   &  27.43/0.8311& \textbf{27.68/0.8327}  &  30.06/0.8601 \\
   \cline{2-8}
    & Type II / $\sigma=\sqrt{2}$   &    27.08/0.8543     &  27.03/0.8596	&  27.25/0.8655   &   27.49/0.8684   &   \textbf{27.62/0.8695}  &  33.64/0.9243  \\
     \cline{2-8}
    & Type III / $\sigma=\sqrt{0.3}$      &  28.21/0.8733        &	27.91/0.8669    &  28.85/0.8875  &   28.87/0.8864 &   \textbf{29.60/0.8929}   &  33.26/0.9121  \\
     \cline{2-8}
    & Type IV / $\sigma=3.0$      &   26.69/0.7826  &    26.42/0.8147      &  27.61/0.8289	  &   27.48/0.8317        &   \textbf{28.03/0.8349}    &  31.91/0.9149   \\
     \cline{2-8}
    & Type IV / $\sigma=4.0$       &  25.90/0.7416  &	25.63/0.7966    &  26.44/0.8003  &  26.73/0.8066  &  \textbf{27.10/0.8091}  &  29.61/0.8473\\
     \cline{2-8}
\hline
\hline
  \multirow{6}{*}{Boat}&  Type I / $\sigma=3.0$  &   27.67/0.8168     &    27.43/0.8137  &   28.26/0.8305   &    28.40/0.8375      &   \textbf{28.59/0.8394}  & 32.40/0.9243     \\
   \cline{2-8}
    & Type I / $\sigma=4.0$ &  26.93/0.7848      &   26.65/0.7869      &  27.12/0.7991  &   27.51/0.8096   &  \textbf{27.64/0.8115}  &  29.88/0.8892  \\
    \cline{2-8}
    & Type II / $\sigma=\sqrt{2}$  &   27.93/0.8399     &   28.02/0.8390     &    28.09/0.8501     &   28.34/0.8501       &    \textbf{28.43/0.8512 }  &  33.46/0.9220   \\
     \cline{2-8}
    & Type III / $\sigma=\sqrt{0.3}$ &   28.89/0.8563     &   28.27/0.8401      &   29.26/0.8684      &   29.29/0.8704       &   \textbf{29.87/0.8776 }   &  32.33/0.8956    \\
  \cline{2-8}
    & Type IV / $\sigma=3.0$ &   26.80/0.7725     &    26.58/0.7663     &   27.35/0.7952      &   27.44/0.7992       &    \textbf{27.73/0.8043 }   &  31.71/0.9121  \\
   \cline{2-8}
    & Type IV / $\sigma=4.0$ &   25.96/0.7361     &    25.73/0.7383     &   26.43/0.7557      &   26.67/0.7672       &    \textbf{26.82/0.7709  }    &  29.33/0.8824 \\
\hline
\hline
  \multirow{6}{*}{Zebra1}&  Type I / $\sigma=3.0$  &   25.75/0.8243     &  24.53/0.8048  &   25.85/0.8317   &  26.14/0.8353   & \textbf{26.94/0.8410}   & 30.15/0.9206  \\
    \cline{2-8}
    & Type I / $\sigma=4.0$  &  24.71/0.7942      &  23.60/0.7684       &   24.73/0.7996      &   25.25/0.8088       &    \textbf{25.67/0.8122 }    &  27.58/0.8900  \\
    \cline{2-8}
    & Type II / $\sigma=\sqrt{2}$ &  25.42/0.8313      &  25.58/0.8307       &   25.45/0.8230      &    25.85/0.8327      &    \textbf{26.05/0.8337}  &  30.58/0.9074    \\
     \cline{2-8}
    & Type III / $\sigma=\sqrt{0.3}$ &  26.05/0.8298      &   25.09/0.8185      &   25.89/0.8299      &    25.76/0.8233      &   \textbf{26.46/0.8348}  &  28.36/0.8882    \\
    \cline{2-8}
    & Type IV / $\sigma=3.0$ &   24.25/0.7478     &  23.55/0.7389       &  24.71/0.7705       &    24.80/0.7699      &     \textbf{25.63/0.7806 }  &  29.33/0.8923    \\
    \cline{2-8}
    & Type IV / $\sigma=4.0$ &    23.16/0.7111    &  22.77/0.7042       &   23.87/0.7358      &    24.11/0.7419      &    \textbf{24.69/0.7501}  &  26.96/0.8633   \\
    \hline
    \hline
  \multirow{6}{*}{Zebra2}&  Type I / $\sigma=3.0$ &    27.67/0.8008    &  27.16/0.8114   &   28.56/0.8250   &  28.58/0.8287   &  \textbf{29.19/0.8363 }  & 32.37/0.8991    \\
  \cline{2-8}
    & Type I / $\sigma=4.0$ &   26.66/0.7696     &   26.25/0.7851      &   27.51/0.7973  & 27.80/0.8027  &  \textbf{28.12/0.8034 } & 30.29/0.8717     \\
  \cline{2-8}
    & Type II / $\sigma=\sqrt{2}$ &   29.79/0.8758     &   30.00/0.8757  &  29.34/0.8705       &    \textbf{30.28/0.8780}  & 30.26/0.8771  &  33.69/0.9020    \\
  \cline{2-8}
    & Type III / $\sigma=\sqrt{0.3}$ &  28.37/0.8381      &   27.22/0.8132   &    28.37/0.8403  &  28.41/0.8445 &  \textbf{29.33/0.8518 }  &  32.23/0.8702     \\
  \cline{2-8}
    & Type IV / $\sigma=3.0$ &   26.15/0.7633     &   27.25/0.7720      &   27.25/0.7948      &   27.06/0.7966       &    \textbf{27.94/0.8017 }  &  31.51/0.8828    \\
    \cline{2-8}
    & Type IV / $\sigma=4.0$ &   25.21/0.7256     &  24.96/0.7481       &   26.24/0.7660      &    26.36/0.7708      &    \textbf{26.92/0.7736 }  &  29.64/0.8606      \\
\hline
\end{tabular}}
 \caption{ The comparison of SNR and SSIM values between proposed algorithm with non-local method \cite{Zhangxiaoqun2010}, wavelet frame based split bregman method \cite{Cai2009a}, MDAL method \cite{Dong2013}, non-local MDAL method \cite{Chen2015}. Bold values denote the highest PSNR or SSIM values besides the Oracle cases.}  \label{Table:1}
\end{table*}

\begin{table}{  
\centering
\begin{tabular}{|c|c|c|c|c|c|}
\hline 
  \multirow{2}{*}{Test image} & Blur type& \multicolumn{1}{|c|}{IDD-BM3D: Initial \cite{Dabov2008}} &  \multicolumn{1}{|c|}{IDD-BM3D: Final \cite{Danielyan2012}}   & \multicolumn{1}{|c|}{ Proposed }  \\
 \cline{3-5}
     &             /Noise level                & \multicolumn{1}{|c|}{PSNR/SSIM} & \multicolumn{1}{|c|}{PSNR/SSIM/(Time)}    & \multicolumn{1}{|c|}{PSNR/SSIM/(Time)}   \\
 \hline
  \multirow{3}{*}{Cameraman}
   & Type II / $\sigma=\sqrt{2}$     &   27.46/0.8415      &	 \textbf{28.18}/0.8681/(86.46)    &    27.62/\textbf{0.8694} /(63.97)   \\
    \cline{2-5}
    & Type IV / $\sigma=3.0$      &  27.19/0.8131     &    \textbf{28.60}/\textbf{0.8451} /(85.61)     &  	 27.91/0.8350/(63.65)       \\
    \cline{2-5}
    & Type IV / $\sigma=4.0$       &   26.40/0.7944  &	 \textbf{27.47}/\textbf{0.8254} /(85.63)   &   27.04/0.8091/(72.03)   \\
    \cline{2-5}
\hline
\hline
  \multirow{3}{*}{Zebra1}

    & Type II / $\sigma=\sqrt{2}$ &  25.78/0.8097      &   \textbf{26.12}/\textbf{0.8384} /(61.45)     &  26.00/0.8335 /(35.89)          \\
    \cline{2-5}
    & Type IV / $\sigma=3.0$ &   24.64/0.7112    &  \textbf{25.64}/0.7729/(67.44)       &  25.50/\textbf{0.7791} /(54.01)           \\
    \cline{2-5}
    & Type IV / $\sigma=4.0$ &    23.94/0.6840    &  \textbf{24.72}/0.7356/(67.97)       &   24.63/\textbf{0.7494} /(58.36)        \\
    \hline

\end{tabular}}
\caption{ Performance comparison of the proposed method with BM3D \cite{Dabov2008, Danielyan2012}. The given values are PSNR (dB)/SSIM/CPU time (second). Bold values denote the highest PSNR or SSIM values.} \label{Table:BM3D}
\end{table}

In Figure 2, we show that the magnitude of sorted wavelet frame coefficients of the 4 tested images has a fast decaying property, which demonstrates that the fast decaying condition under wavelet frame transform is a mild assumption. The quality of recovered image is quantitatively measured by means of the peak signal-to-noise ratio (PSNR) defined as
\begin{equation}
\mathrm{PSNR}(u,u^*) = -20\mathrm{lg}\left \{ \frac{||u-u^*||_2}{255N} \right \}
\end{equation}
where $u$ and $u^*$ denote the original and recovered images, respectively. $N$ denotes the total number of pixels in the image $u$. In addition, we also use another image quality assessment: Structural SIMilarity (SSIM) index between two images proposed in \cite{Wang2004}, which aims to be more consistent with human eye perception. Besides, we choose the following stopping criterion for all the wavelet frame based methods:
\begin{equation}
\mathrm{min}\left\{\frac{||u^k-u^{k-1}||_2}{||u^k||_2},\frac{||Au^k-f||_2}{||f||_2}\right\}<5\times10^{-4}
\end{equation}

\begin{figure}[h]
  \centering
   \includegraphics[width=0.40\textwidth]{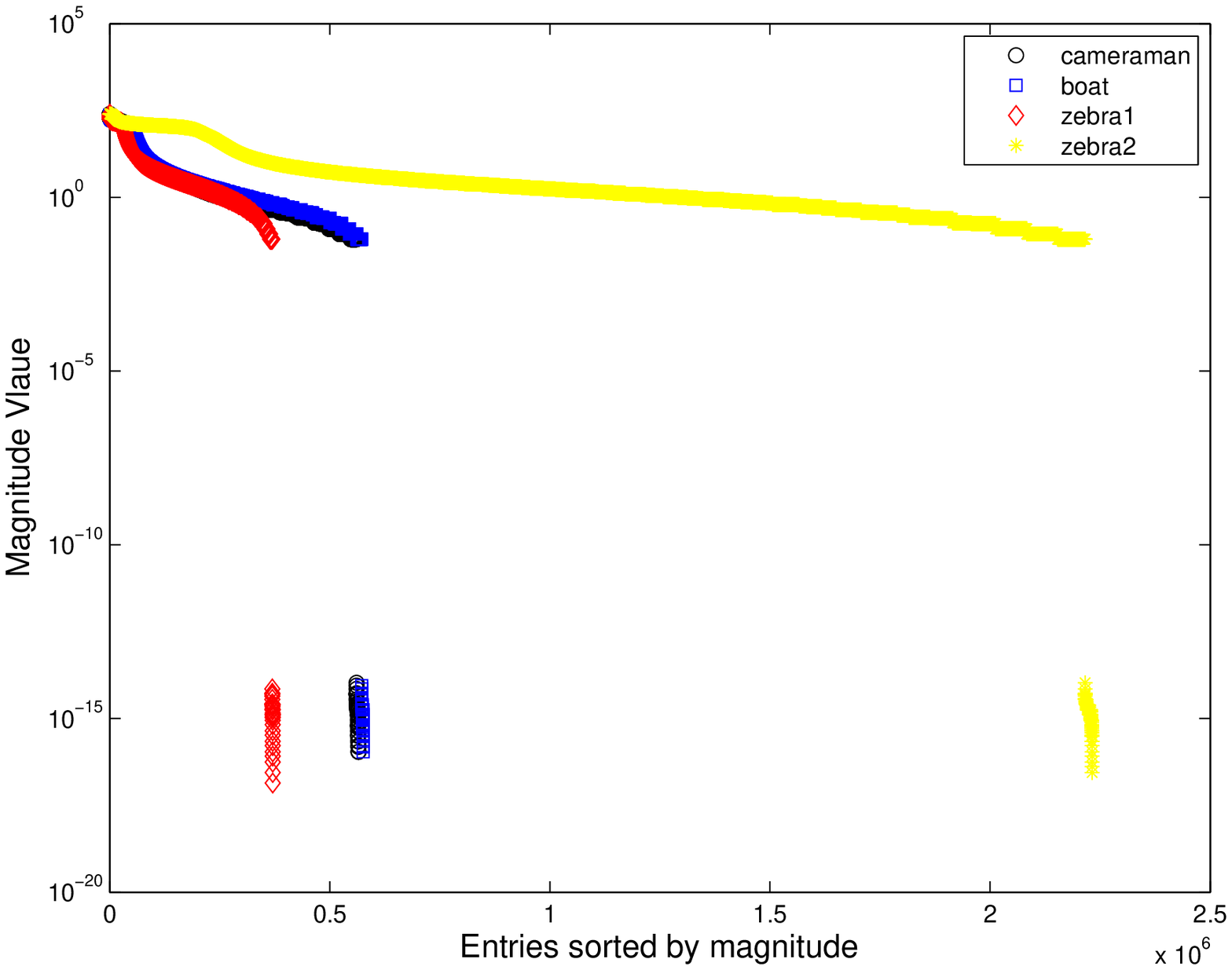}\\
  \caption{\small The sorted magnitudes of the wavelet frame coefficients of the 4 test images have the property of fast decaying property.}
\label{fig:fastdecaying}
\end{figure}

\begin{figure}[h]
  \centering
   \includegraphics[width=0.22\textwidth]{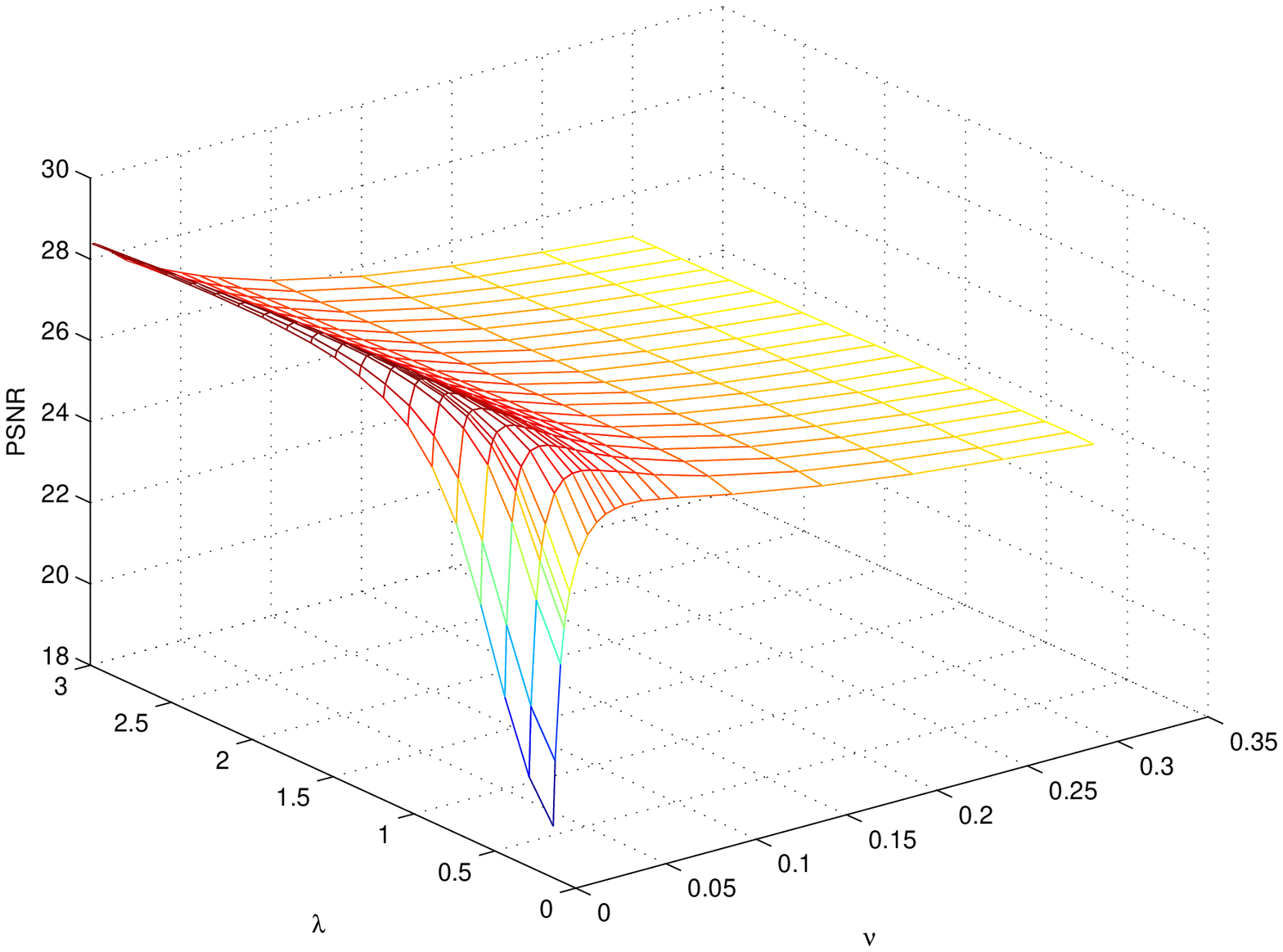}
   \includegraphics[width=0.22\textwidth]{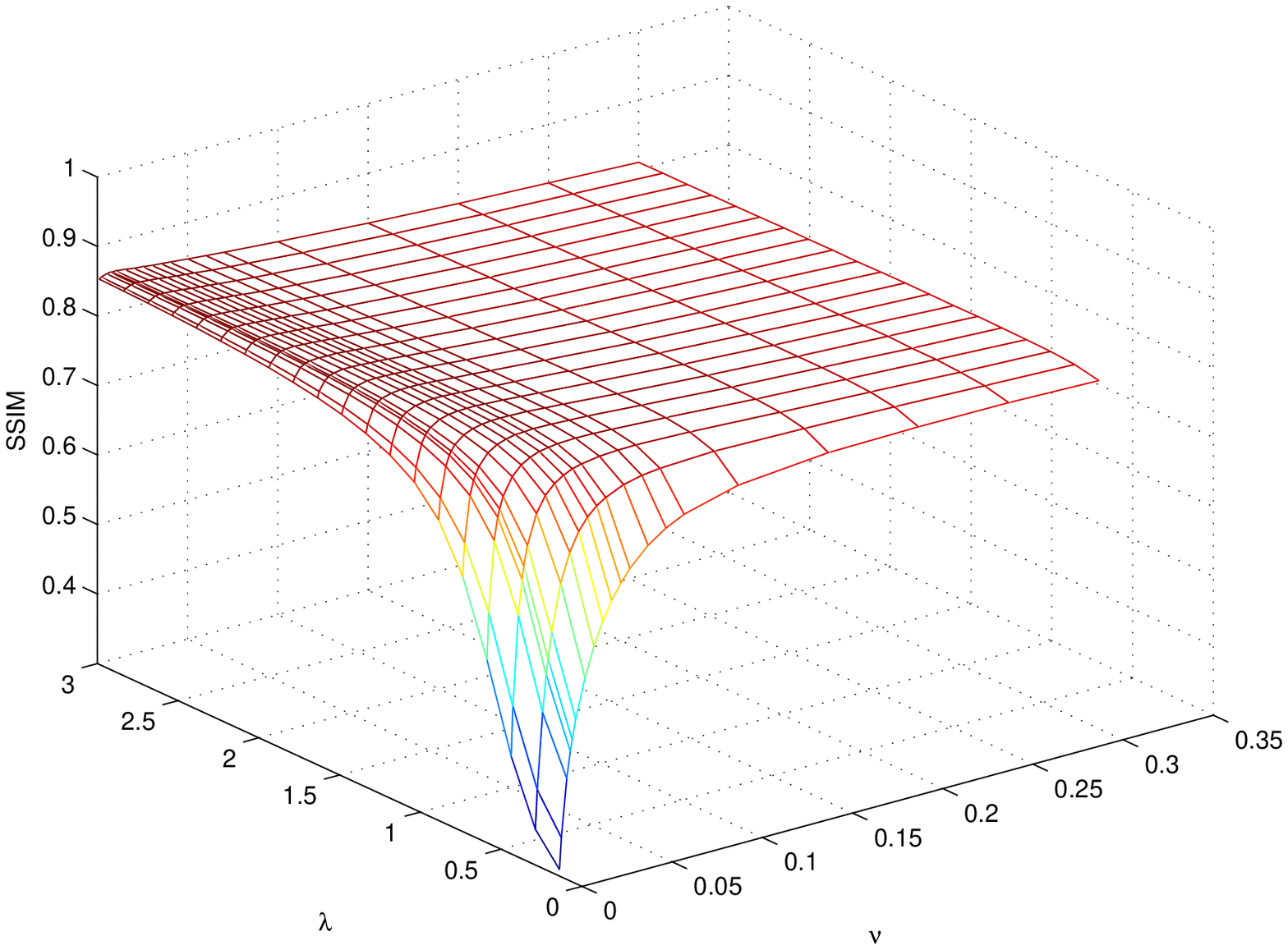}
   \includegraphics[width=0.22\textwidth]{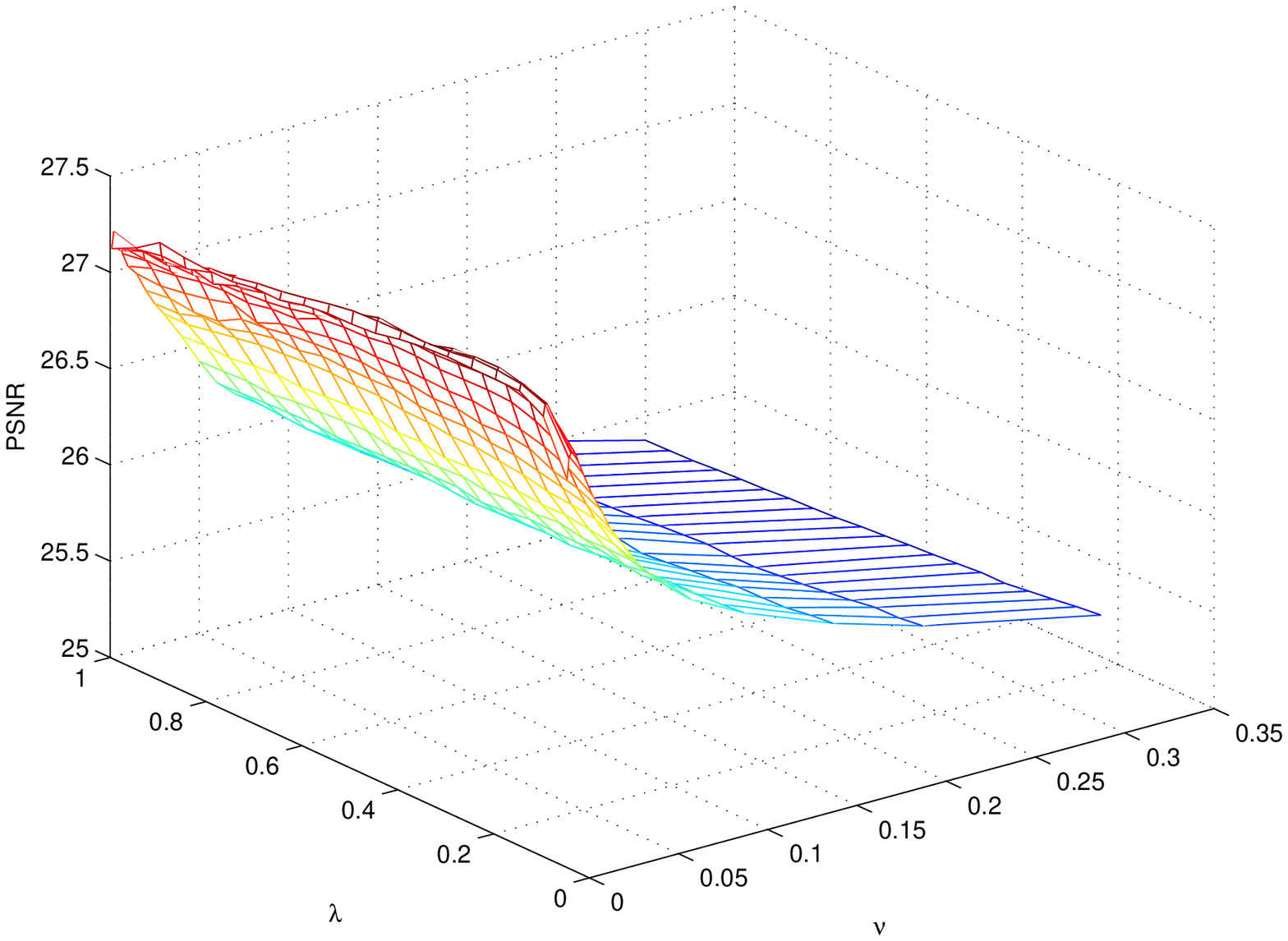}
   \includegraphics[width=0.22\textwidth]{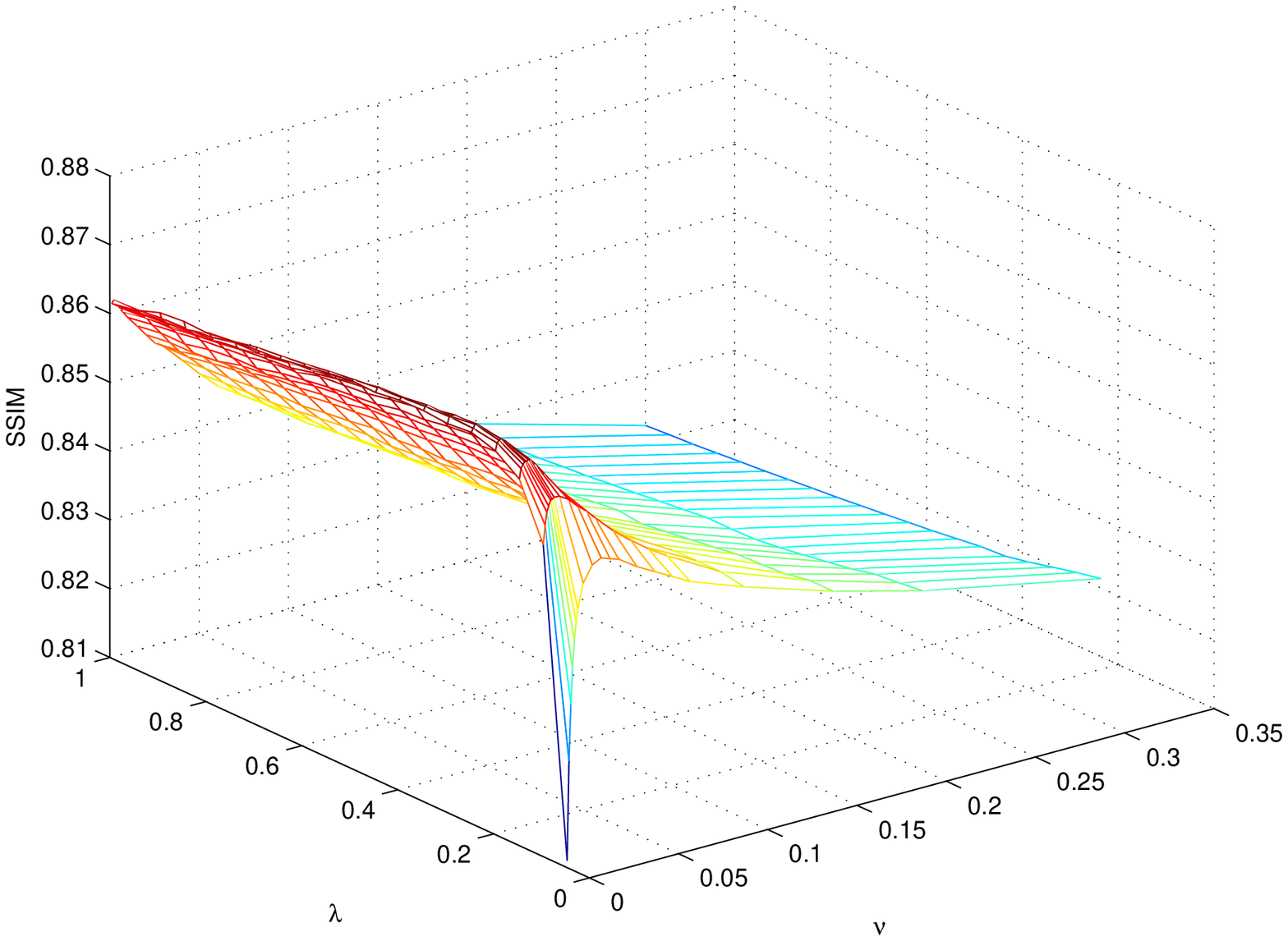}\\
  \caption{\small  The PSNR and SSIM values of Non-local MDAL method with the parameters $\lambda$ and $\nu$ vary. The first column and second column correspond to the recovery results of blur kernel Type I with noise level $\sigma=3.0$. The third column and fourth column correspond to the recovery results of blur kernel Type II with noise level $\sigma=\sqrt{2}$. Test image: cameraman.}
\label{fig:lambda_nu}
\end{figure}

According to the suggestion of \cite{Chen2015}, linear B-spline framelet is adopted for our experiments. Considering that the framelet decomposition level to be 1 is a commonly desirable choice \cite{Chen2015}, we fix it to be 1 (i.e., $L=1$) for all wavelet frame based methods, for fair comparison. For all the cases, we fix the penalty parameter $\mu=0.05$ for the split bregman method, and we fix $\mu=0.01$, $\gamma=0.003$ for the MDAL method. In addition, we select the best regularization parameter $\lambda$ for optimal image recoveries. For the Non-local MDAL method, based on the suggestion in \cite{Chen2015},  the sizes of image patches and the searching window are fixed to be $5\times 5$ and $11\times 11$ respectively, for the computation of the nonlocal weights. The 15 nearest patches (i.e., $m=15$) are used for the computation of the nonlocal prior. We fix the penalty parameter $\mu=0.01$ and $\gamma=0.003$, and the regularization parameter $\lambda$ and $\nu$ are selected for optimal PSNR and SSIM values. In Figure \ref{fig:lambda_nu}, we show the PSNR and SSIM variation trends as the parameters $\lambda$ and $\nu$ vary. Here for the conciseness of the paper, we only show the observations of test image cameraman, since the other cases have the similar conclusions.
For our proposed Algorithm 1, for simplicity, the parameters are set to be the same as Non-local MDAL method. In addition, we set $\rho=3$ in (\ref{eq:threshold-value setting}). Empirically, the performance of the proposed algorithm is unsensitive to the settings of these parameters. We believe our comparison will be fair under these specific settings.

In what follows, we also show the oracle recovered results of our proposed algorithm, i.e., the support estimation in (\ref{eq: support detection}) is based on the underlying true image. Clearly, this case is not possible in practice. However, it demonstrates the advantage by exploiting the support prior of frame coefficients, which serves as a benchmark and chalks out a path for us to explore: serving as an ideal golden upper bound of the performance of support detection based methods.

\begin{figure}[h]
  \centering
   \includegraphics[width=0.22\textwidth]{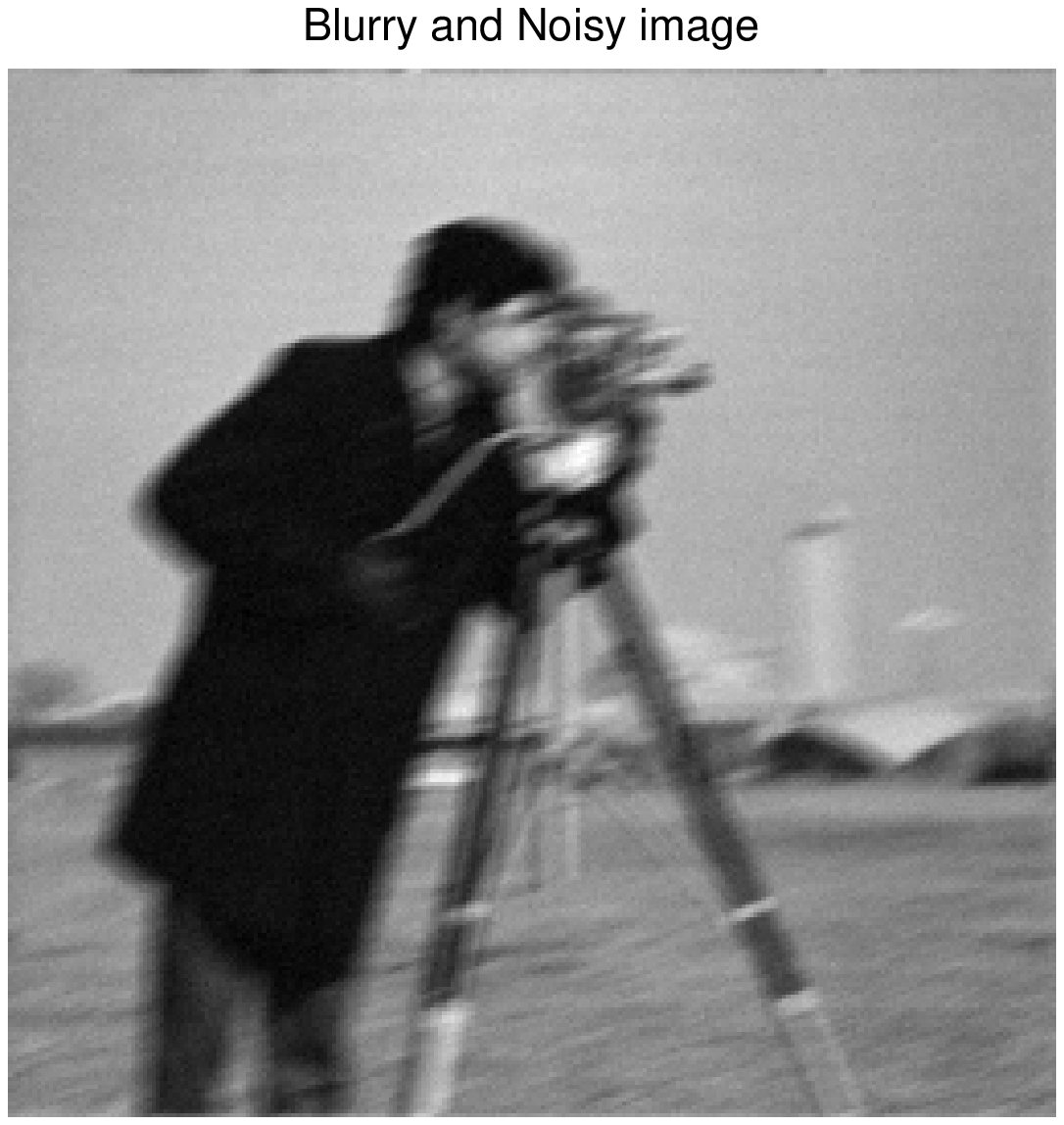}
   \includegraphics[width=0.22\textwidth]{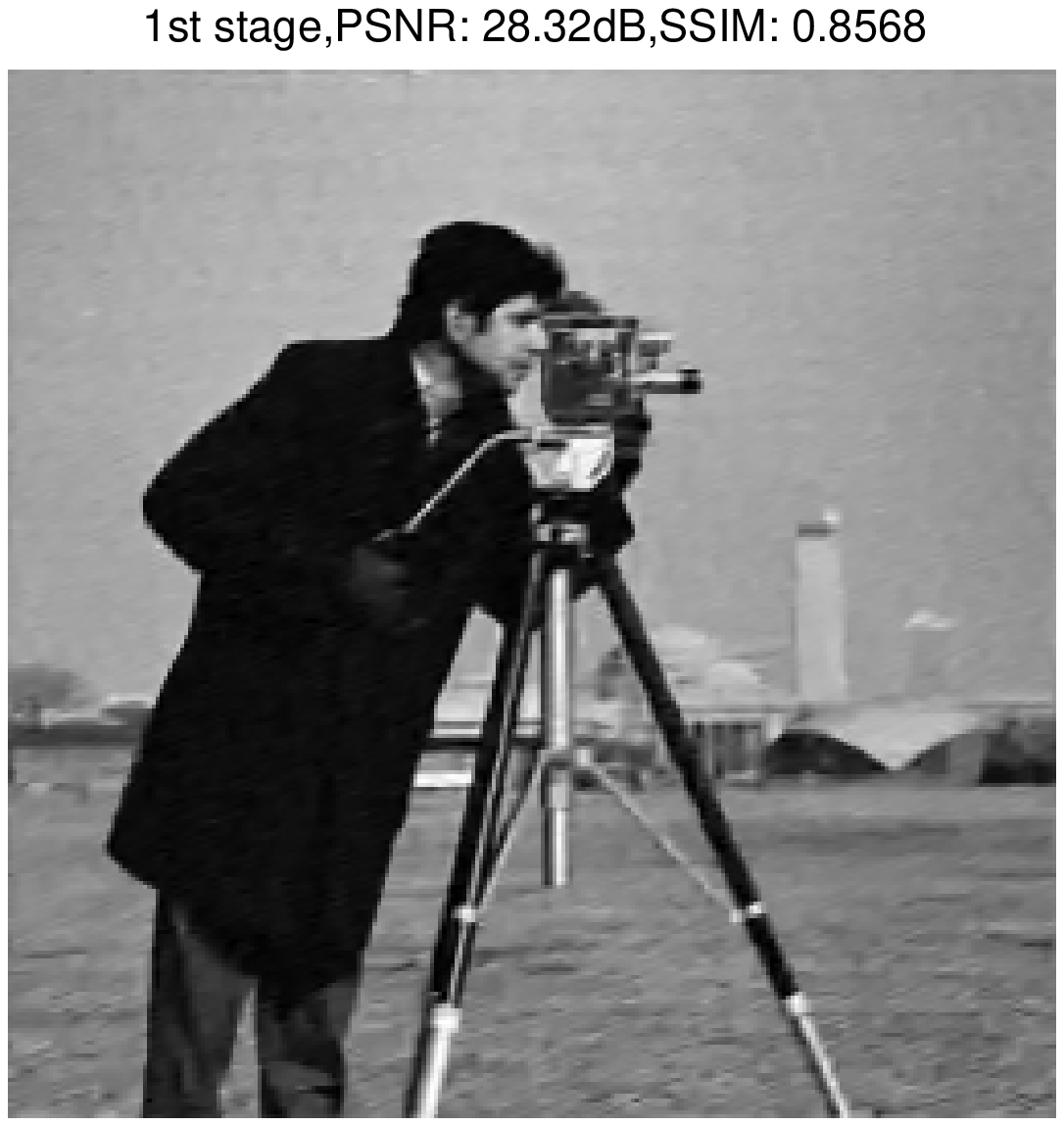}
   \includegraphics[width=0.22\textwidth]{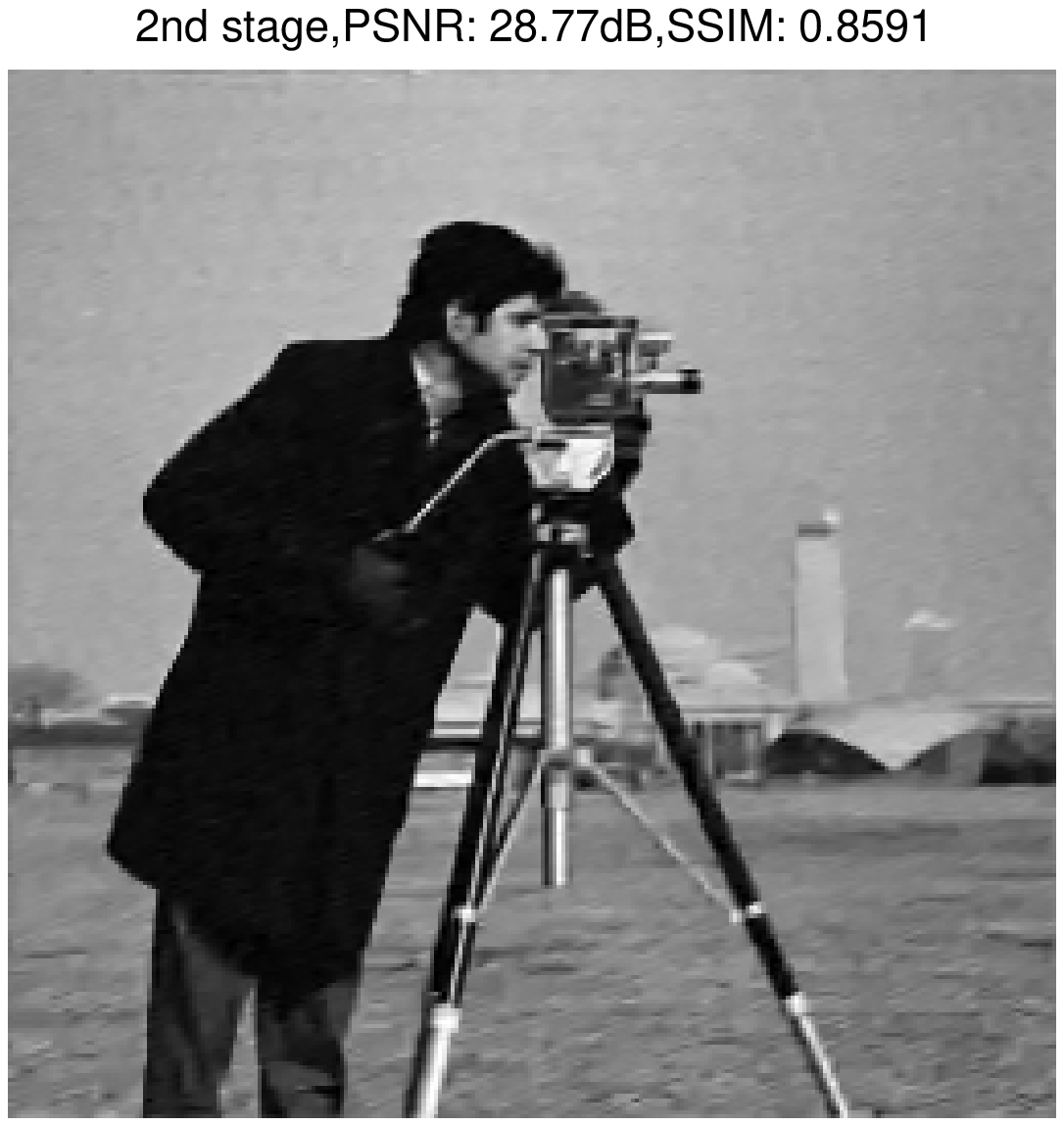}
   \includegraphics[width=0.22\textwidth]{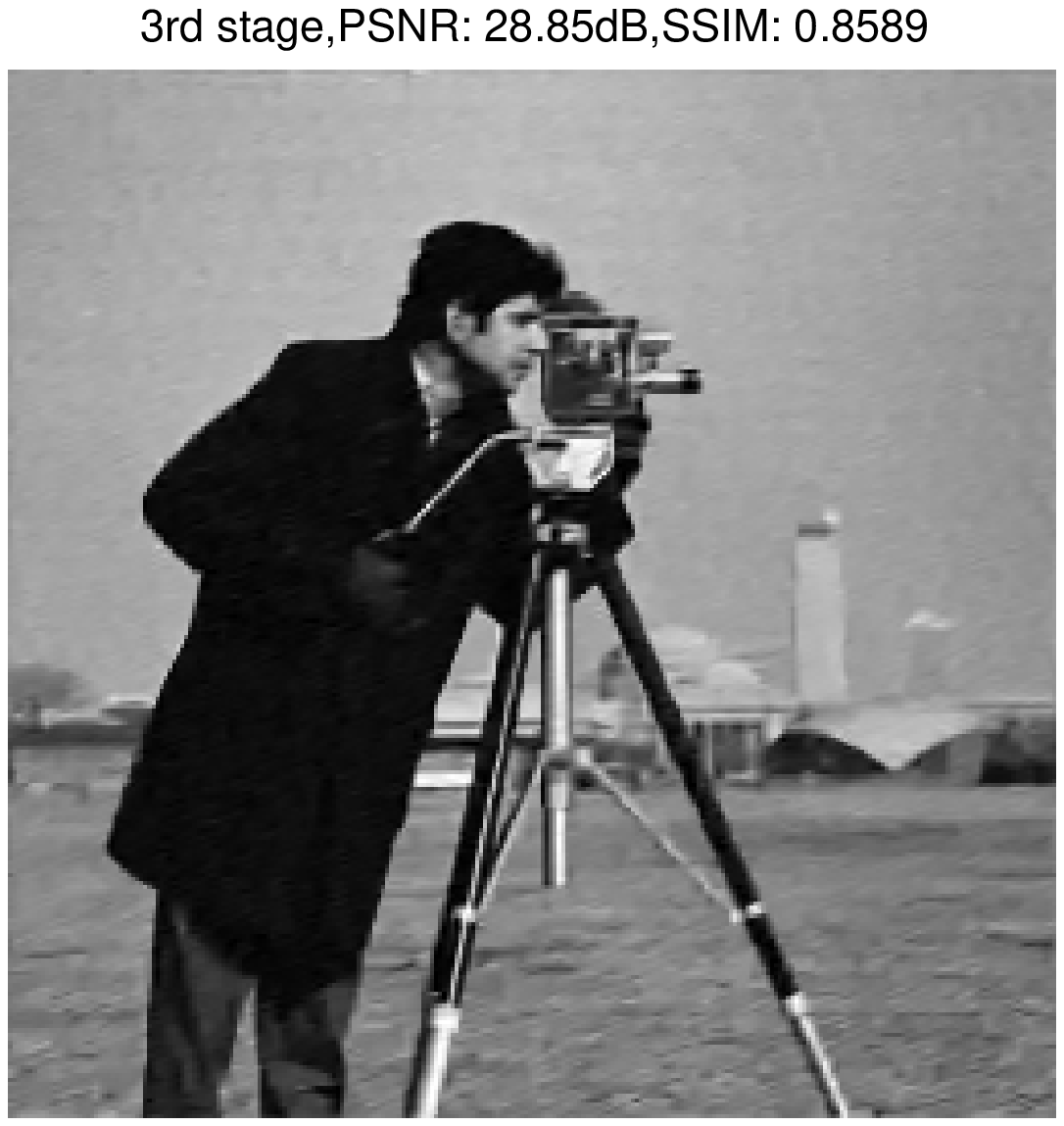}\\
   \includegraphics[width=0.22\textwidth]{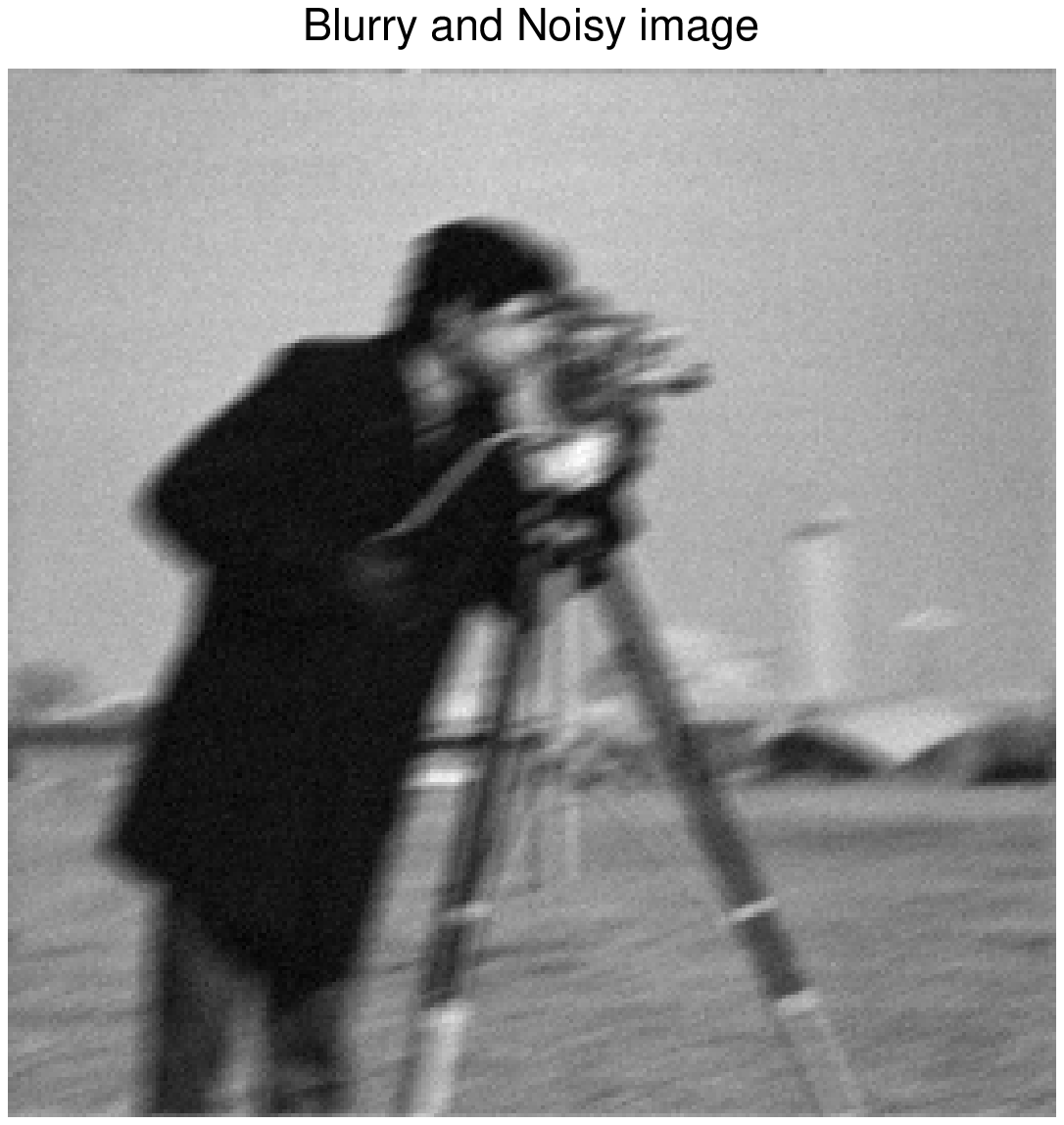}
   \includegraphics[width=0.22\textwidth]{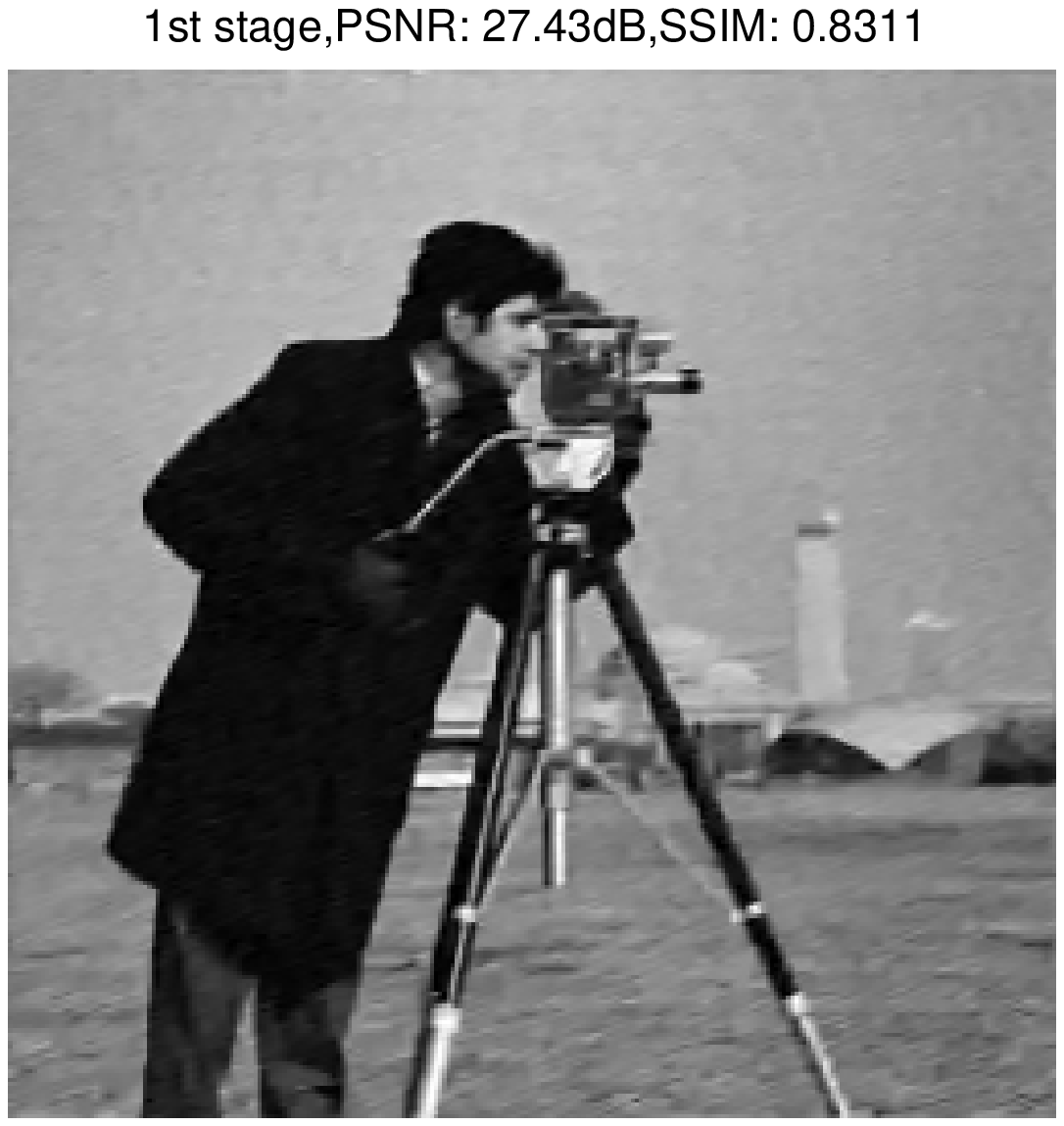}
   \includegraphics[width=0.22\textwidth]{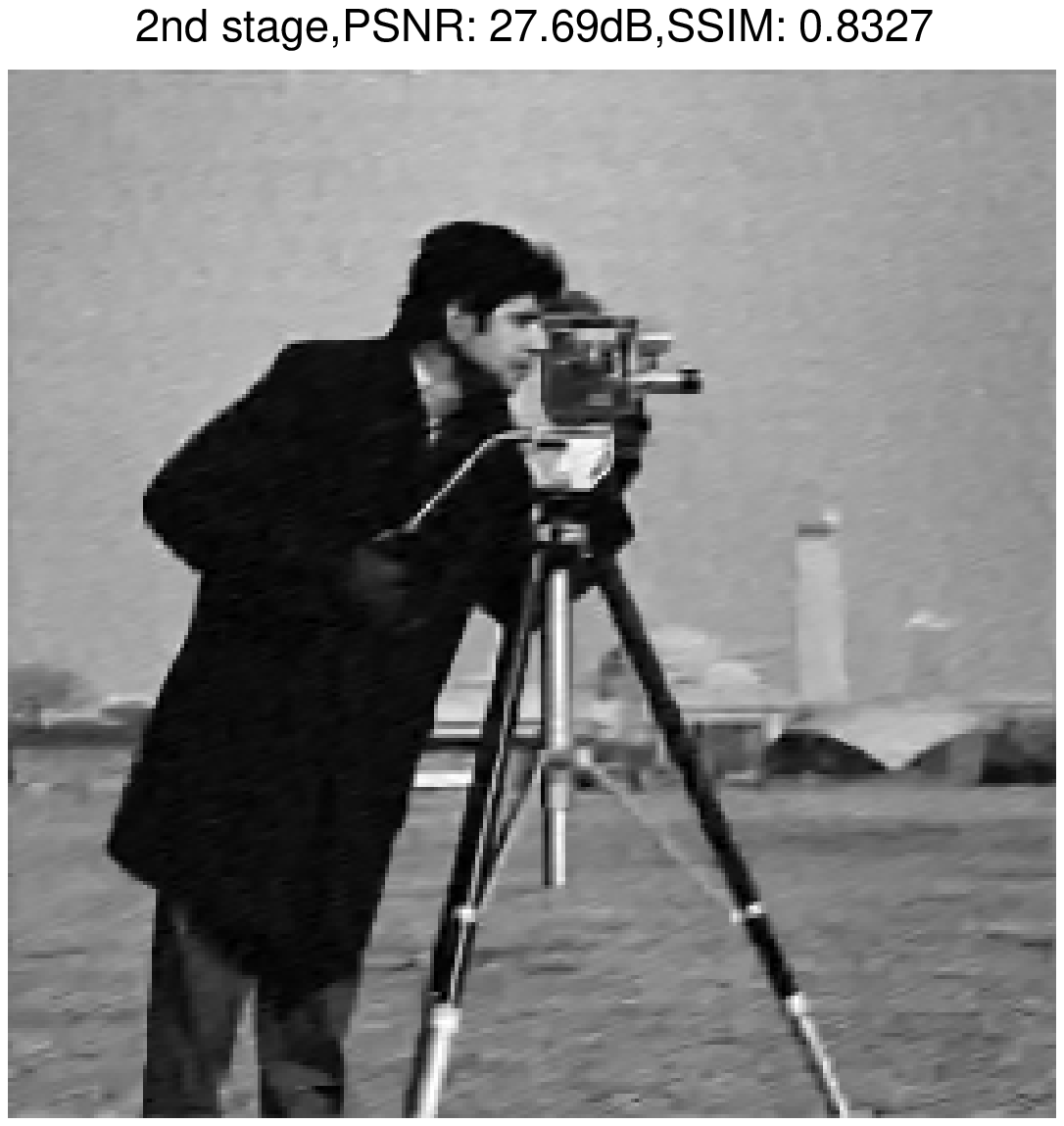}
   \includegraphics[width=0.22\textwidth]{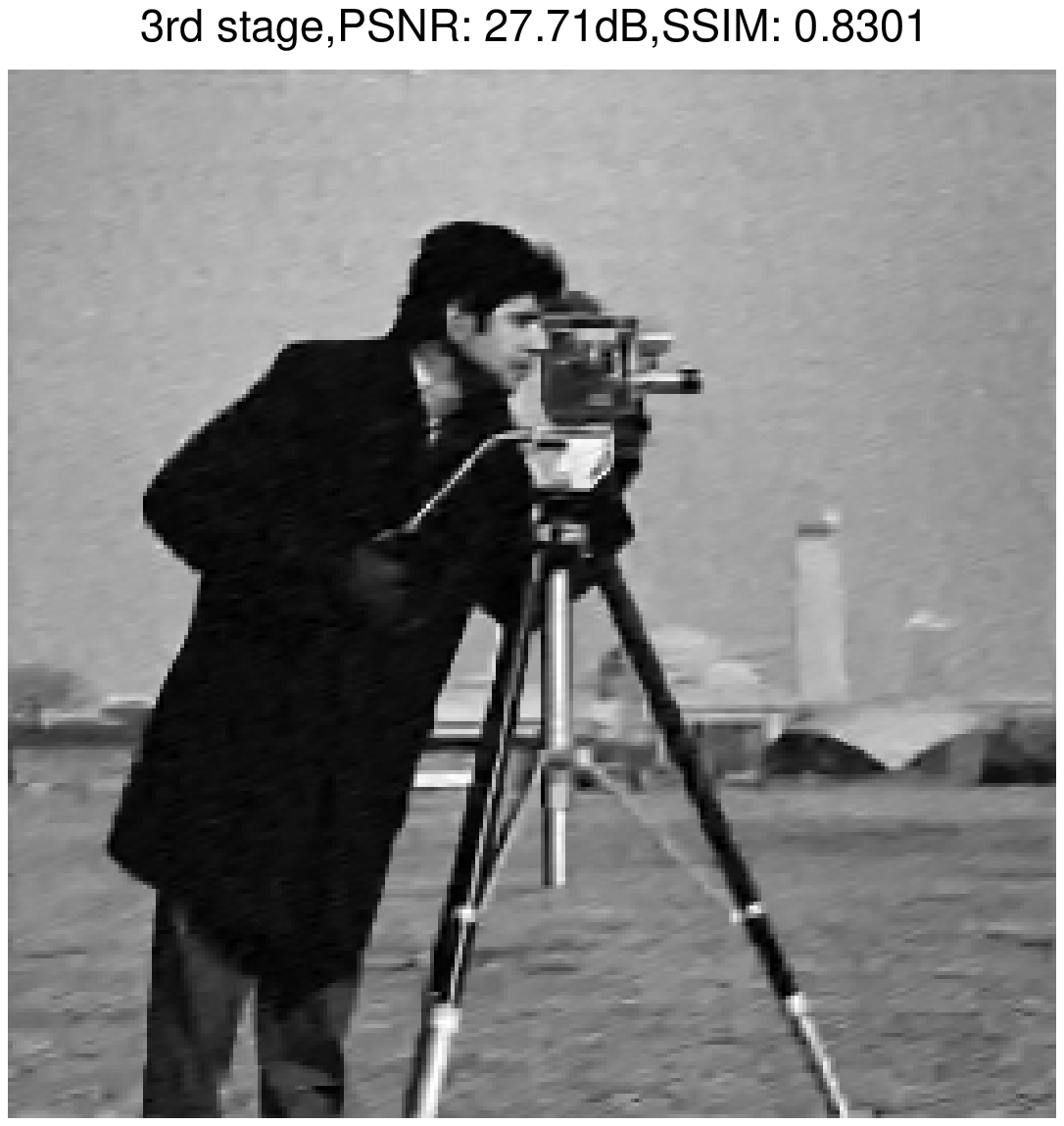}\\
   \includegraphics[width=0.22\textwidth]{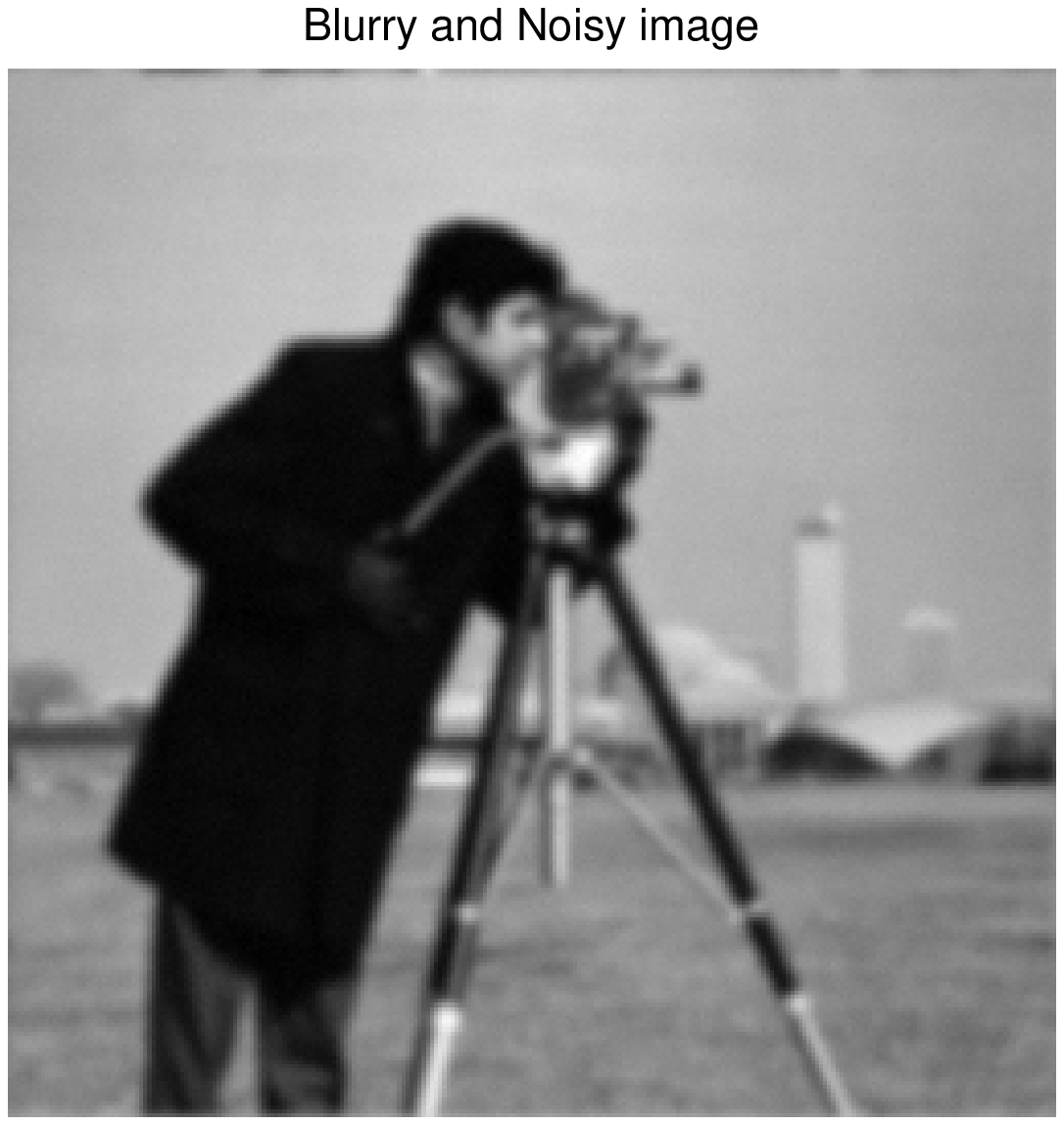}
   \includegraphics[width=0.22\textwidth]{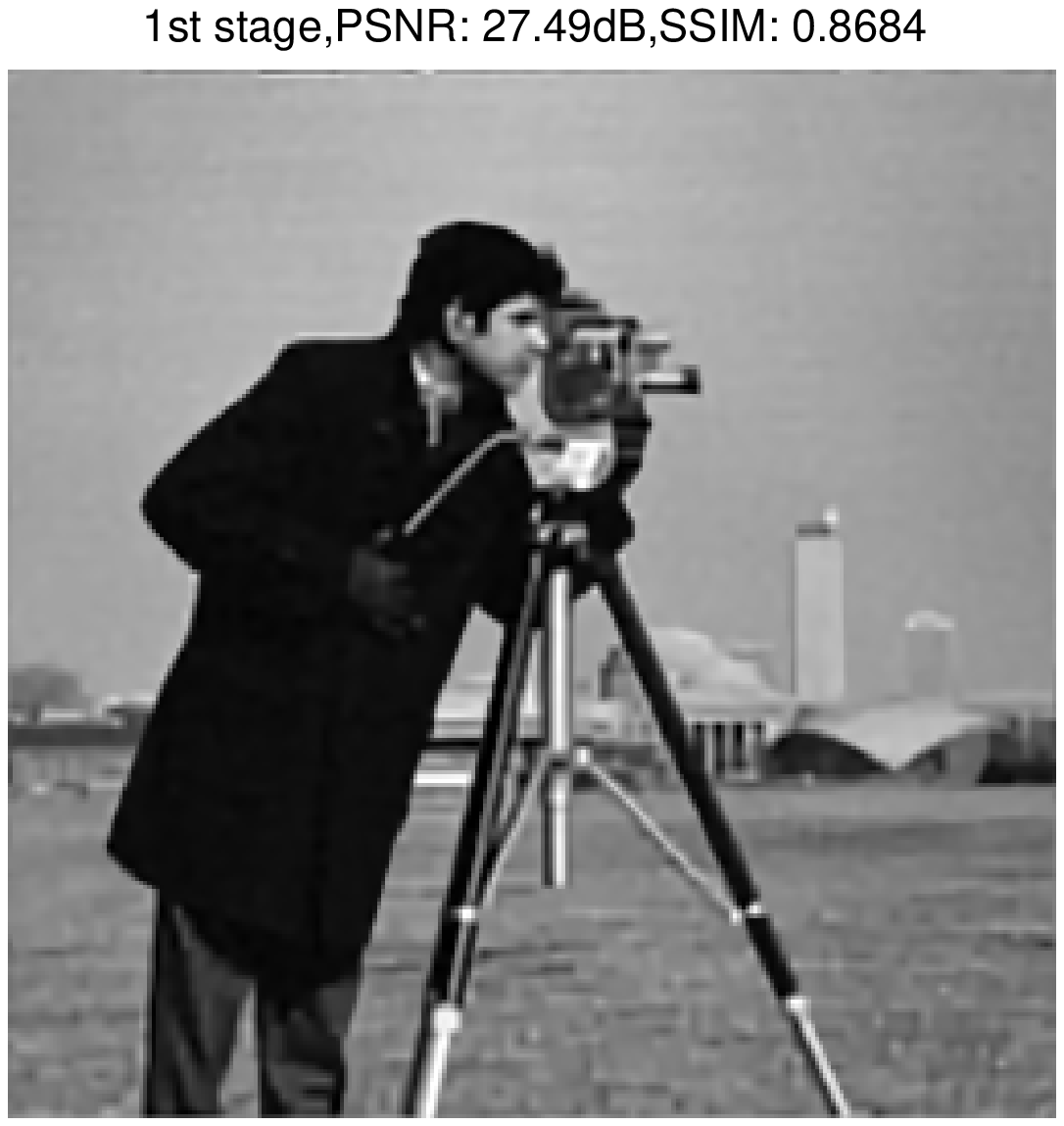}
   \includegraphics[width=0.22\textwidth]{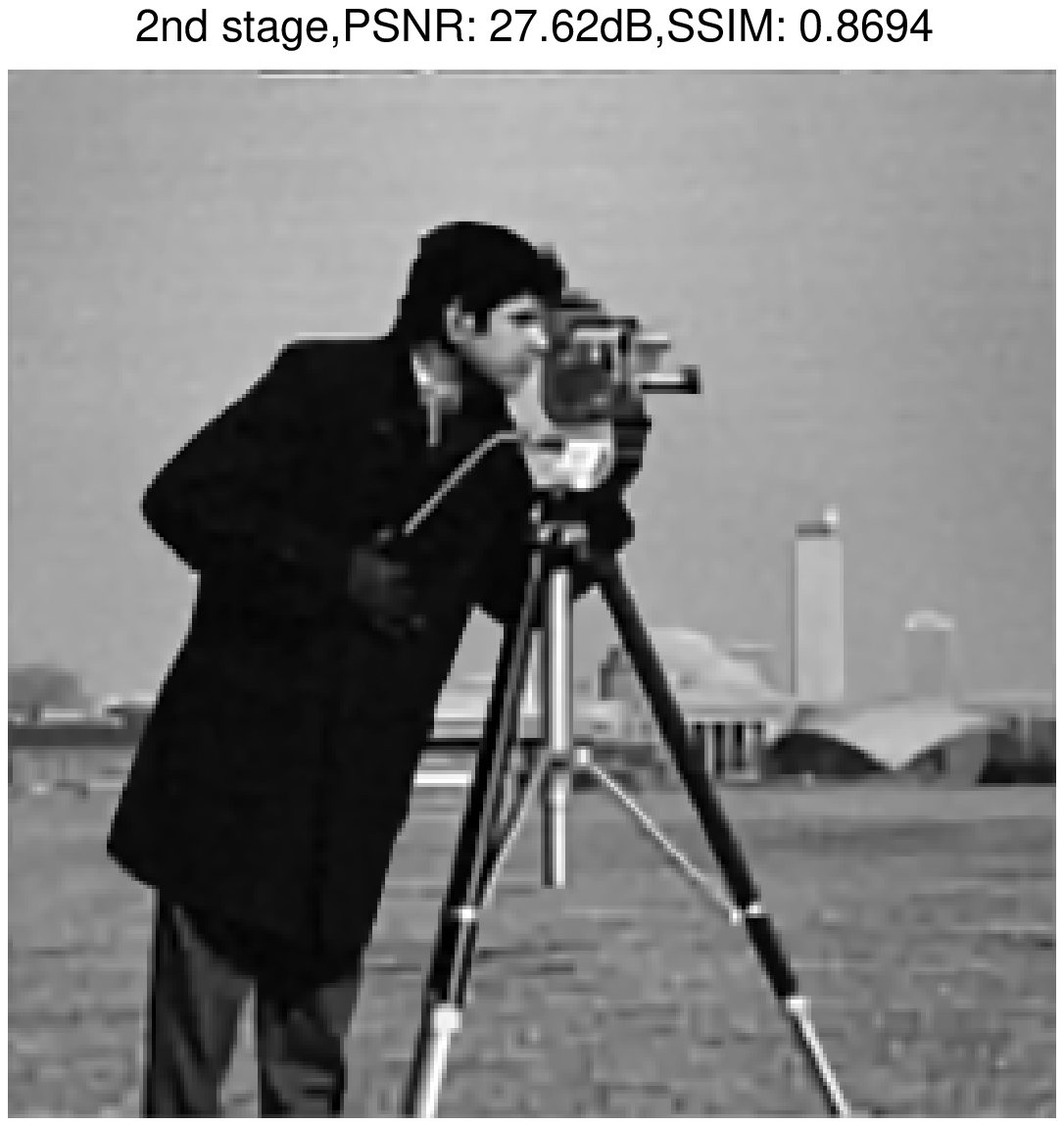}
   \includegraphics[width=0.22\textwidth]{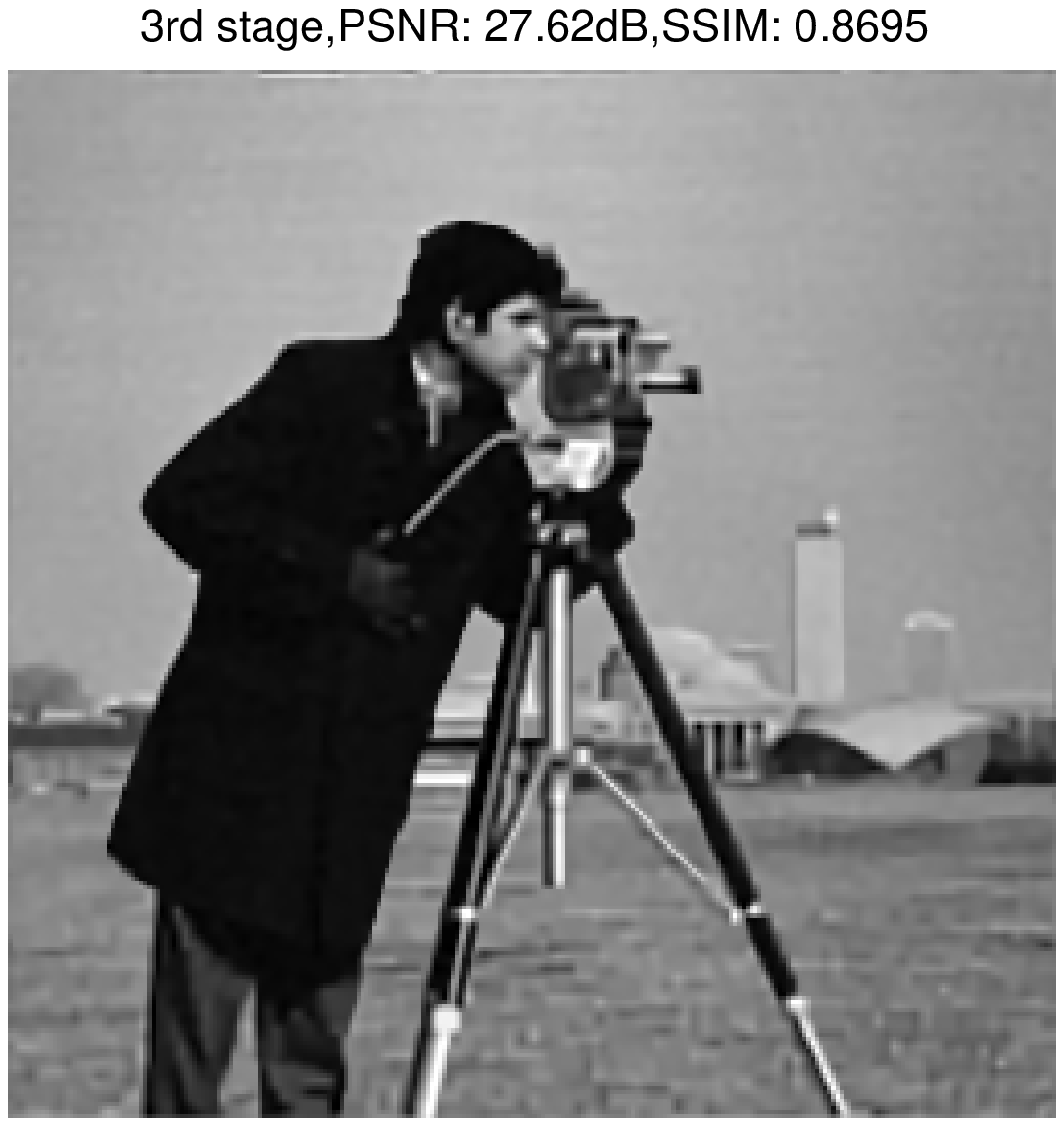}\\
   \includegraphics[width=0.22\textwidth]{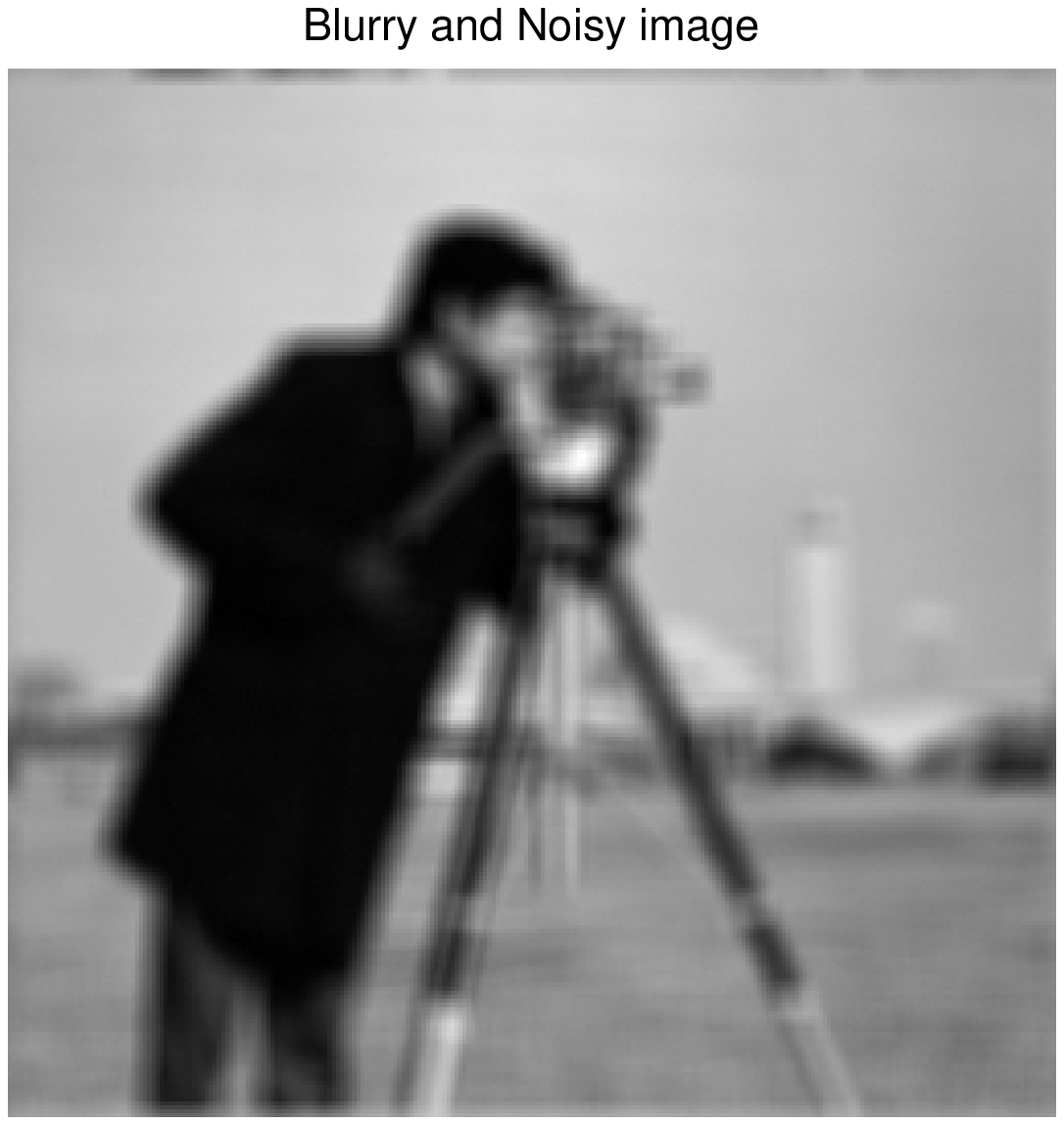}
   \includegraphics[width=0.22\textwidth]{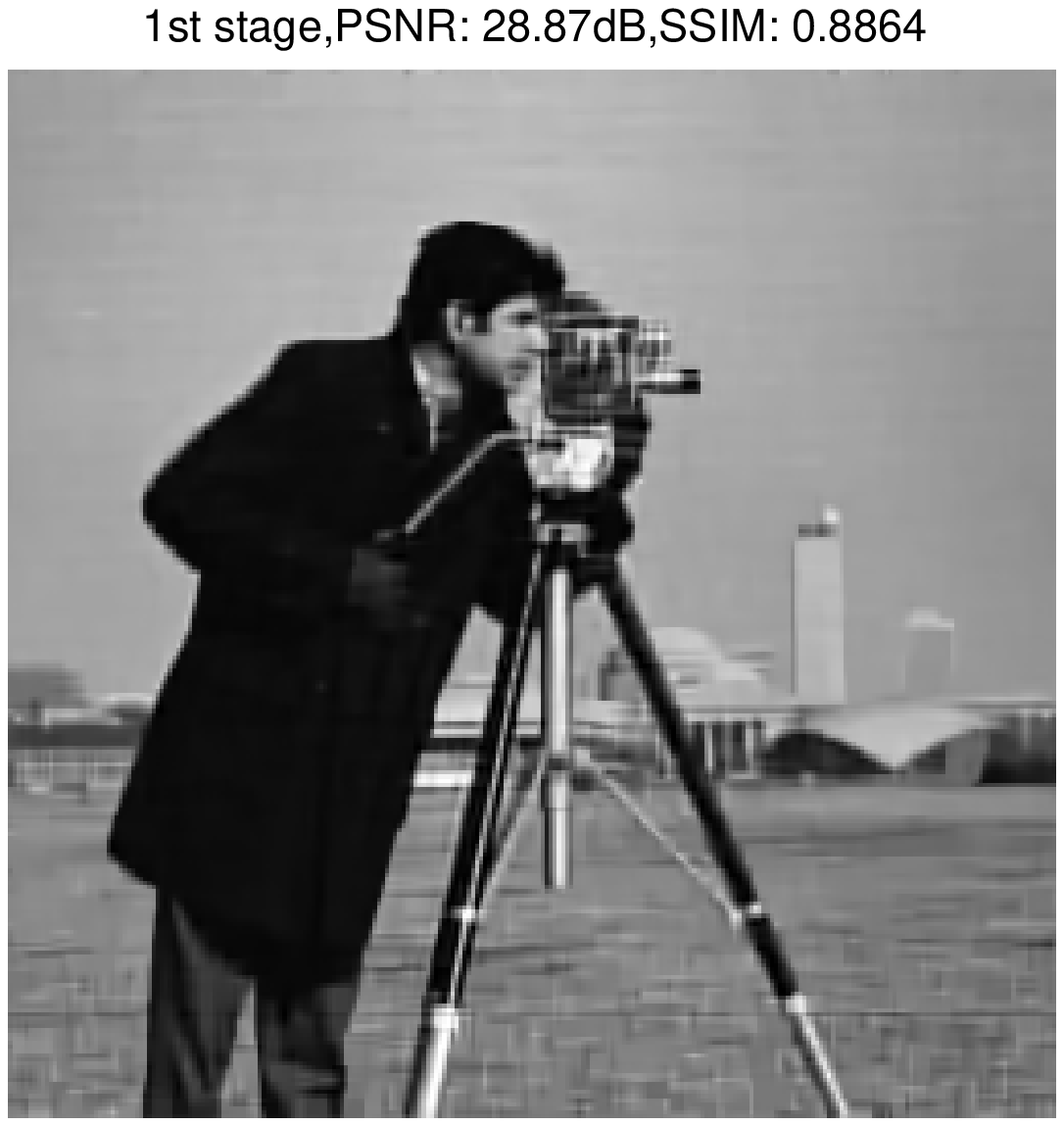}
   \includegraphics[width=0.22\textwidth]{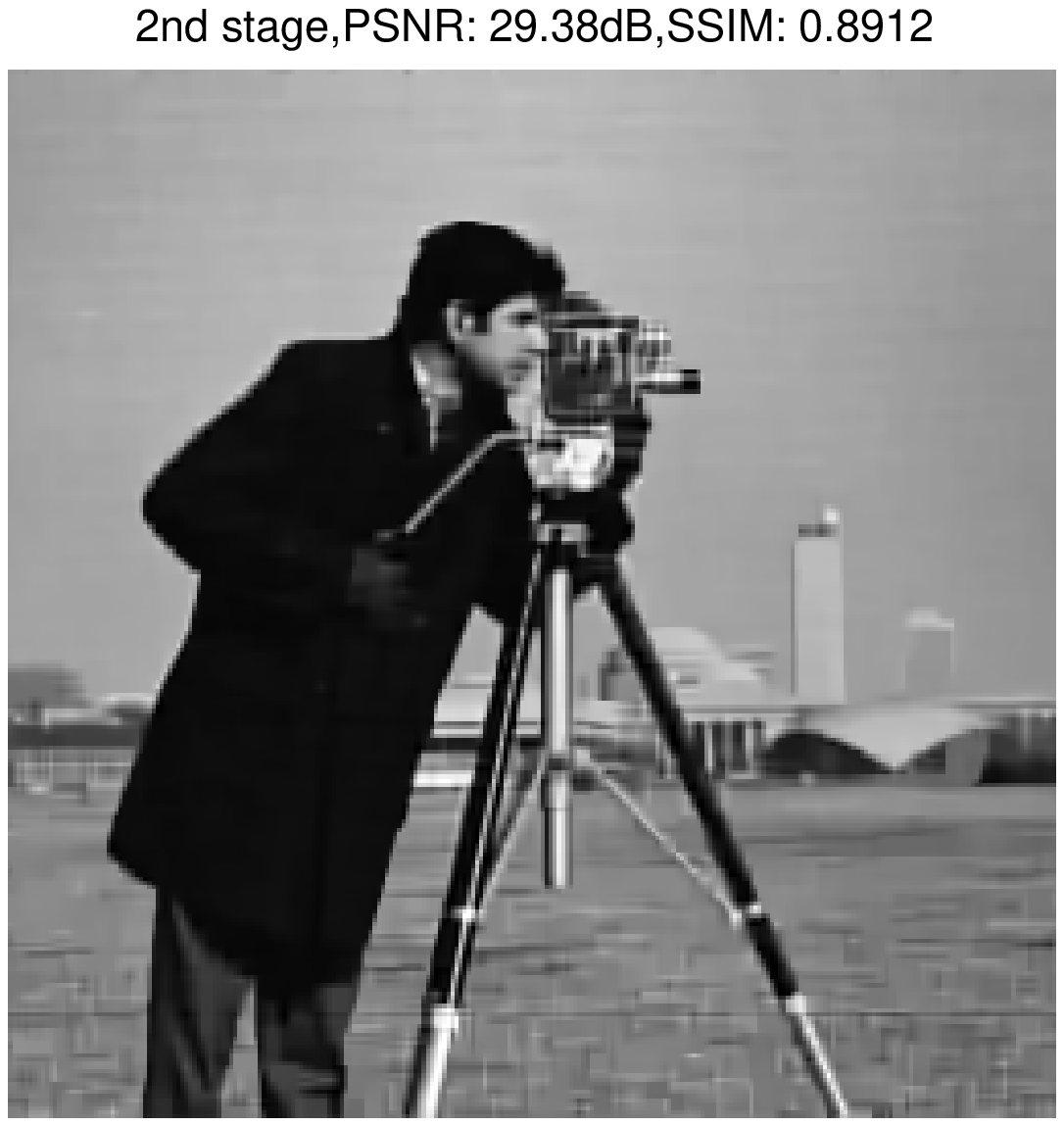}
   \includegraphics[width=0.22\textwidth]{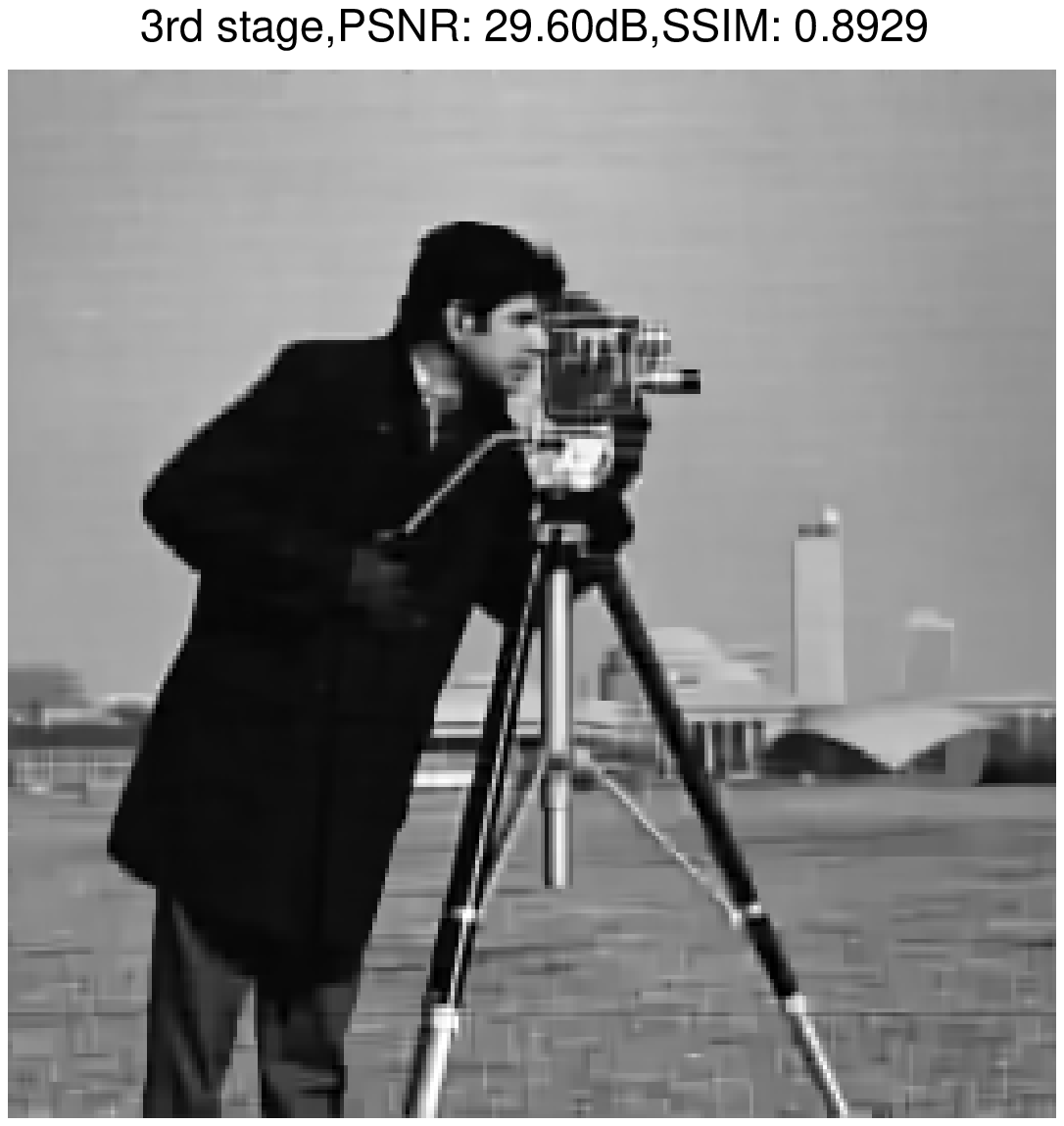}\\
   \includegraphics[width=0.22\textwidth]{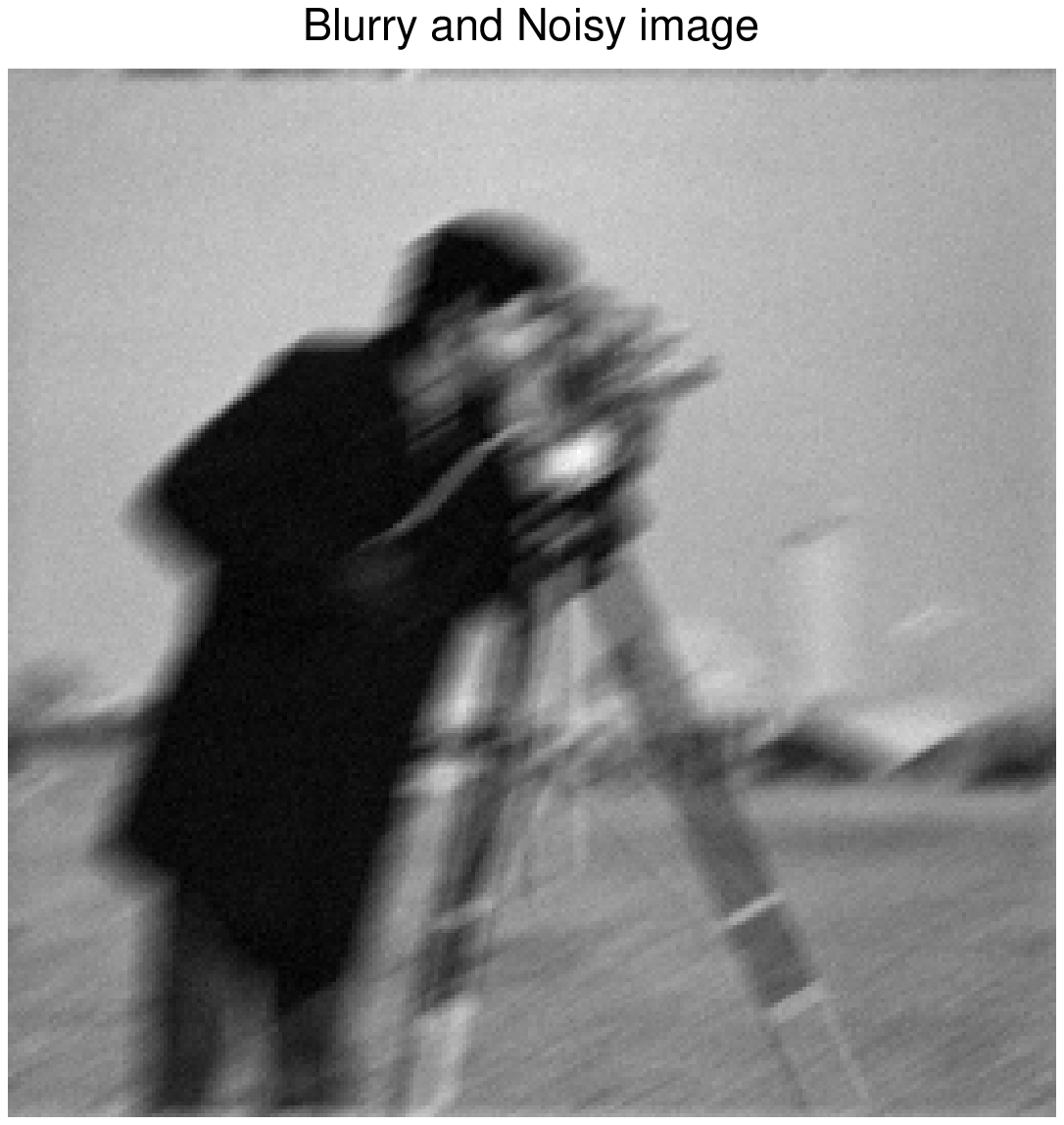}
   \includegraphics[width=0.22\textwidth]{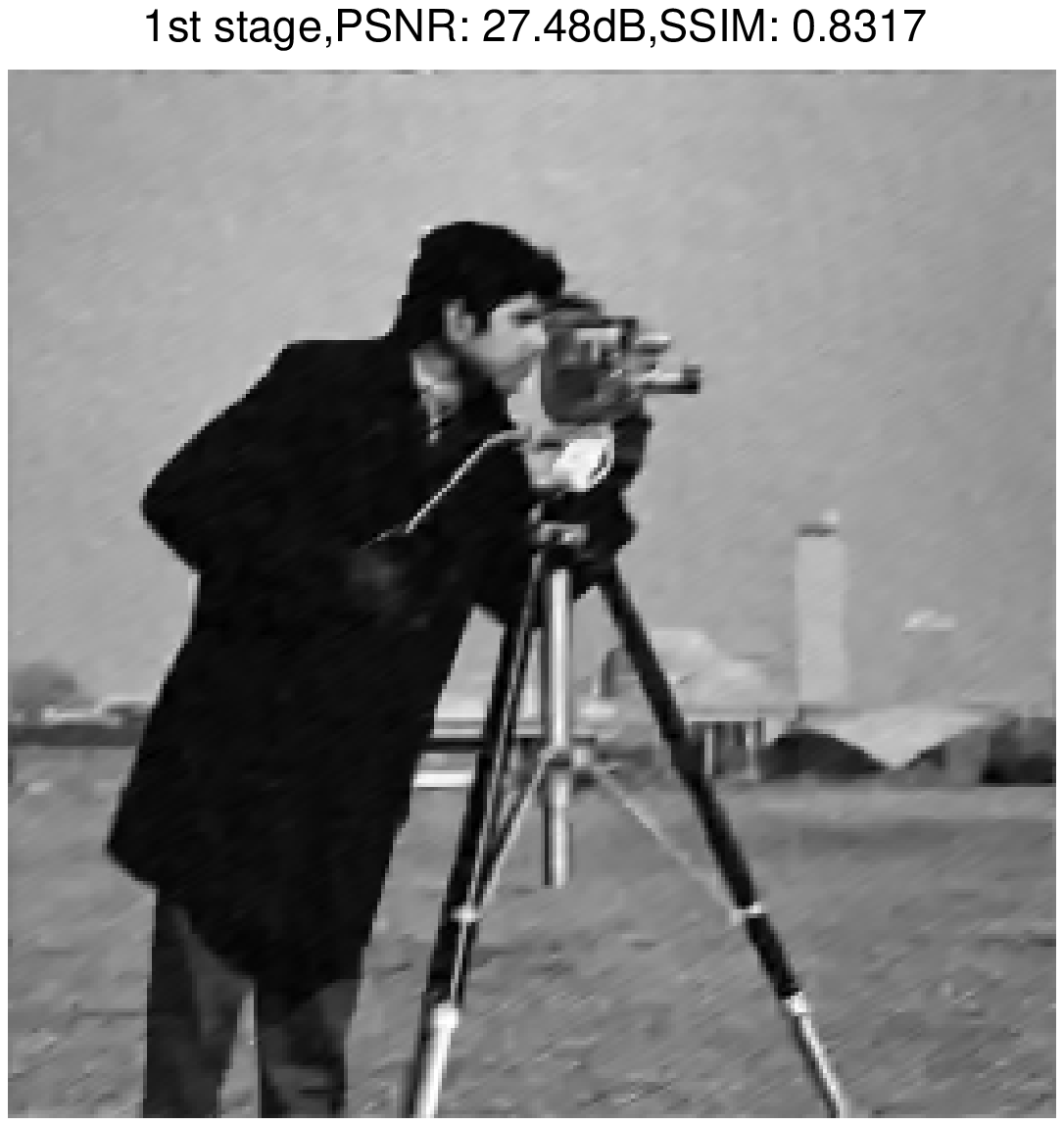}
   \includegraphics[width=0.22\textwidth]{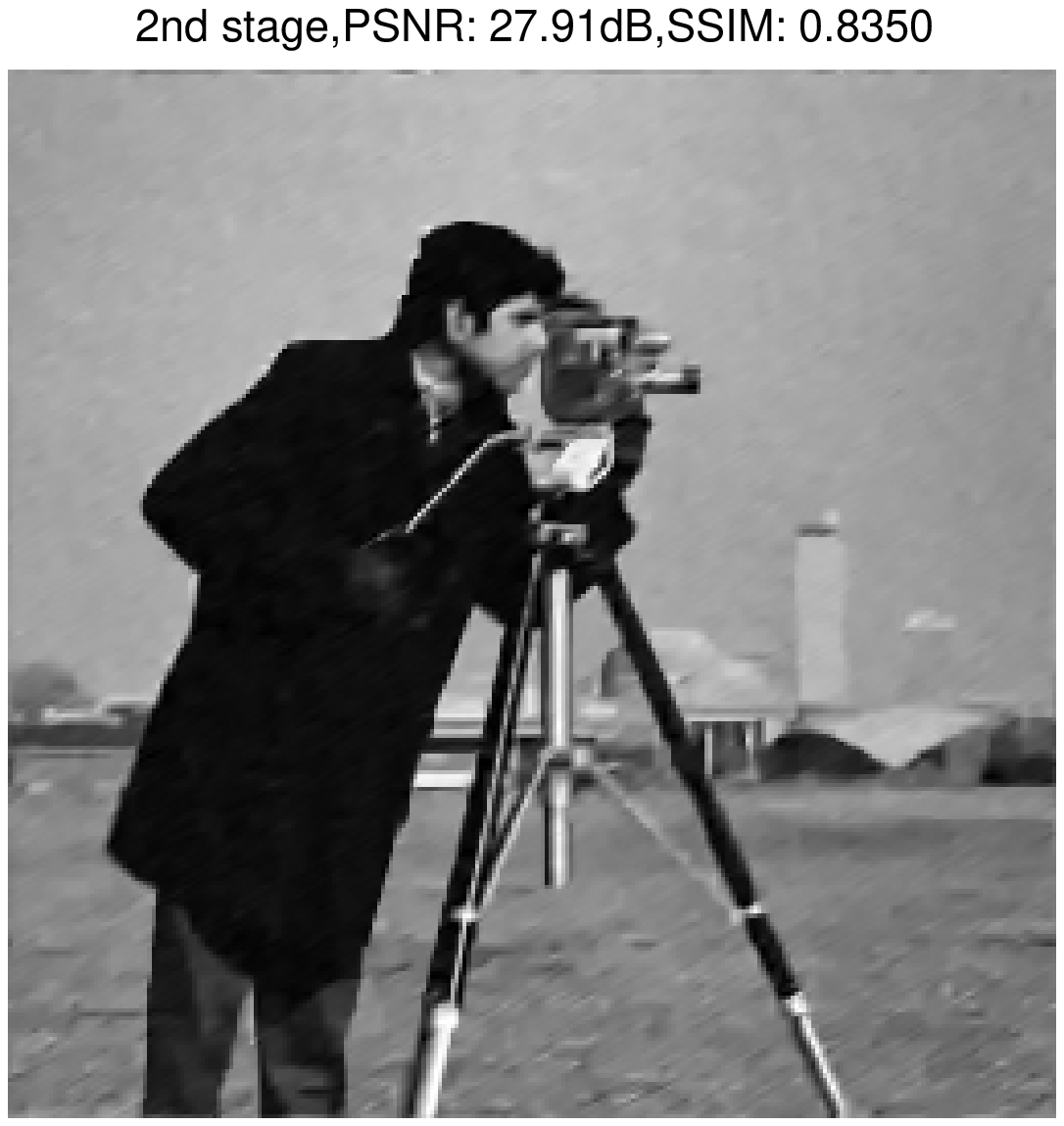}
   \includegraphics[width=0.22\textwidth]{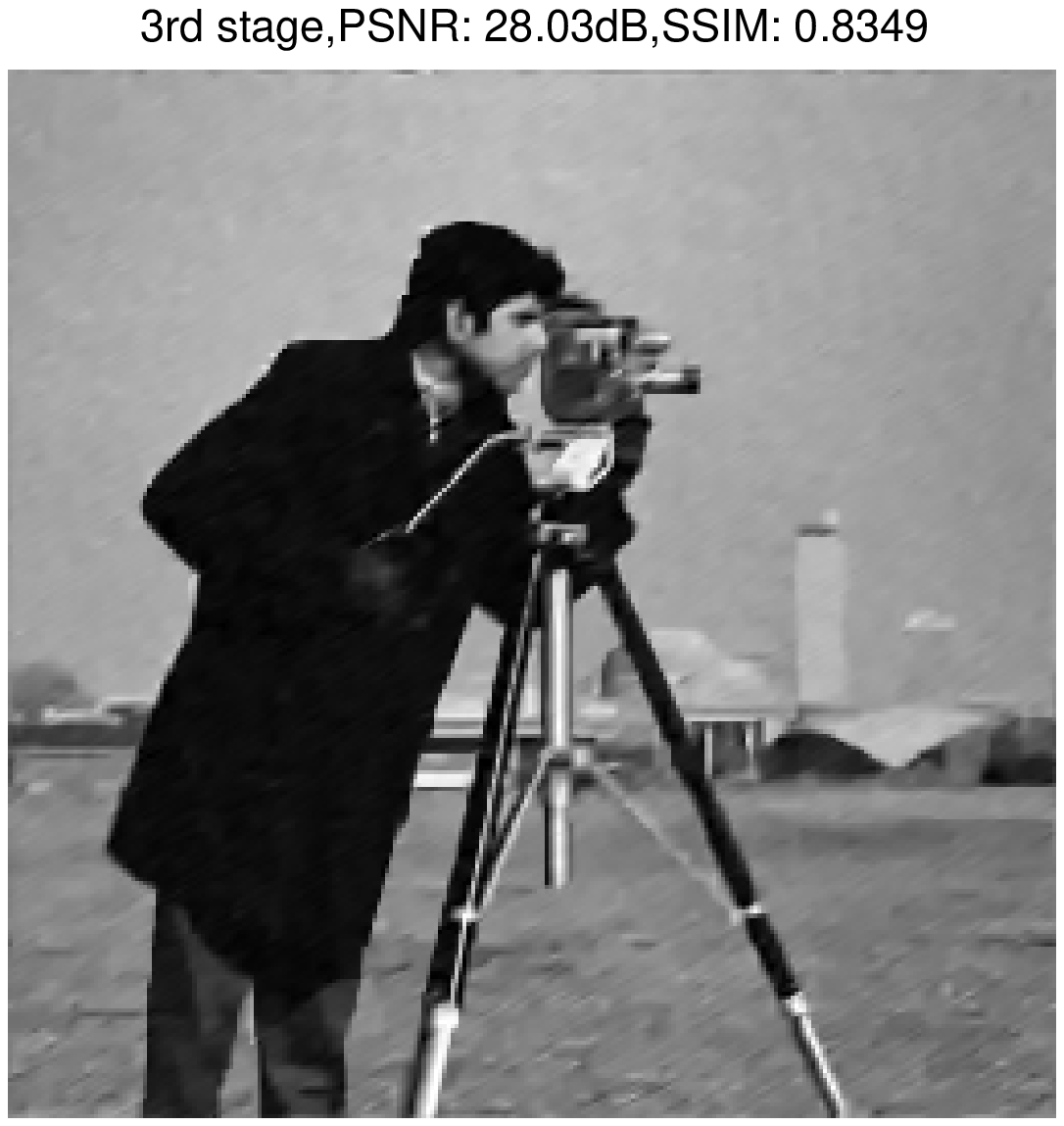}\\
   \includegraphics[width=0.22\textwidth]{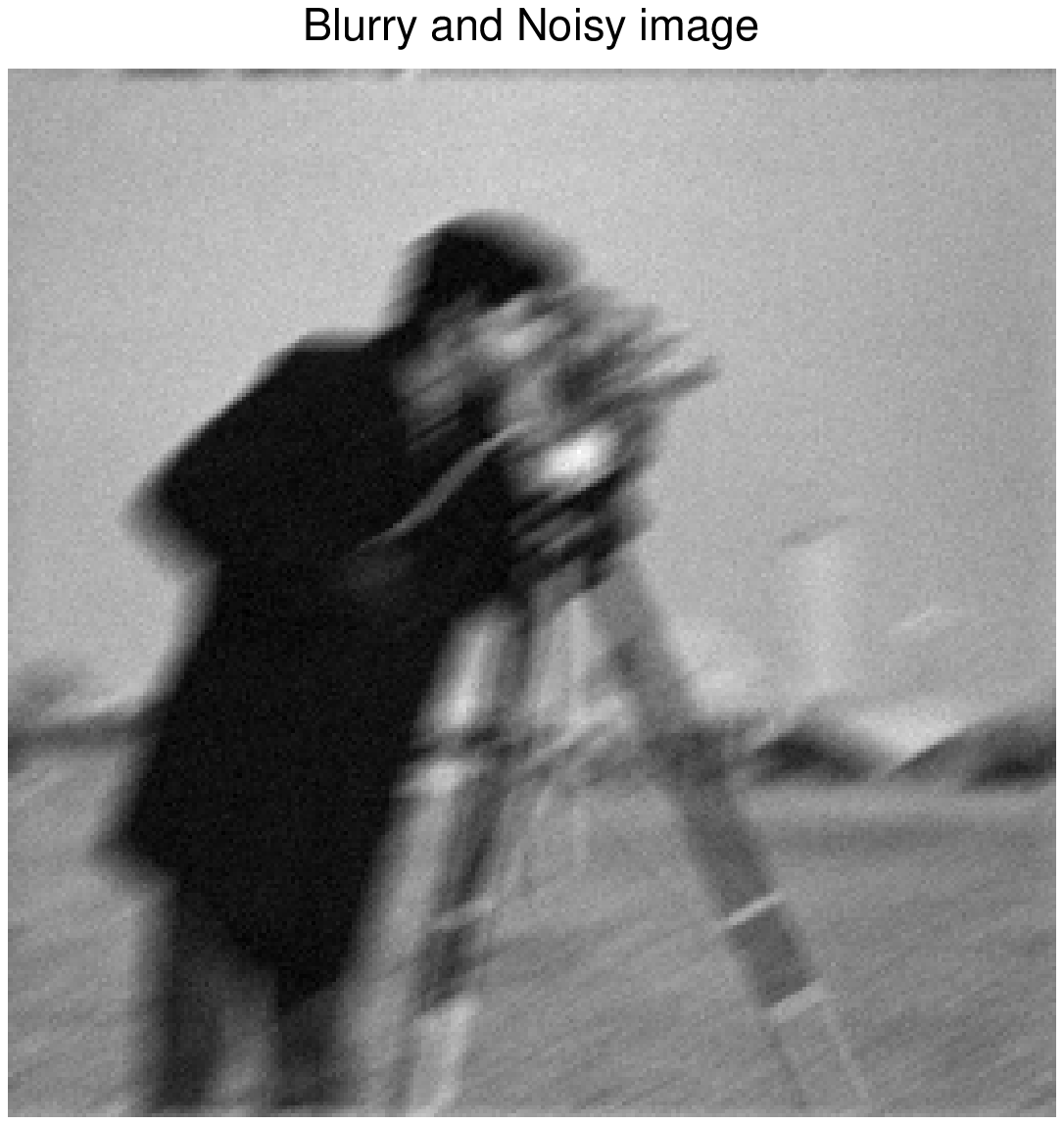}
   \includegraphics[width=0.22\textwidth]{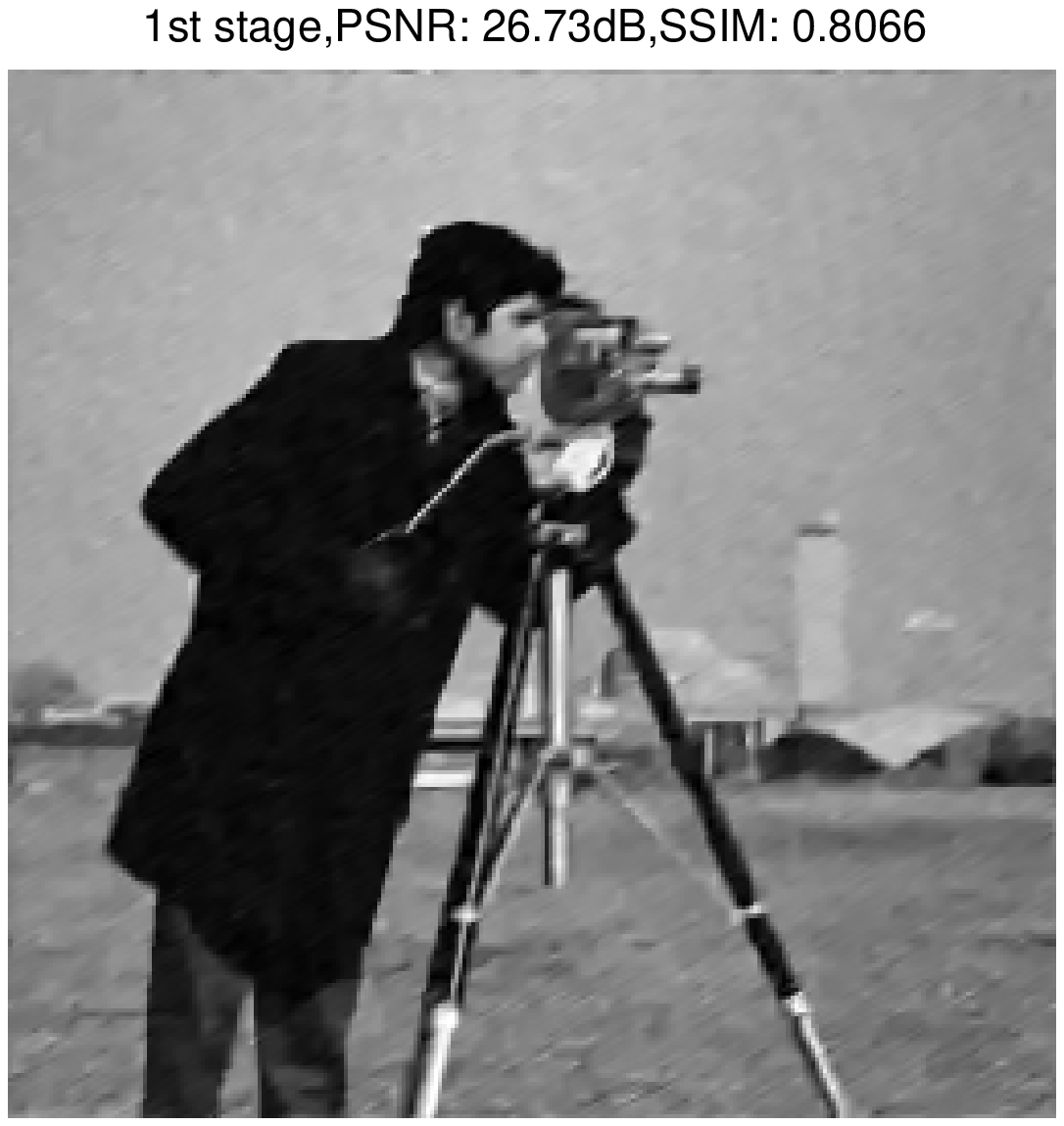}
   \includegraphics[width=0.22\textwidth]{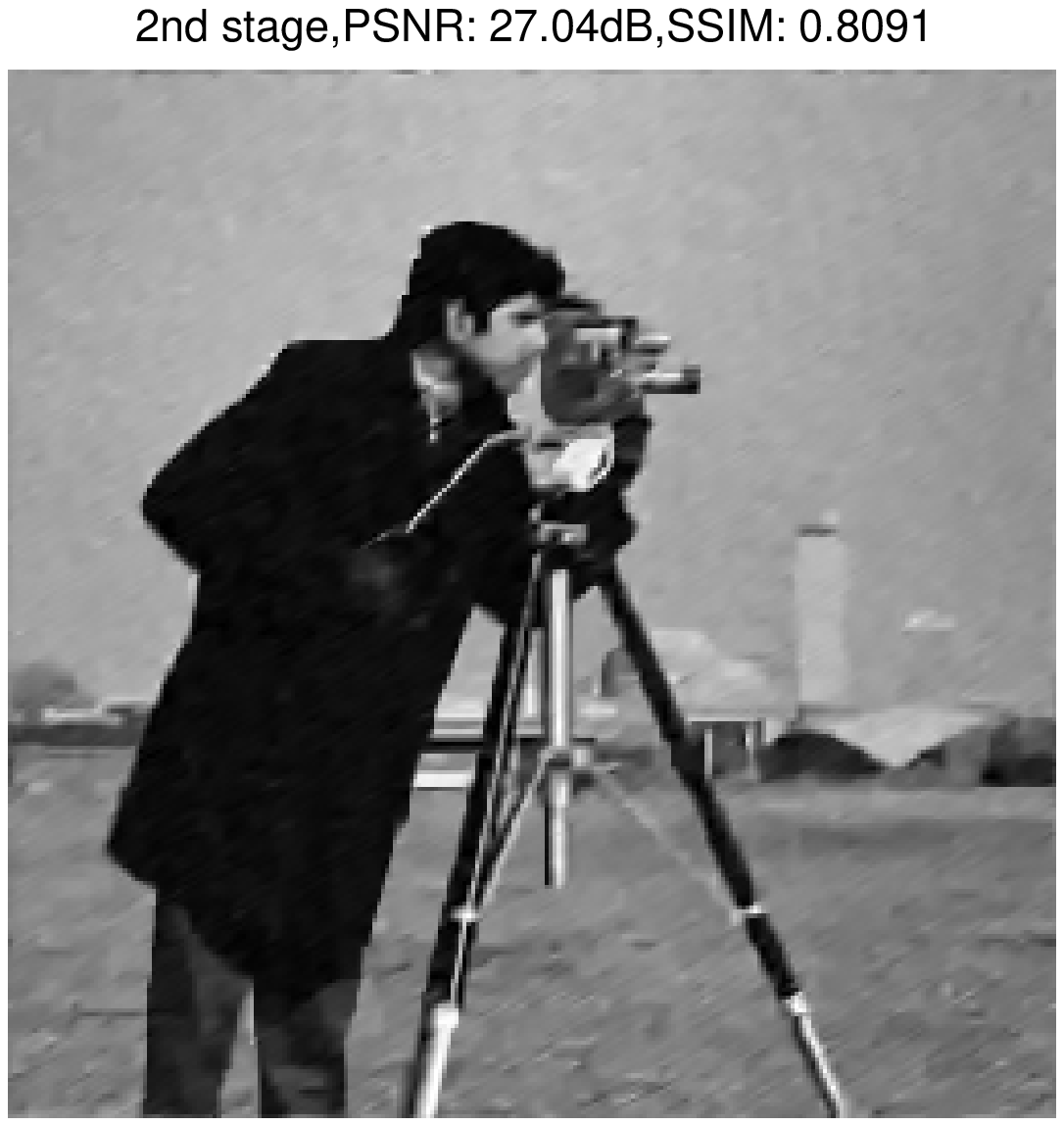}
   \includegraphics[width=0.22\textwidth]{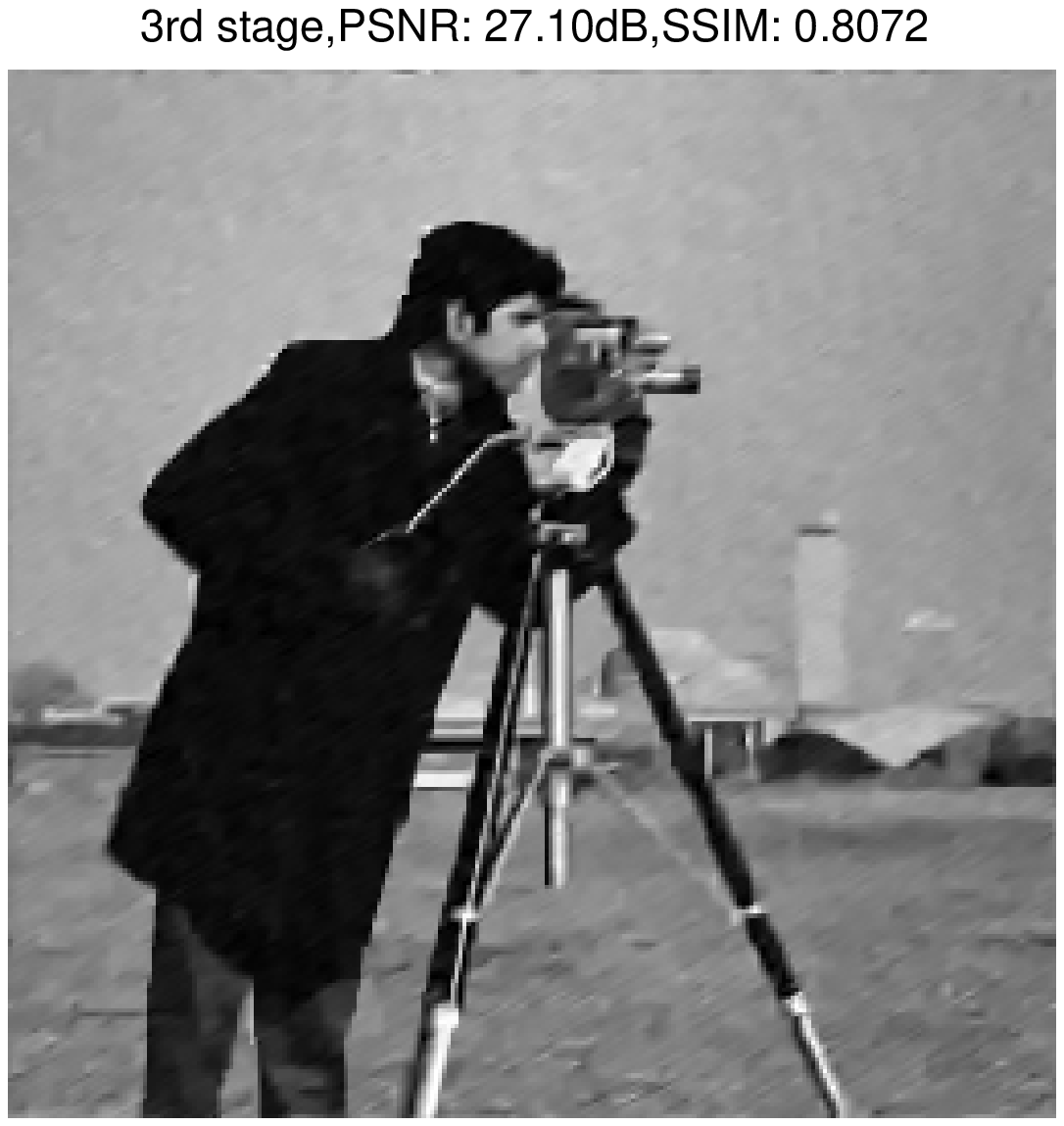}\\
  \caption{\small The intermediate stage results of proposed algorithm. Where the first to fourth columns are corresponding to the blurry and noisy image, the results of first stage, second stage and third stage, respectively. }
\label{fig: intermediate results}
\end{figure}

\subsection{Results and discussions}
In this subsection, we report the experiments results, comparing the proposed algorithm with the PBOS algorithm for the nonlocal TV model, the Split Bregman method for the wavelet frame based $l_1$ minimization model, MDAL method for the $l_0$ model, Nonlocal MDAL method for the $l_0$-$l_2$ model, and the famous IDD-BM3D algorithm.

First of all, in Table 1, we summarize the results of the five algorithms for all the test images with different blur kernels and noise levels. Note that the PBOS algorithm for the nonlocal TV model can not satisfy the stopping criterion based on the error between two successive iteration values such as $||u^{k+1}-u^k||_2/||u^k||_2 < 5\times 10^{-4}$ for many iterations, hence the stopping criterion of $||Ku^k - f||_2< \sigma$ adopted  \cite{Zhangxiaoqun2010} is also used here, i.e., the iterative process stops if either of the two criterions is satisfied. We can observe that our proposed algorithm has an overall better performance than the other methods in terms of both PSNR and SSIM values.

\begin{figure}[h]
  \centering
   \includegraphics[width=0.22\textwidth]{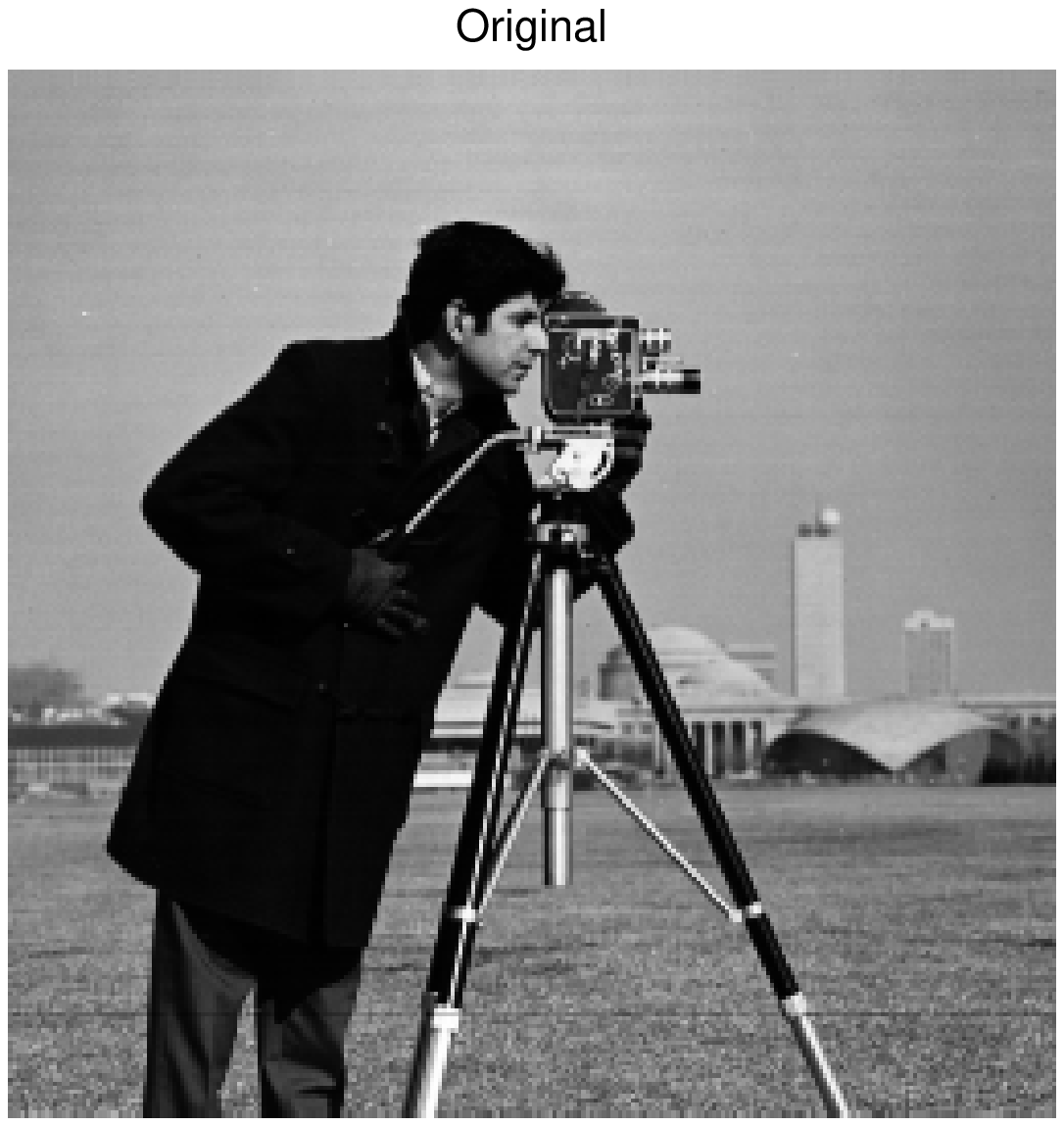}
   \includegraphics[width=0.22\textwidth]{cameraman_corrupted_motion154.eps}
    \includegraphics[width=0.22\textwidth]{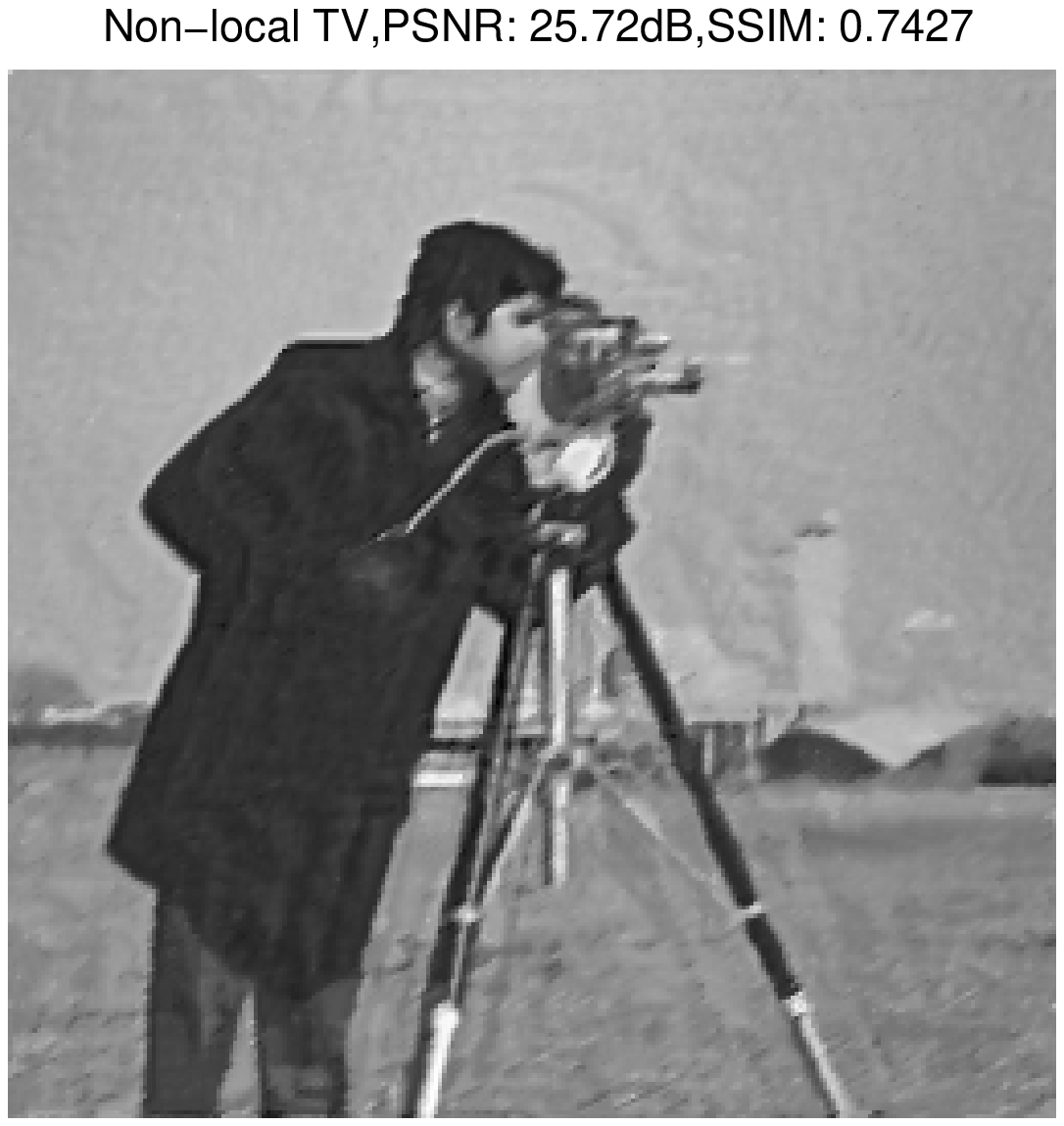}
   \includegraphics[width=0.22\textwidth]{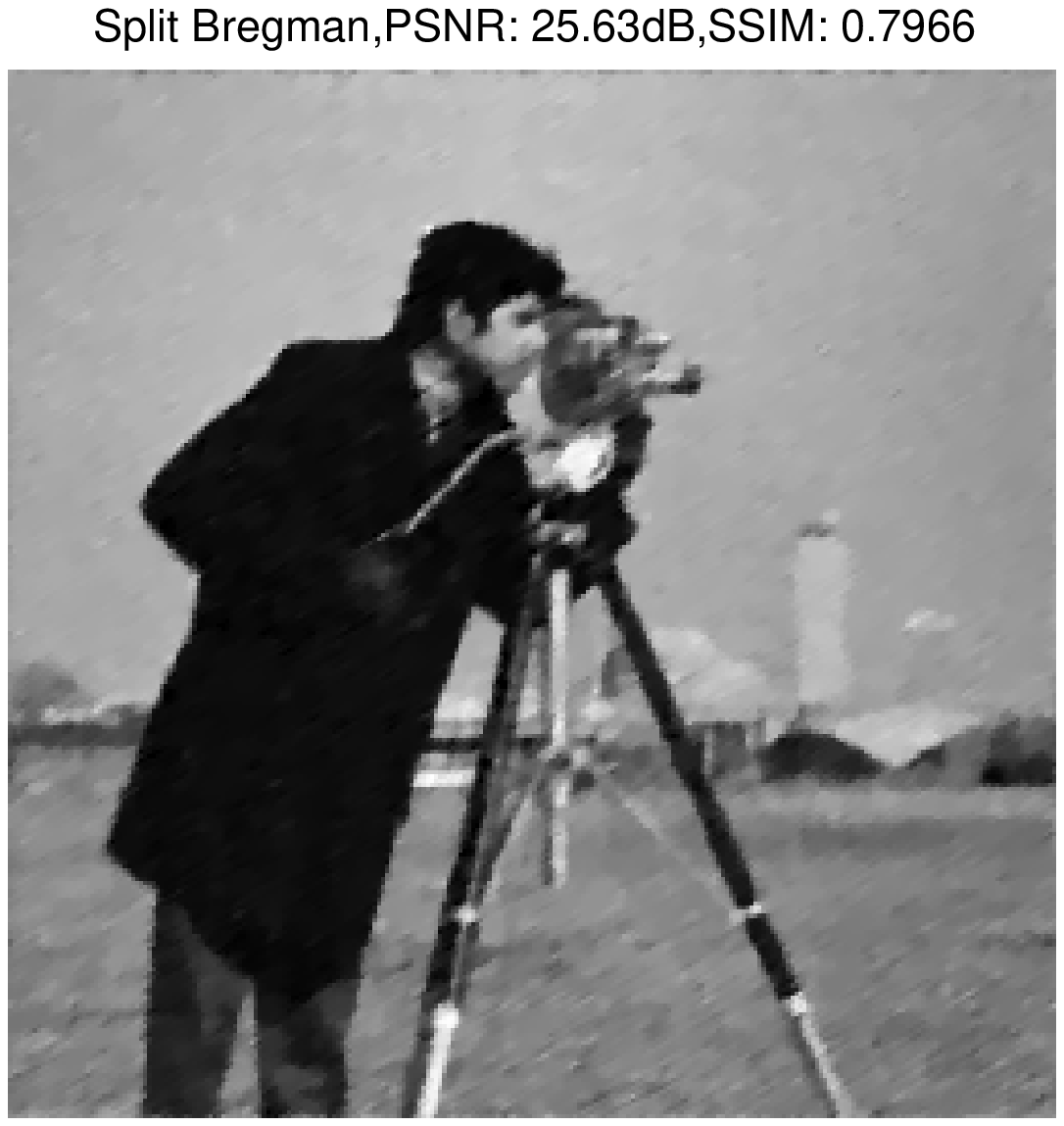} \\
   \includegraphics[width=0.22\textwidth]{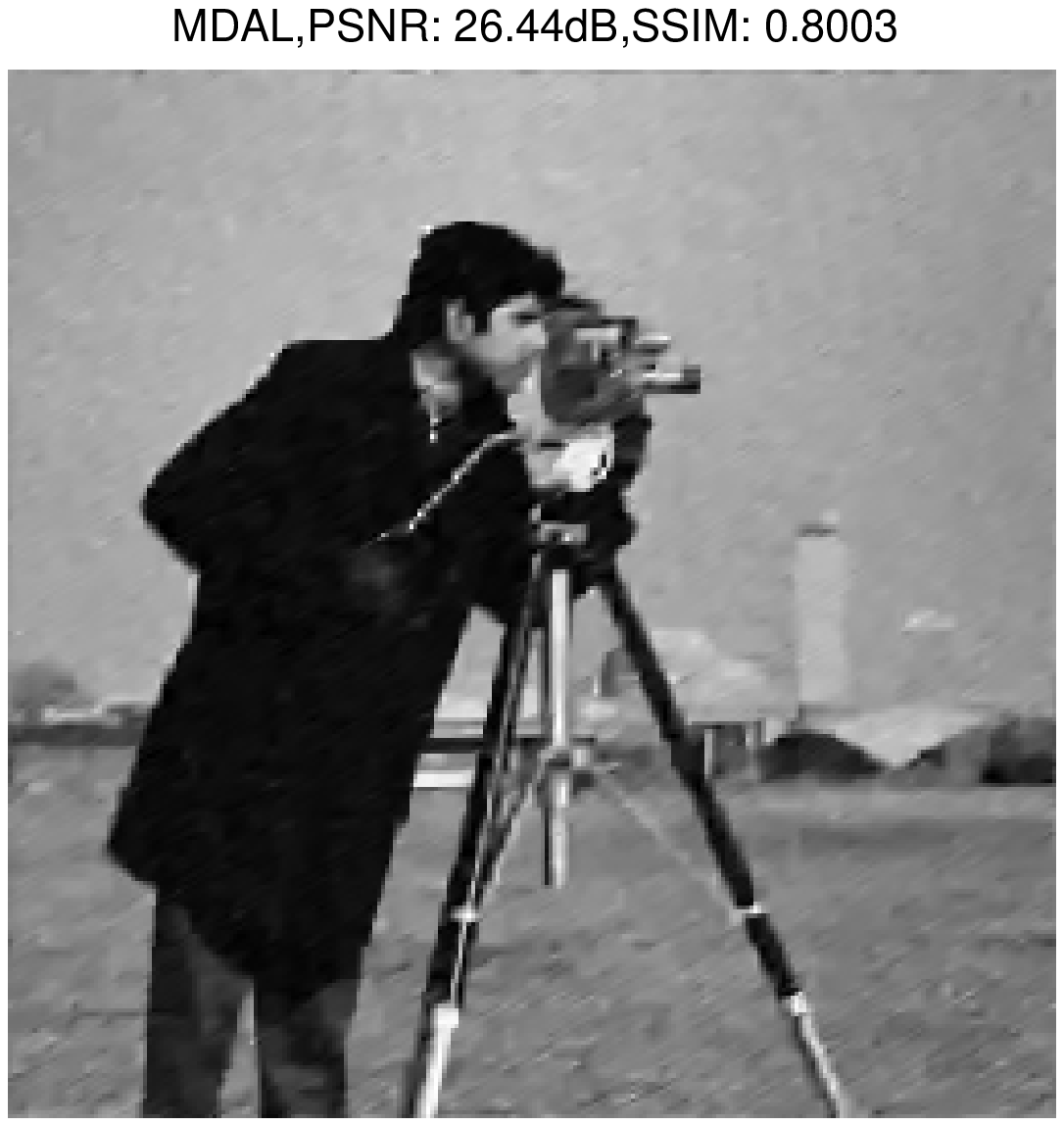}
   \includegraphics[width=0.22\textwidth]{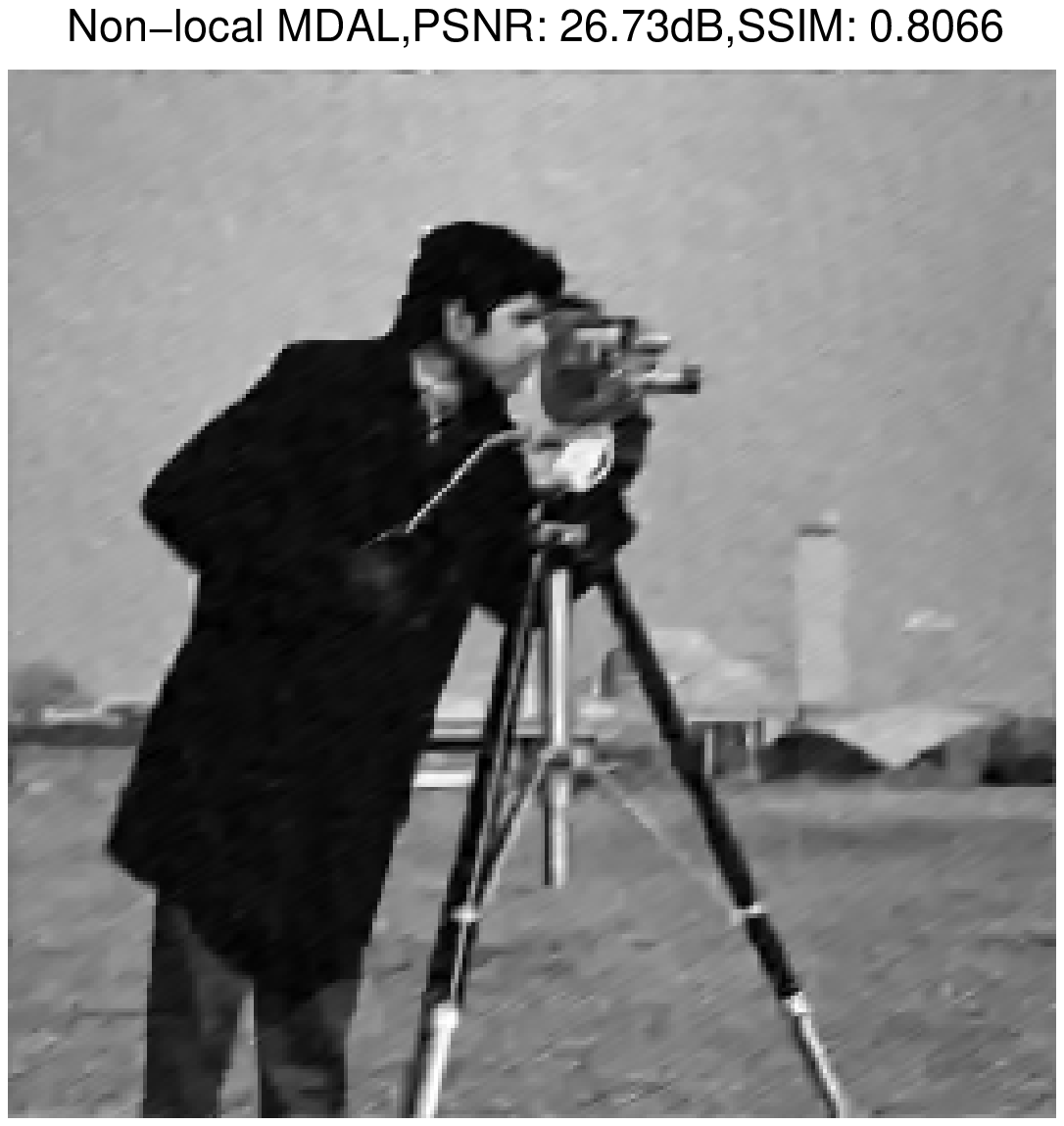}
   \includegraphics[width=0.22\textwidth]{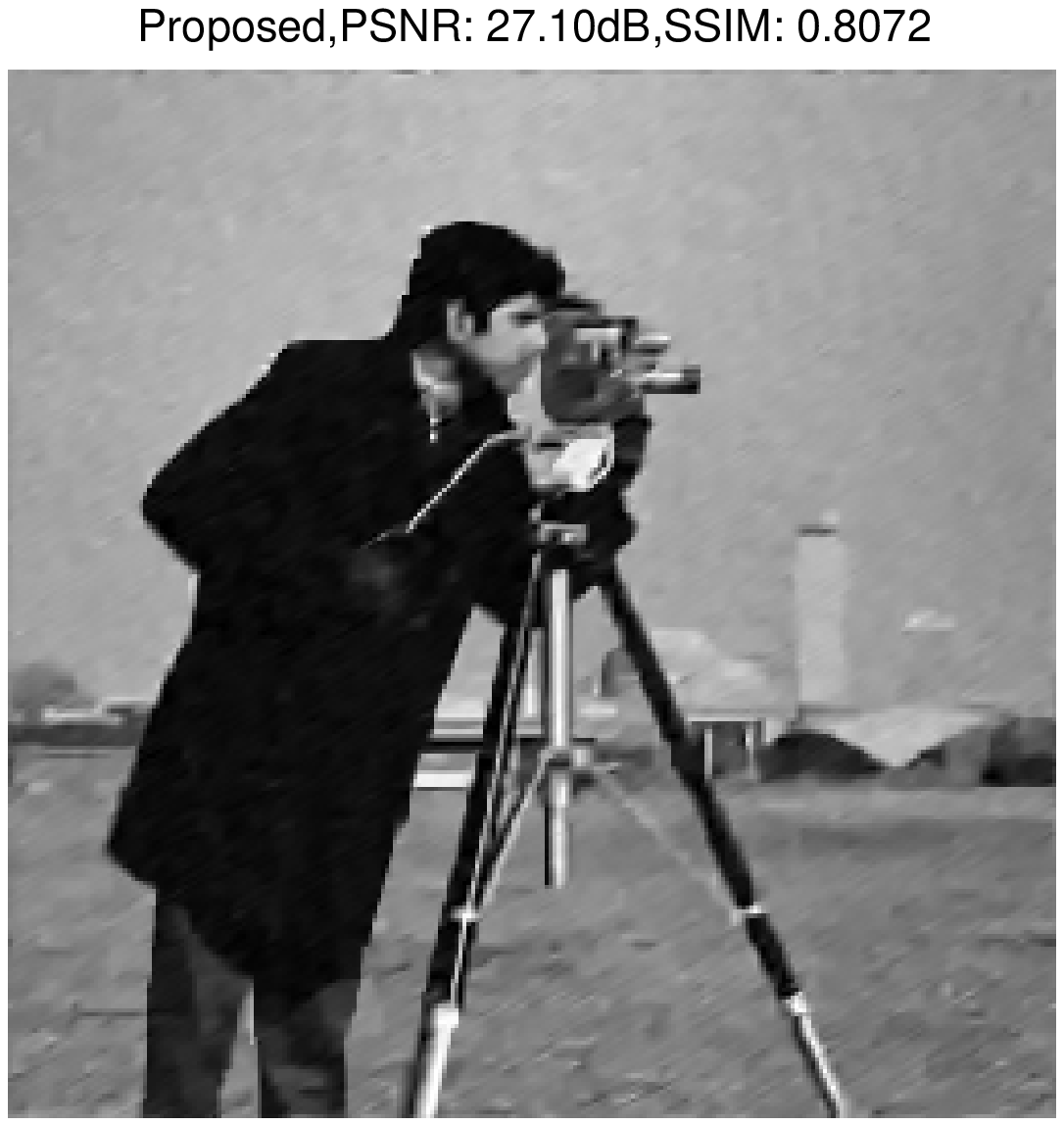}
   \includegraphics[width=0.22\textwidth]{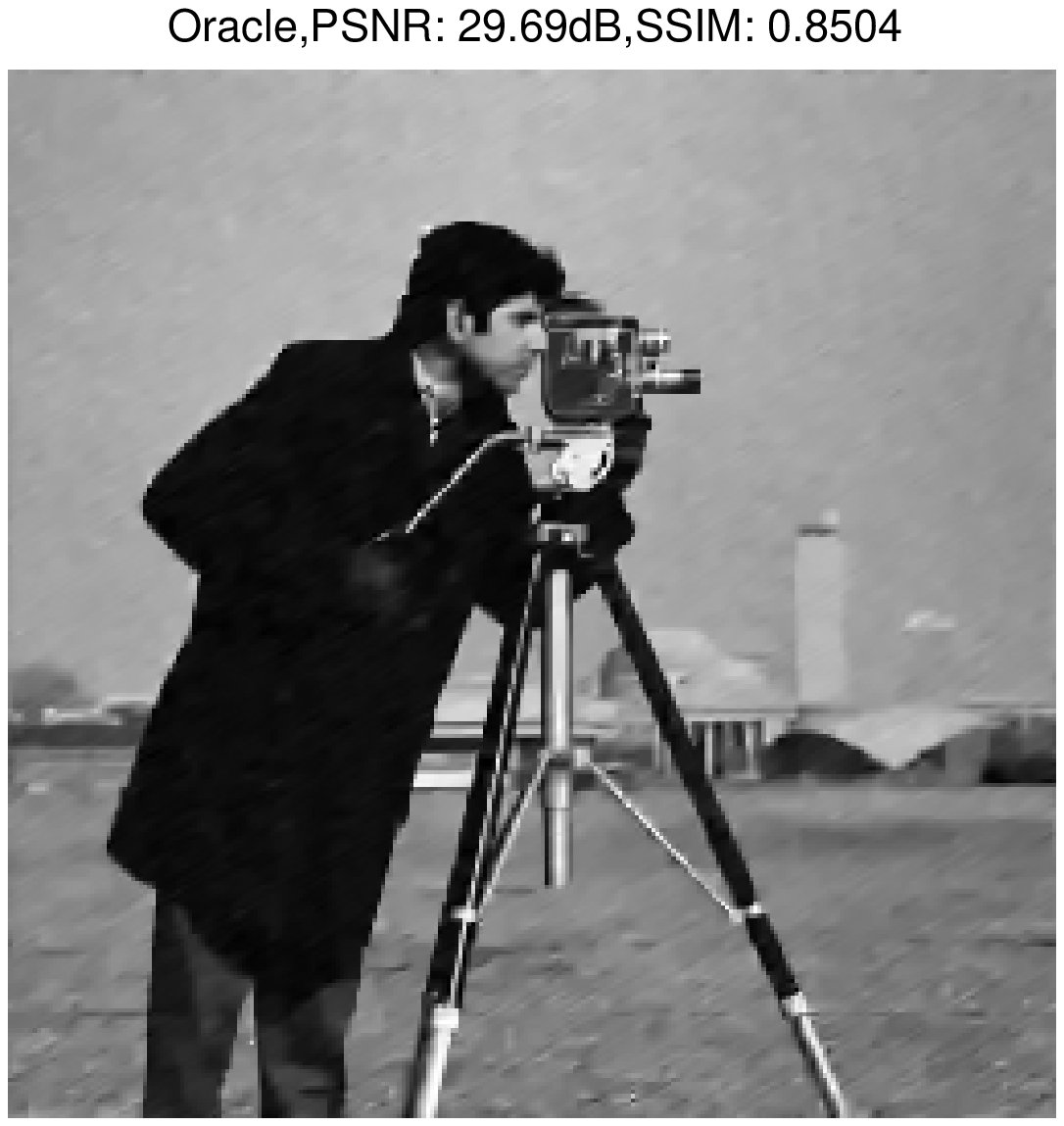} \\
      \includegraphics[width=0.22\textwidth]{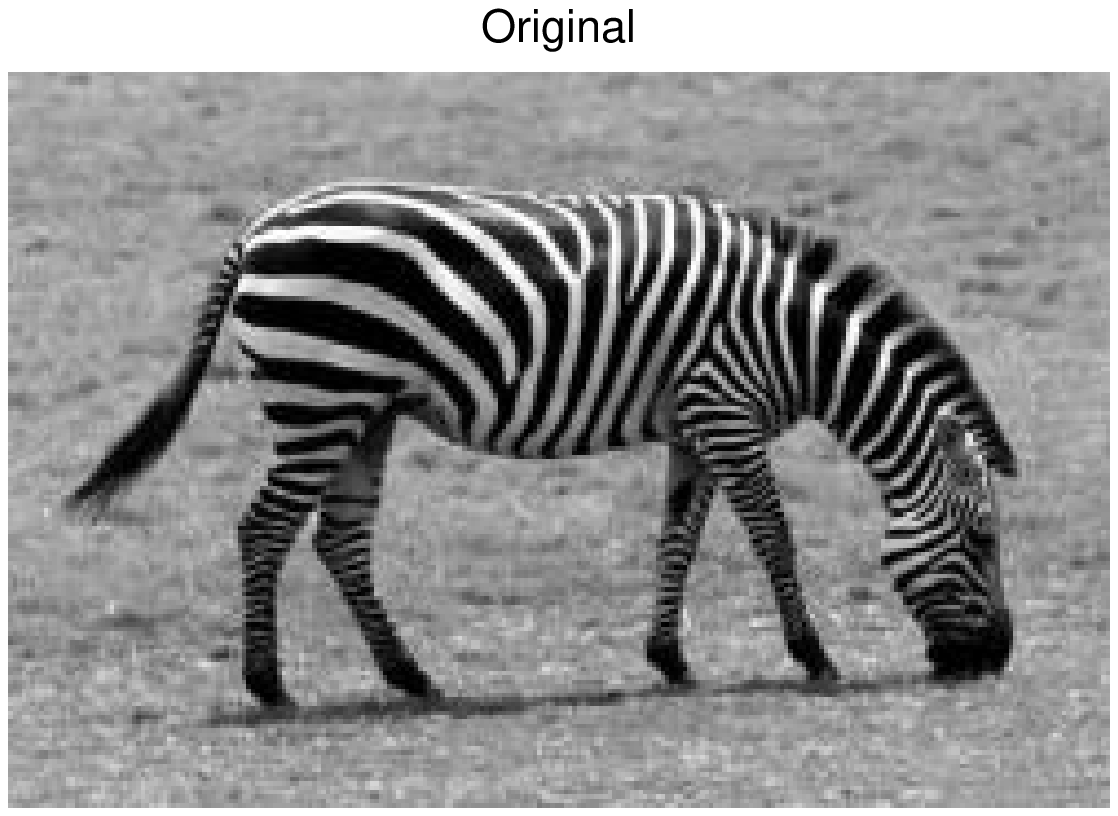}
   \includegraphics[width=0.22\textwidth]{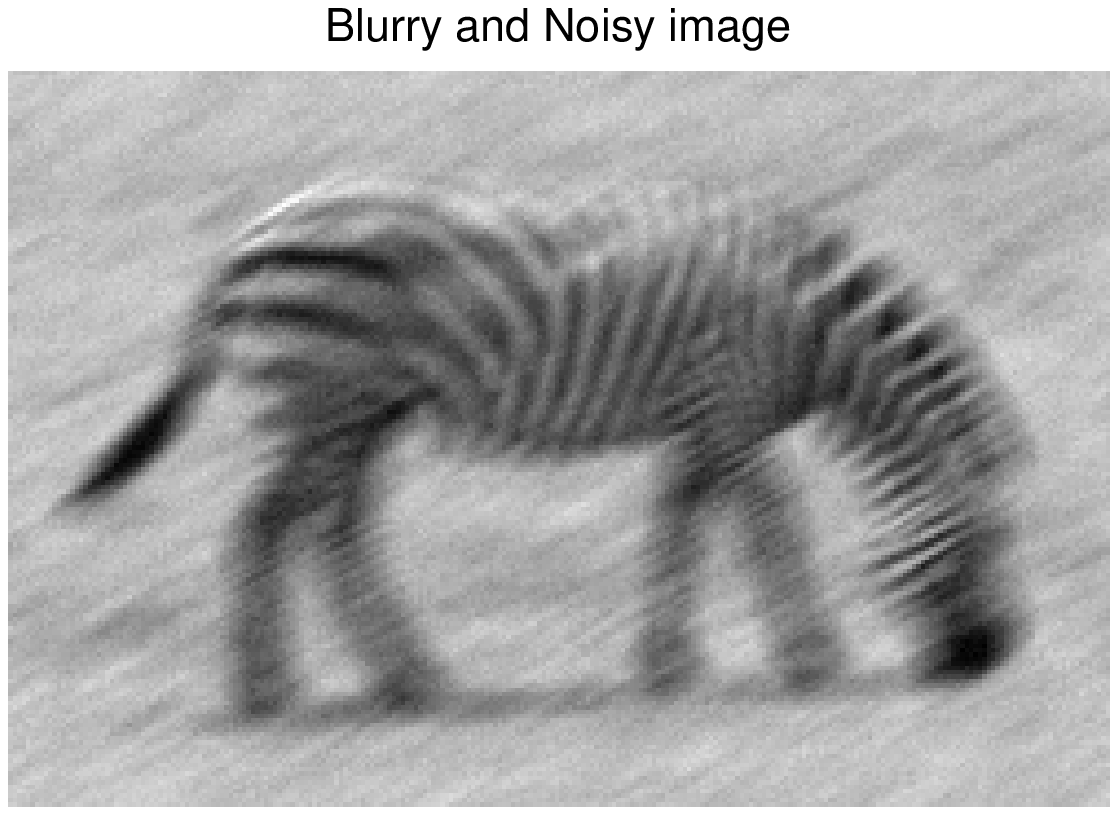}
    \includegraphics[width=0.22\textwidth]{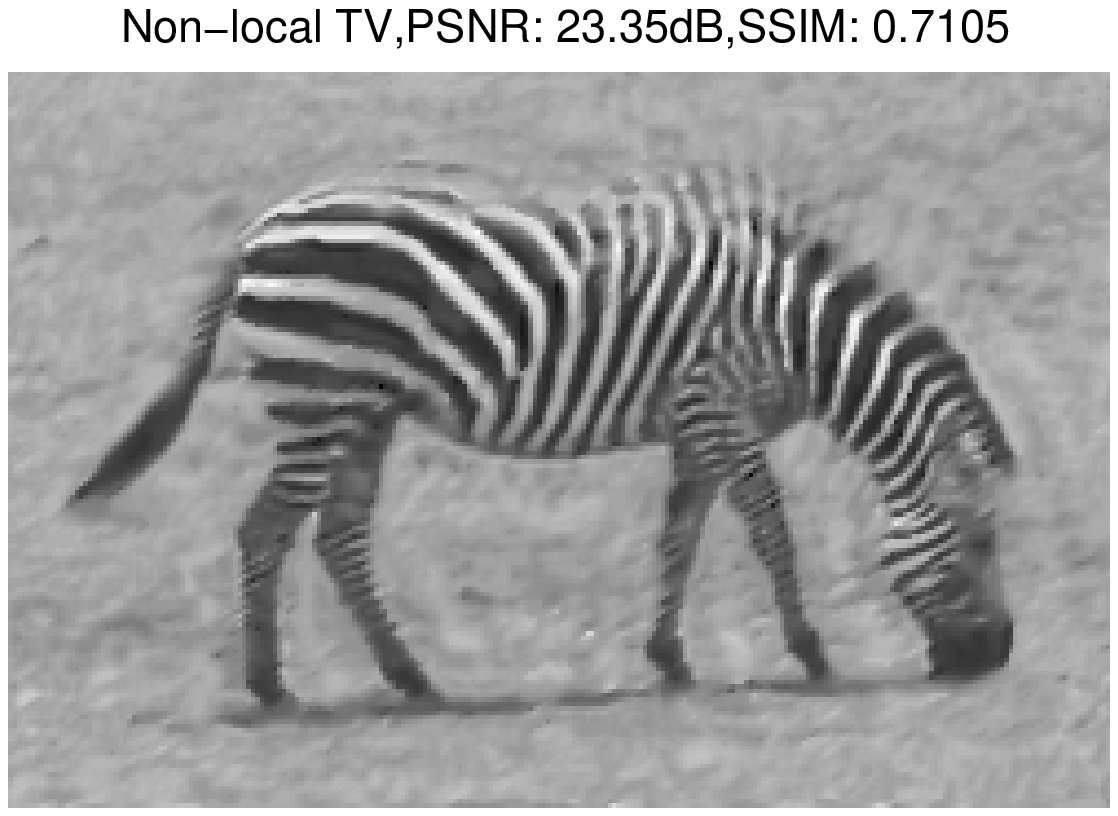}
   \includegraphics[width=0.22\textwidth]{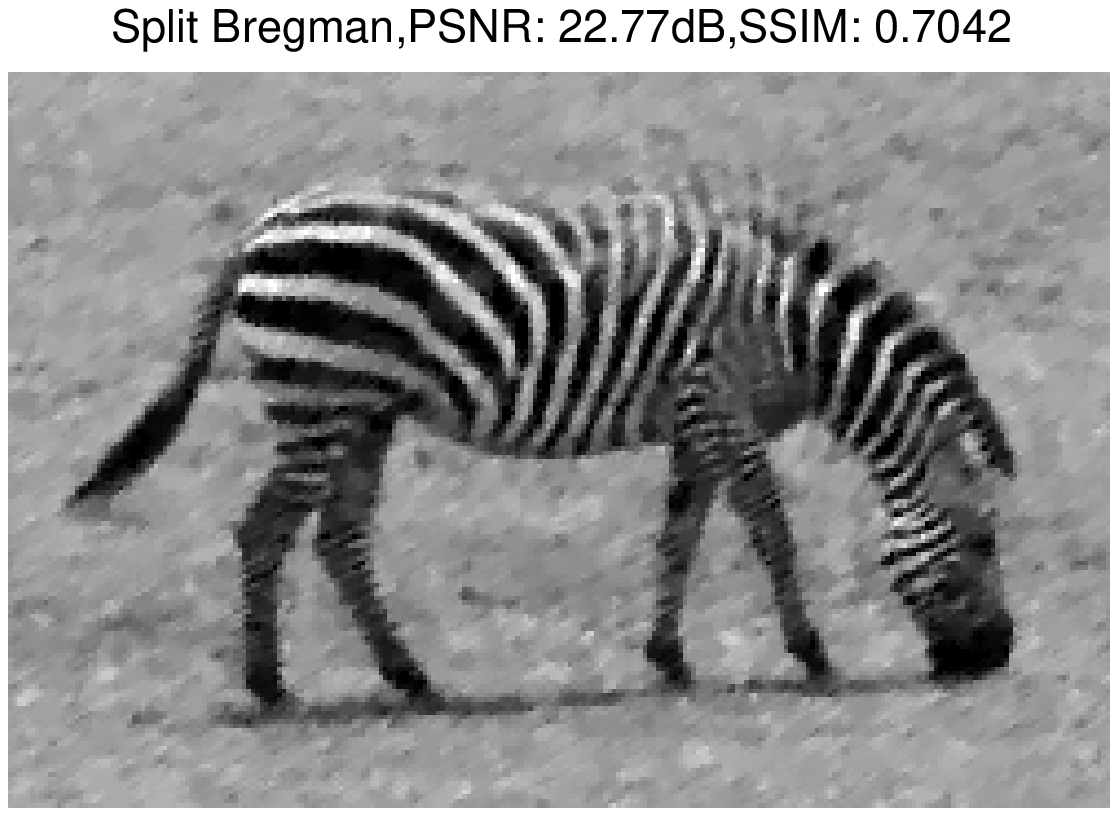} \\
   \includegraphics[width=0.22\textwidth]{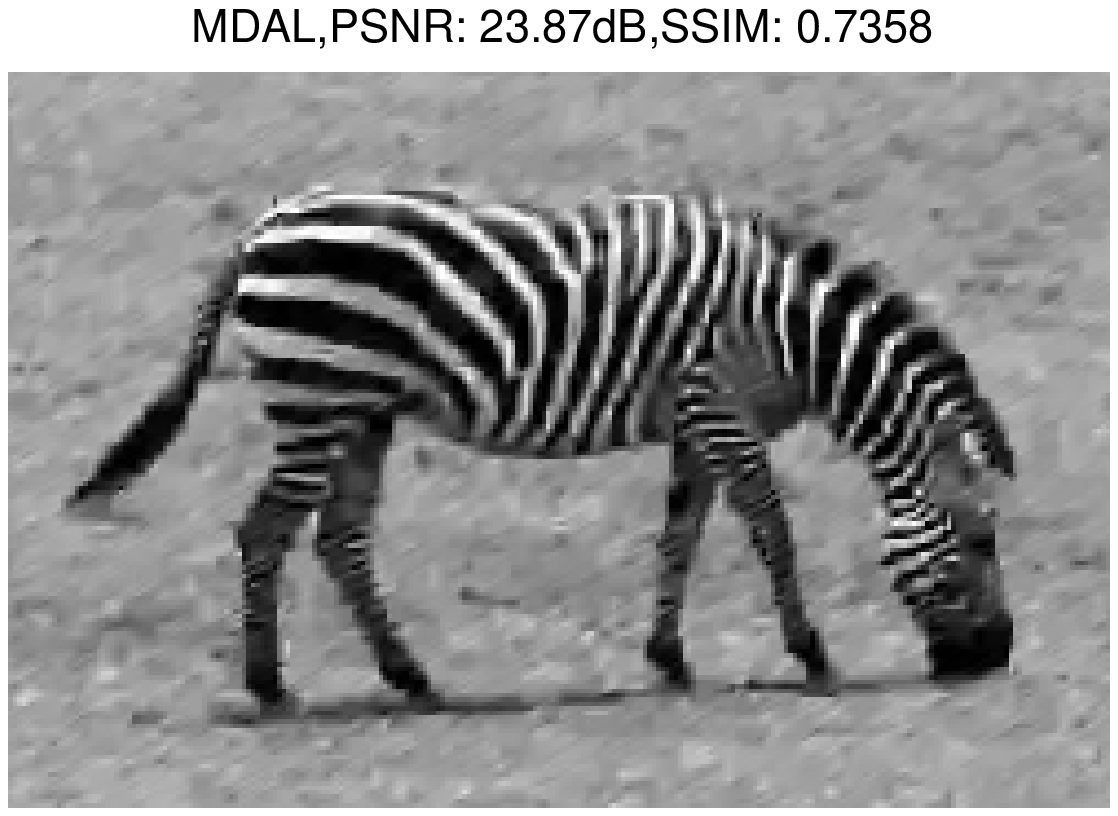}
   \includegraphics[width=0.22\textwidth]{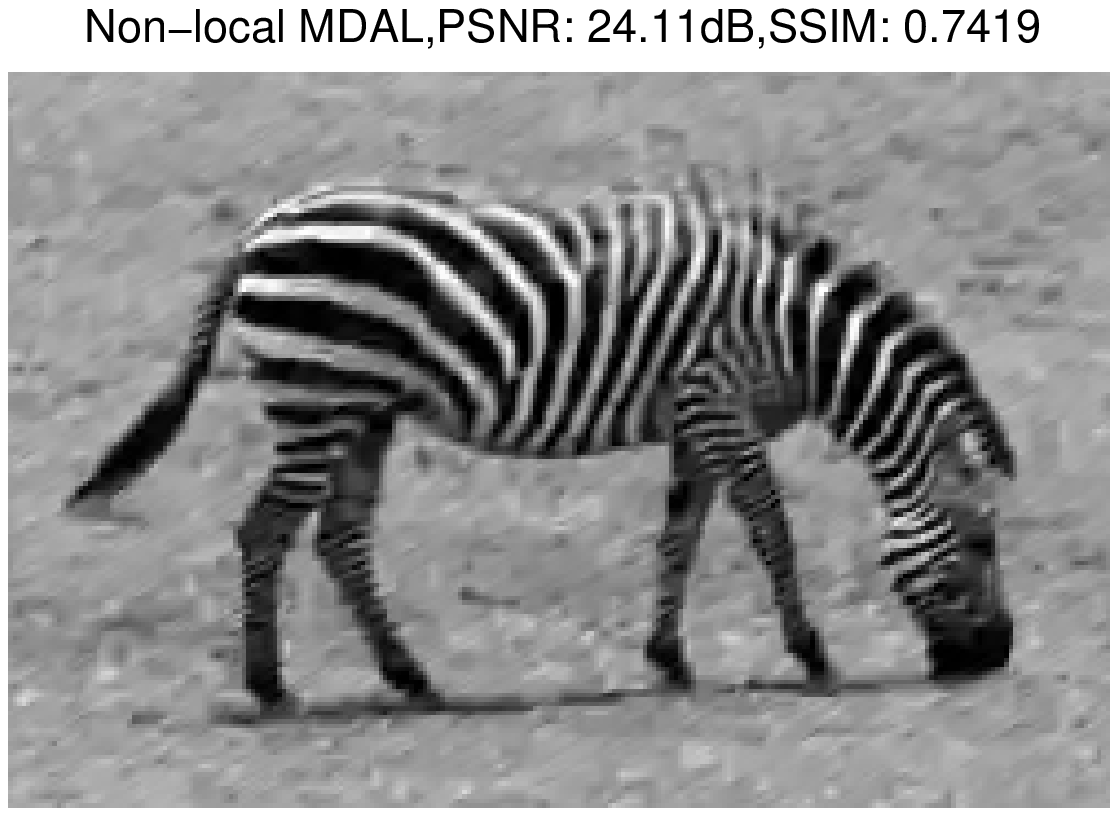}
   \includegraphics[width=0.22\textwidth]{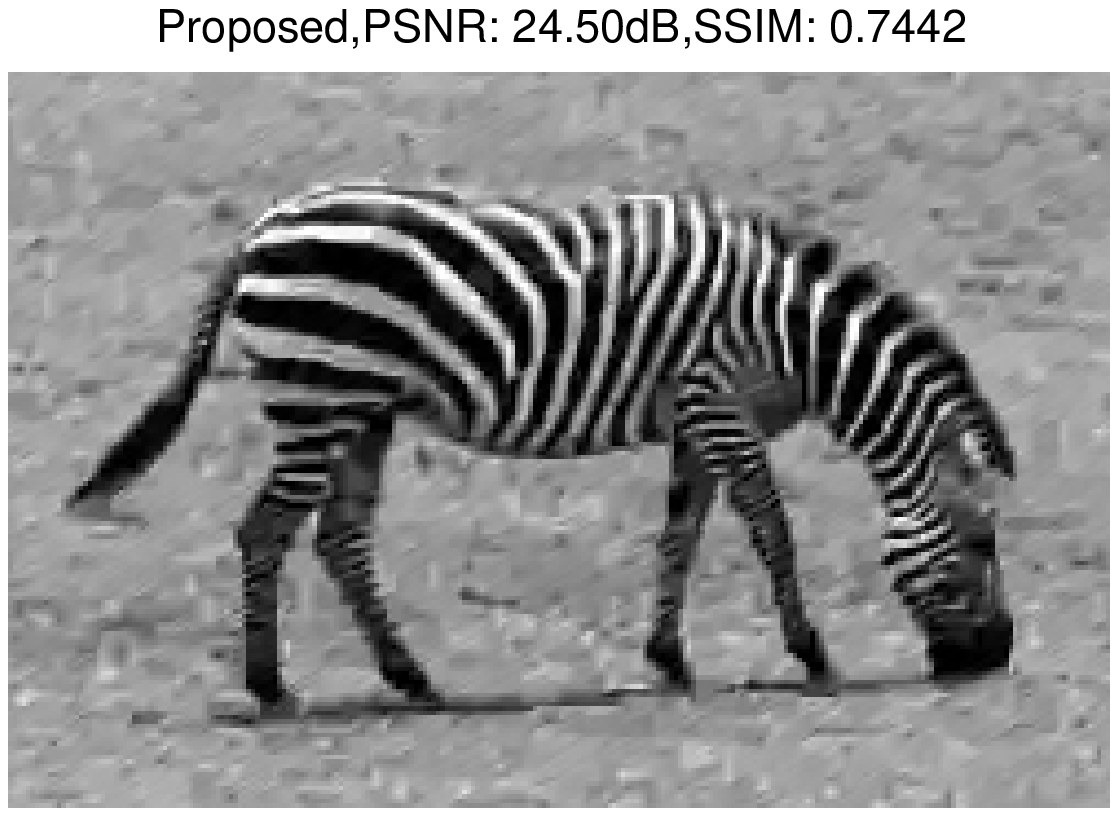}
   \includegraphics[width=0.22\textwidth]{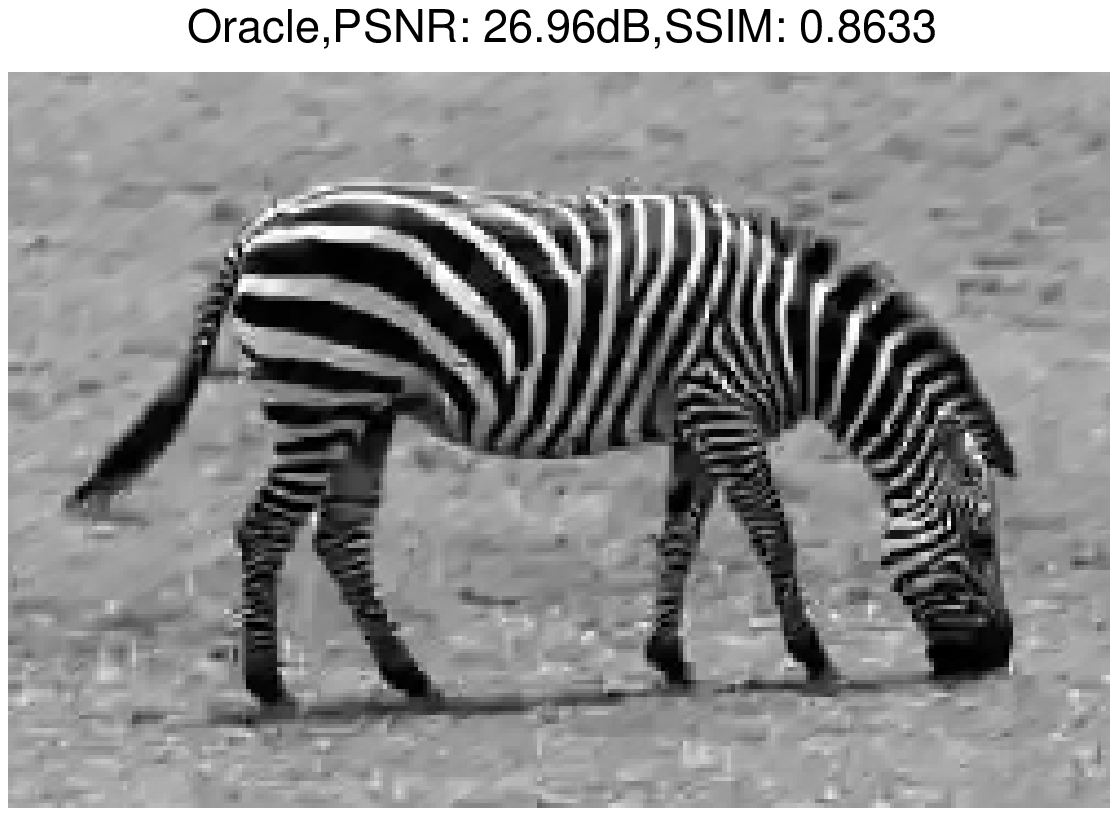} \\

  \caption{\small The visual comparison between different methods. From the left to right and up to down: the original image, the blurry and noisy image (blur kernel Type IV with noise level $\sigma=4.0$), recovered results by Non-local TV method \cite{Zhangxiaoqun2010}, Split Bregman method \cite{Cai2009a}, MDAL method \cite{Dong2013}, Non-local MDAL method \cite{Chen2015}, Proposed method, Oracle method, respectively.}
\label{fig: motion154 comparisons}
\end{figure}

In Figure \ref{fig: intermediate results}, we present the recovery results of the intermediate stages of our proposed algorithm. Due to the page limit, we just choose to show the results of the test image cameraman here, since the other cases have the similar conclusions. Recall that the first stage of our proposed algorithm is to run the Non-local MDAL method, and we can observe that our proposed algorithm can bring gradual improvements as the support detection proceeds. Empirically, our proposed algorithm achieves the best final results when the maximum stage number is only 2 or 3 in most cases.

\begin{figure}[h]
  \centering

    \includegraphics[width=0.32\textwidth]{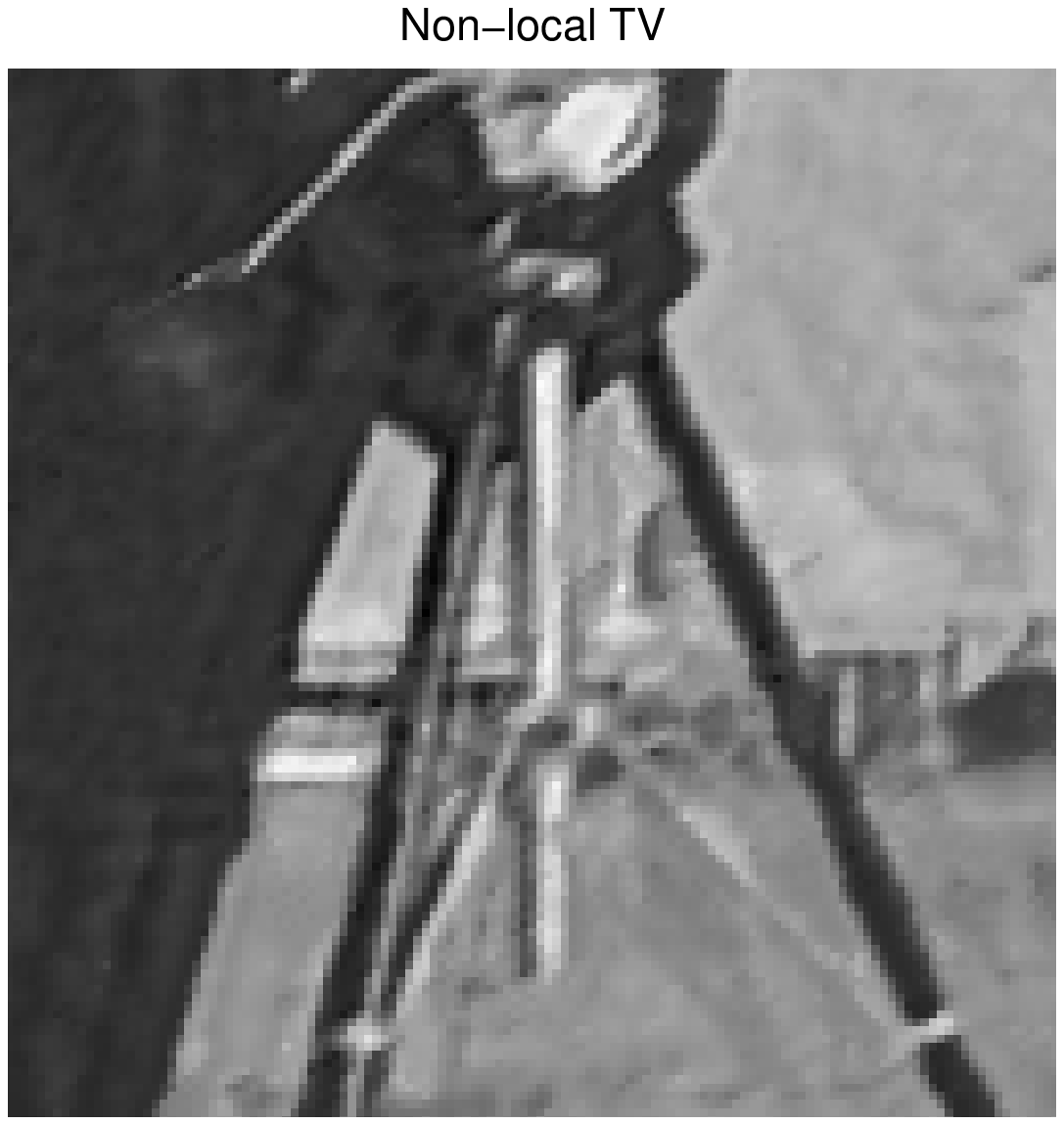}
   \includegraphics[width=0.32\textwidth]{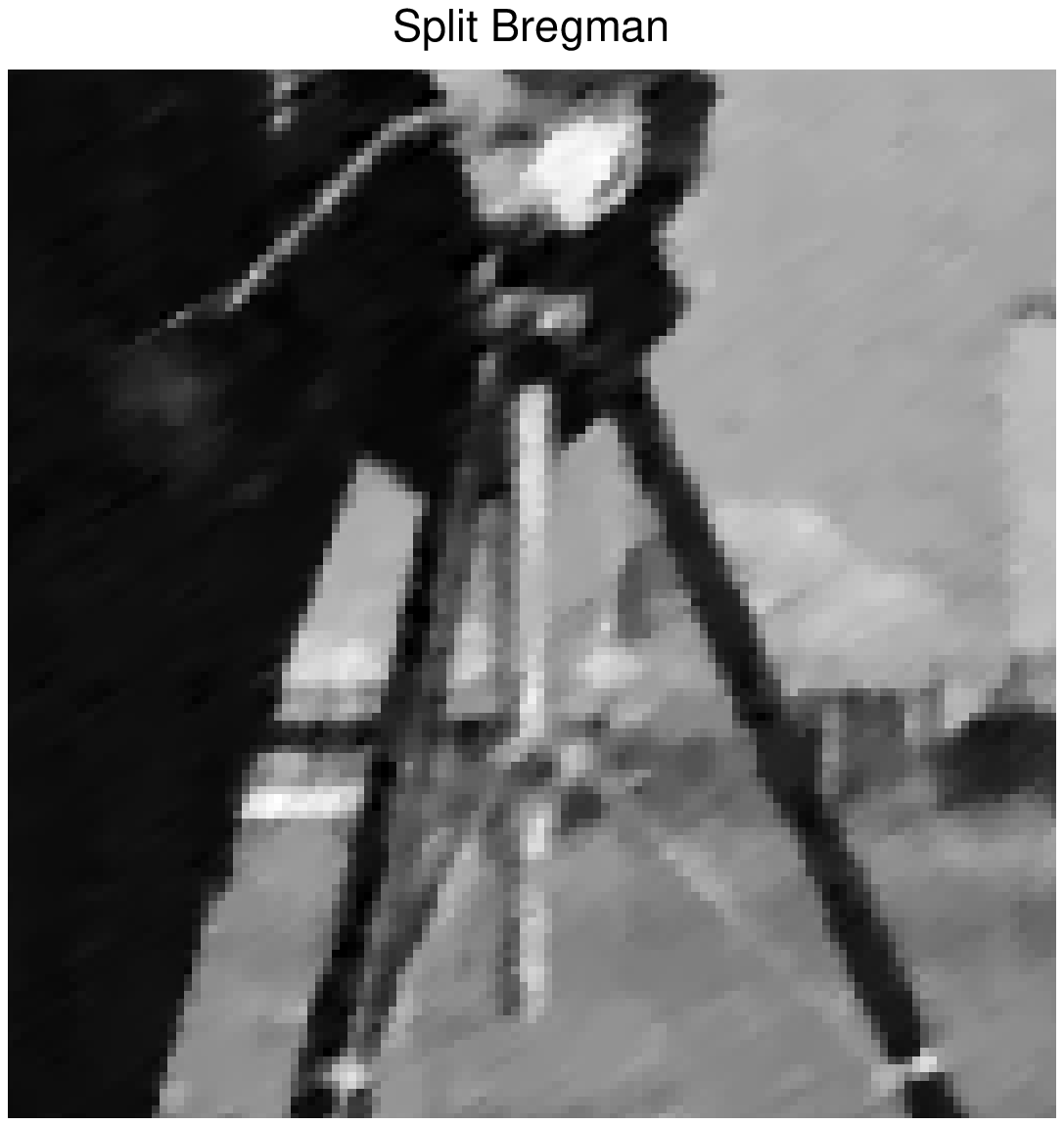}
   \includegraphics[width=0.32\textwidth]{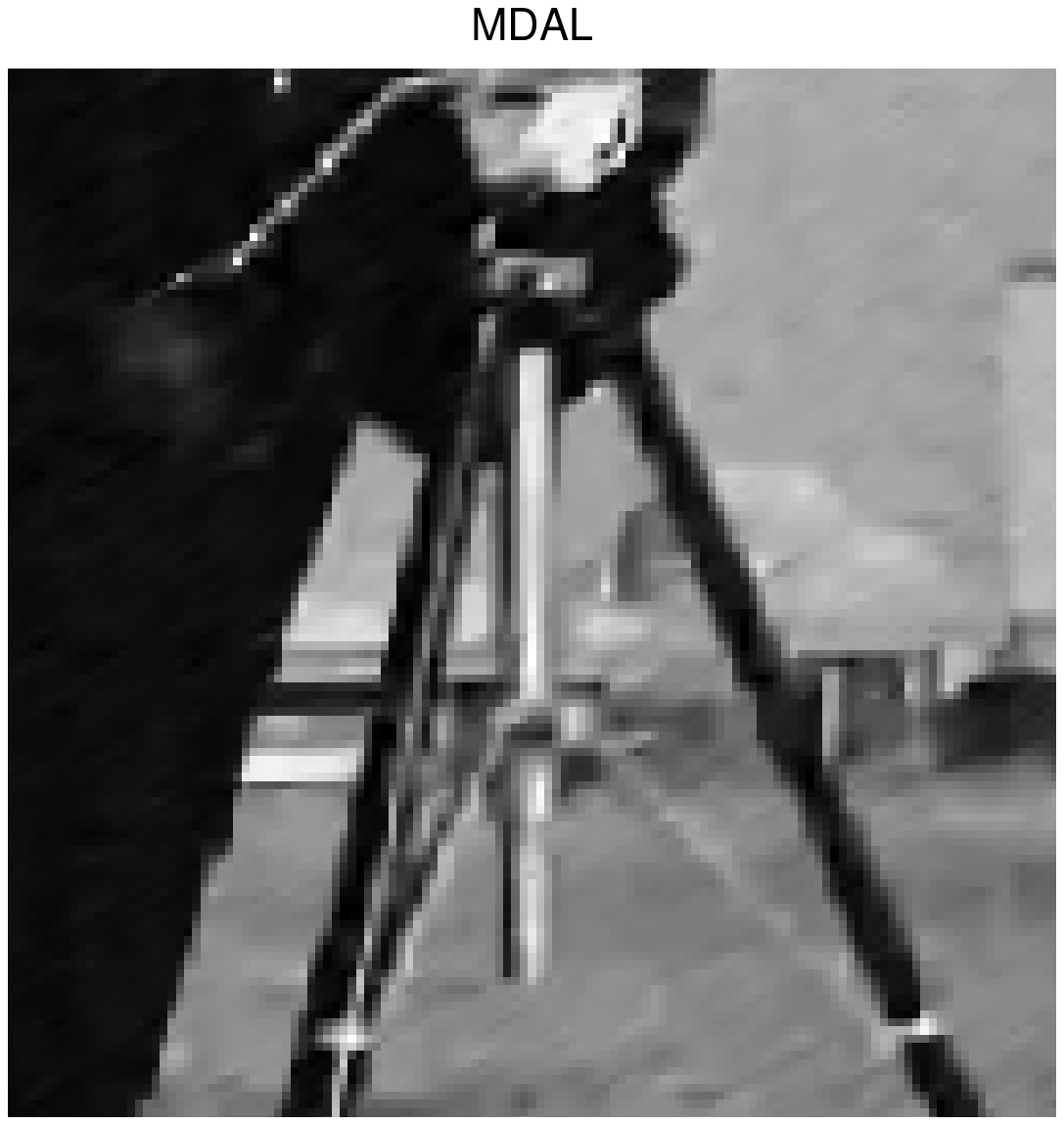} \\
   \includegraphics[width=0.32\textwidth]{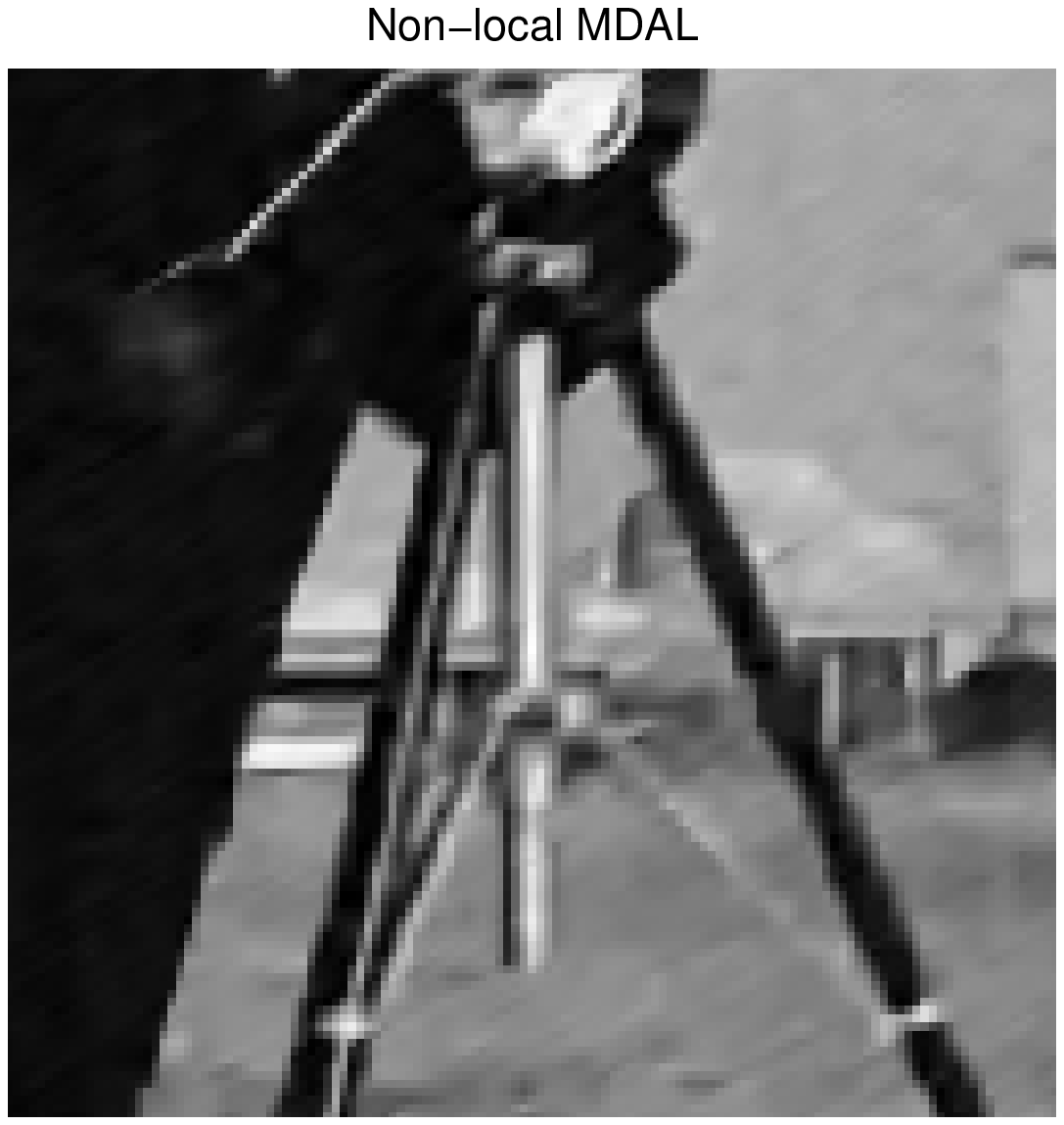}
   \includegraphics[width=0.32\textwidth]{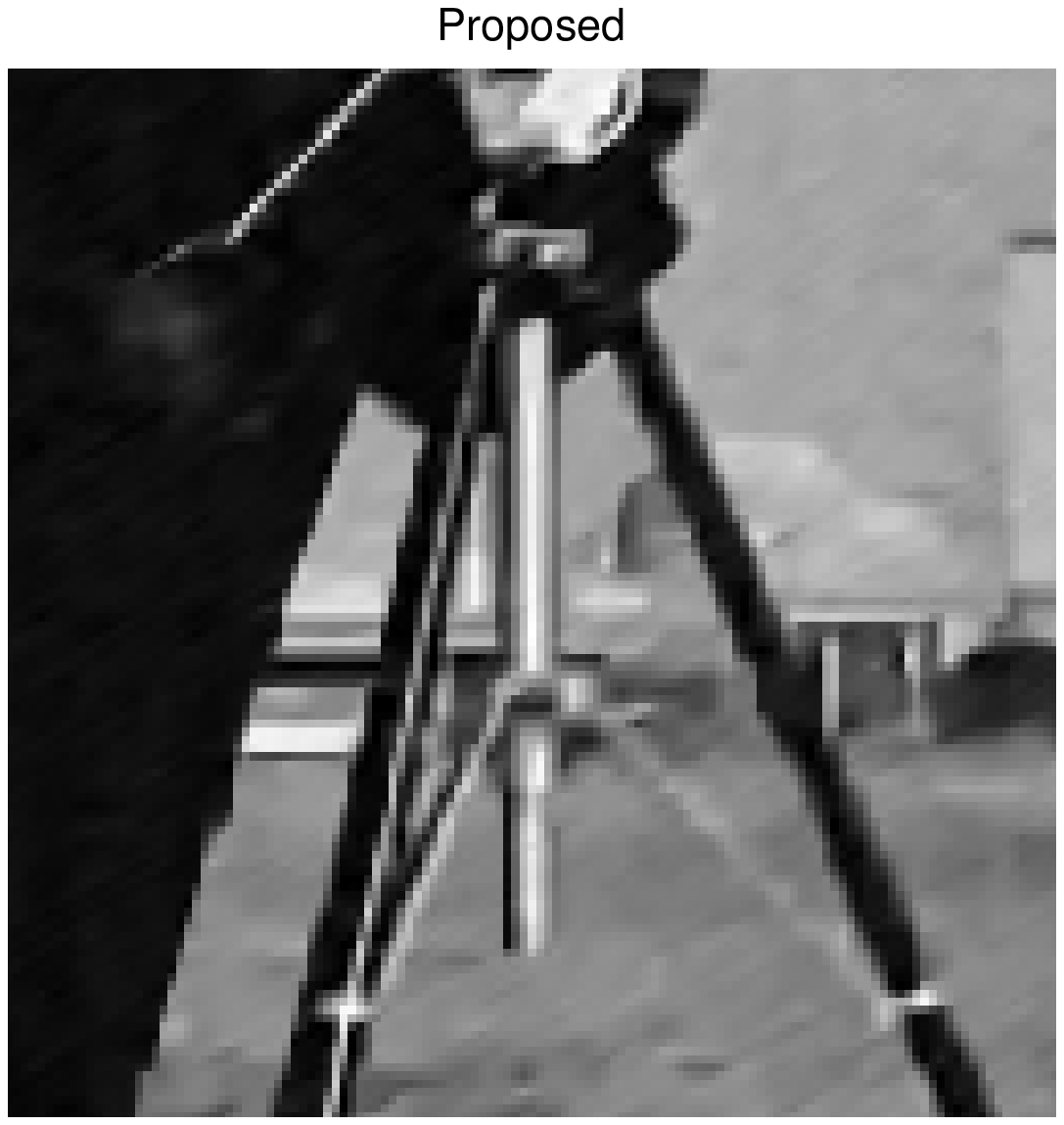}
   \includegraphics[width=0.32\textwidth]{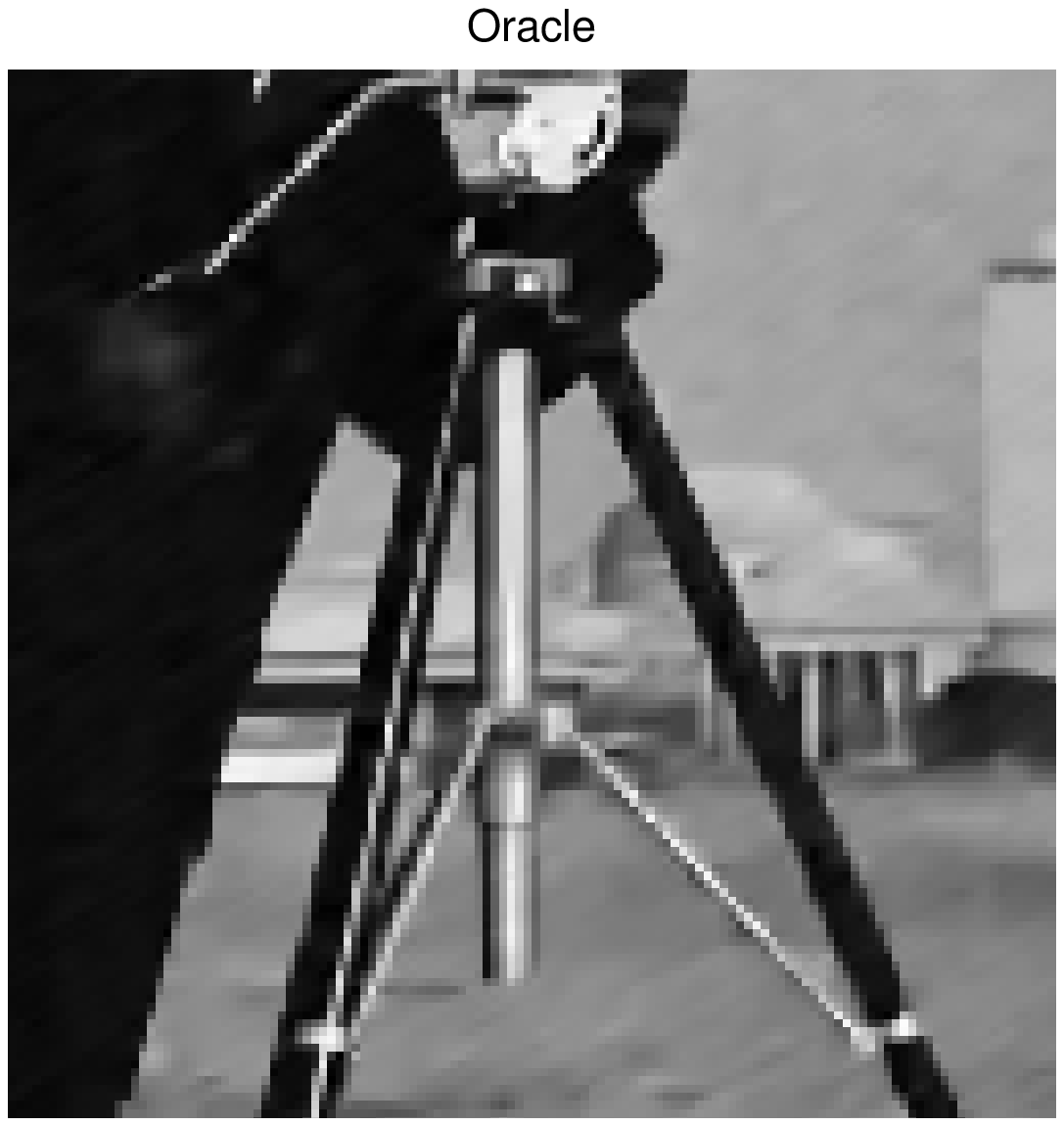} \\
       \includegraphics[width=0.32\textwidth]{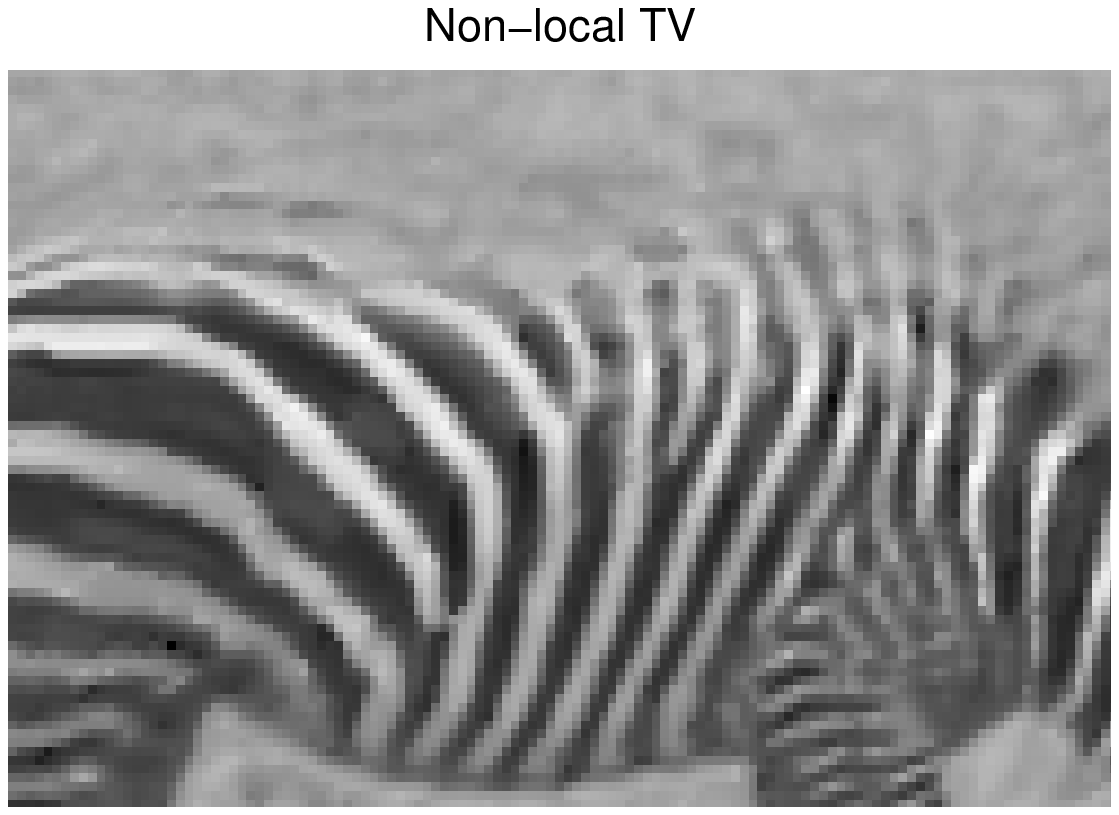}
   \includegraphics[width=0.32\textwidth]{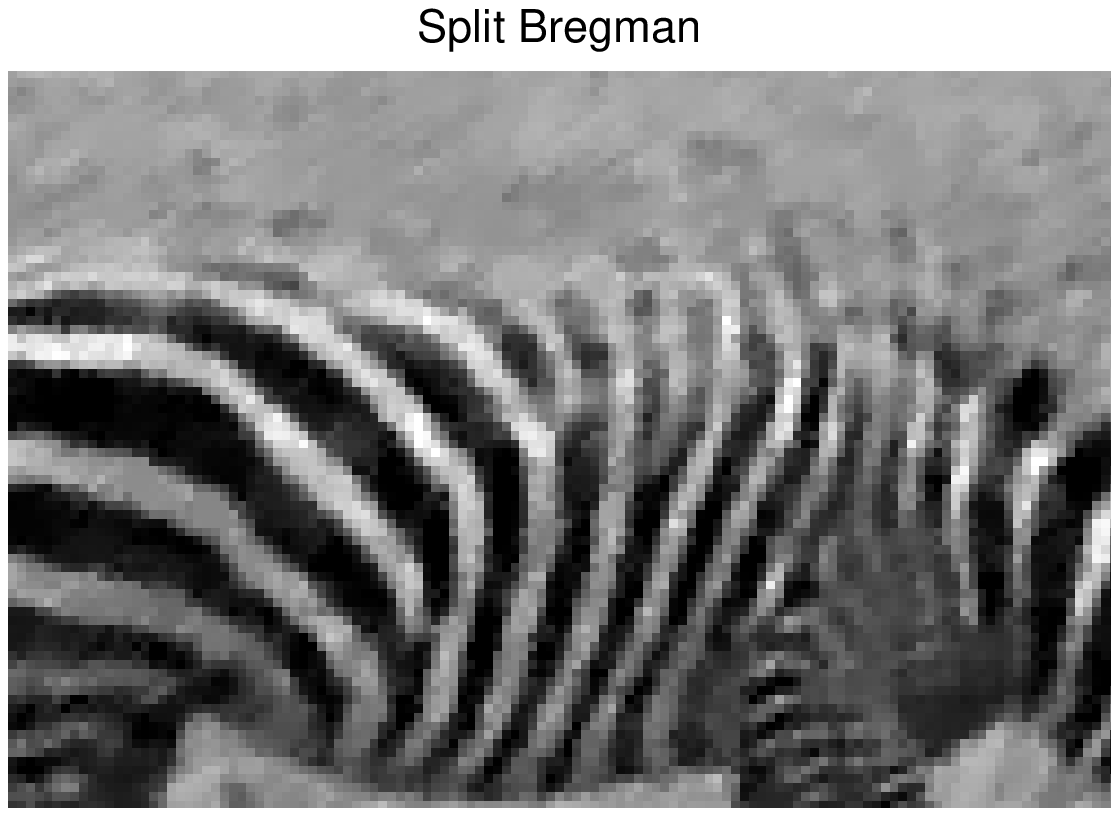}
   \includegraphics[width=0.32\textwidth]{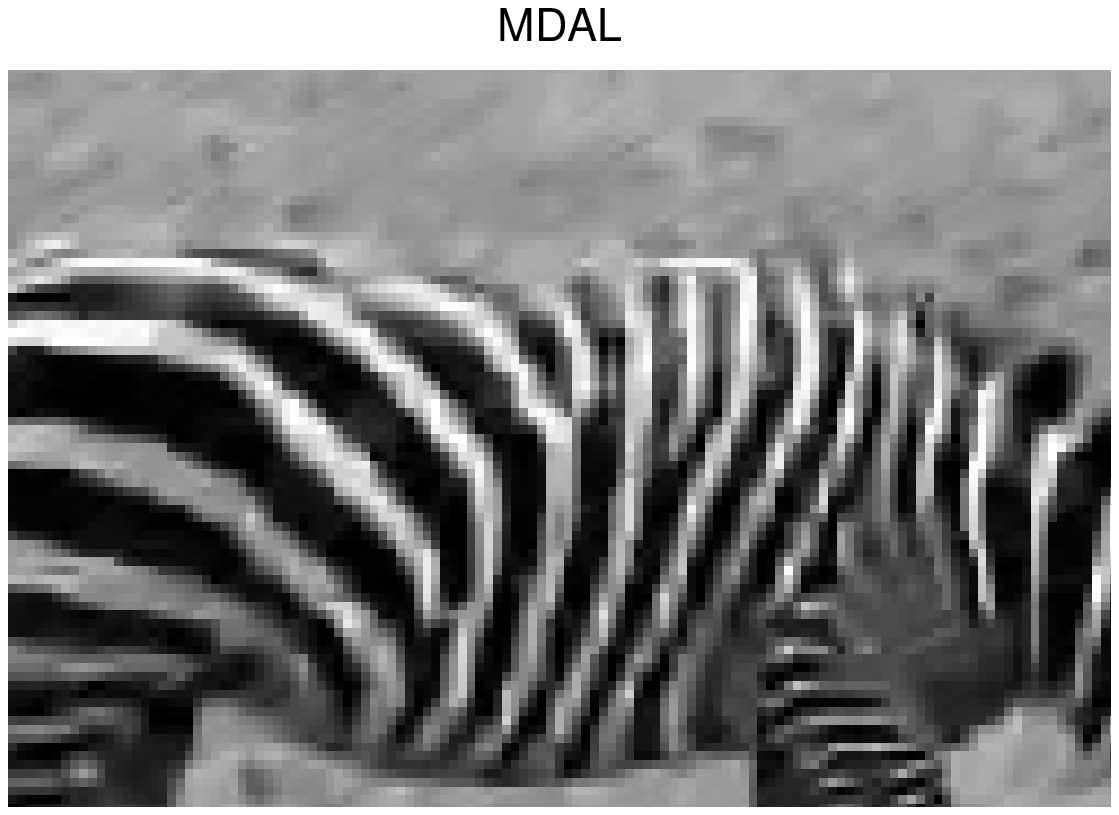} \\
   \includegraphics[width=0.32\textwidth]{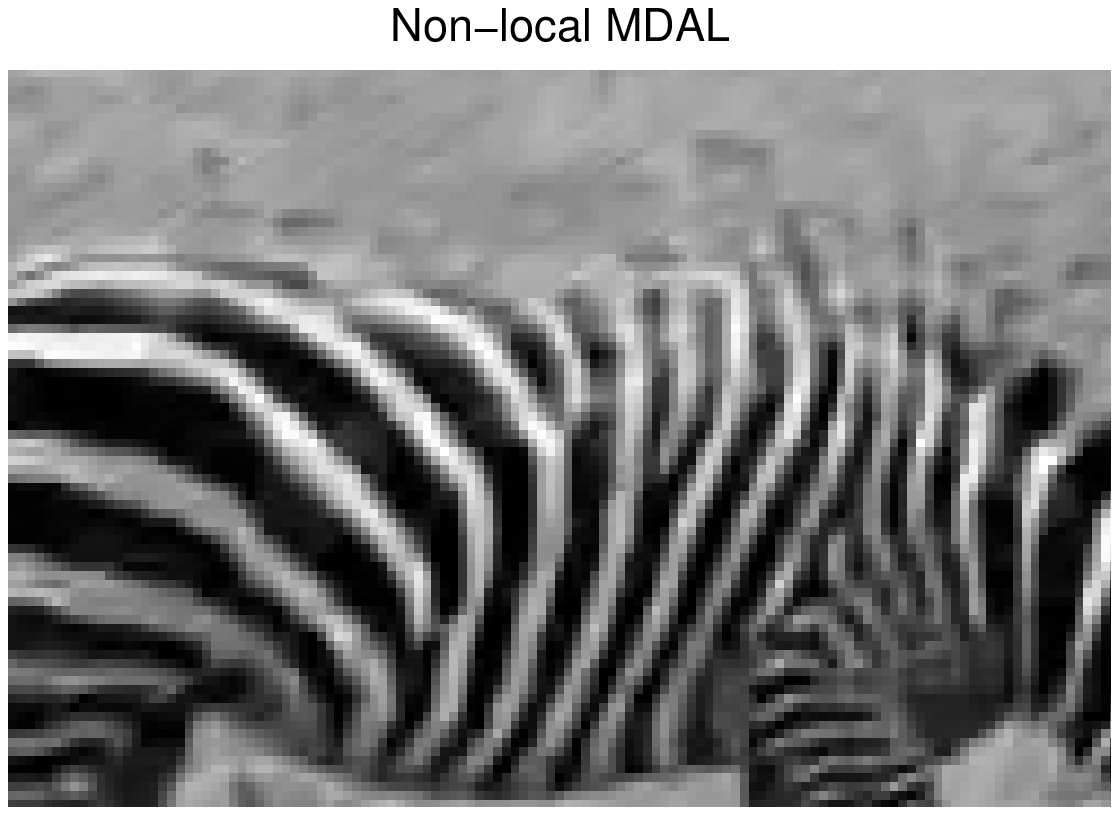}
   \includegraphics[width=0.32\textwidth]{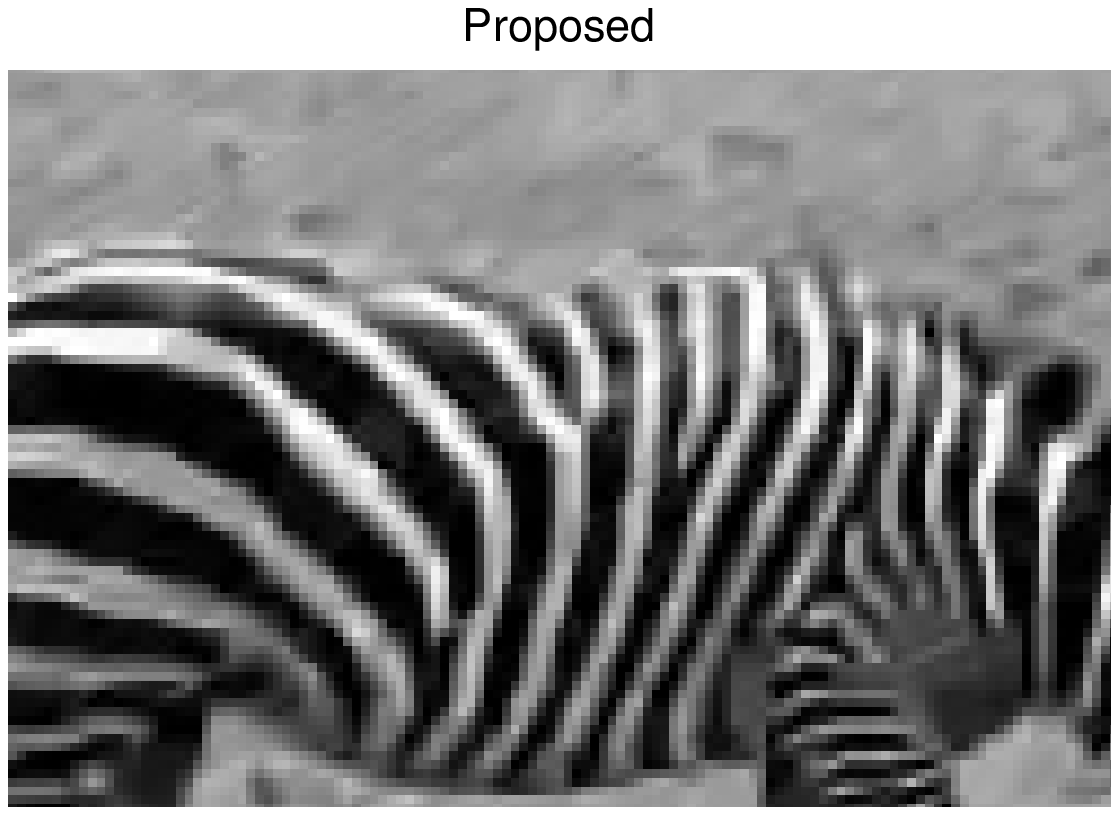}
   \includegraphics[width=0.32\textwidth]{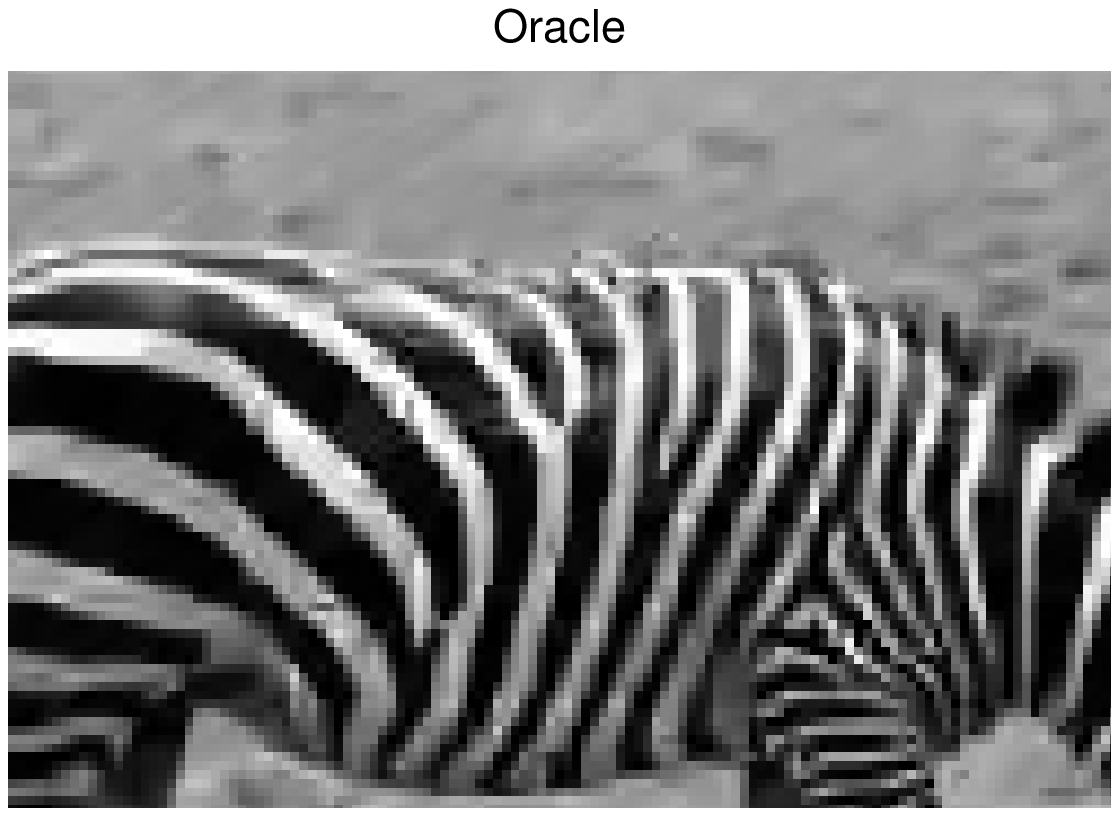} \\

  \caption{\small The zoom in visual detail comparisons between different methods corresponding to Figure \ref{fig: motion154 comparisons}. }
\label{fig: fig: zoomin motion154 comparisons}
\end{figure}

In what follows, we further analyze the advantage of the proposed algorithm through the visual comparisons between different methods. We choose the test images cameraman and zebra1 here, and the images are convoluted by blur type IV and contaminated by noise with $\sigma=4.0$. We can observe that the proposed algorithm has the potential to recover the edge regions and textures simultaneously. In addition, in order to make the visual comparisons more clear, we give the close-up details (zoom-ins) of recovery results displayed in Figure \ref{fig: motion154 comparisons}, and in Figure \ref{fig: fig: zoomin motion154 comparisons}, respectively. It can be observed that our proposed algorithm brings significant visual enhancements in the sharp edges of the recovery images compared to the other methods, e.g., the tripod of the cameraman image and the stipes of the zebra1 image.

\begin{figure}[h]
  \centering

   \includegraphics[width=0.22\textwidth]{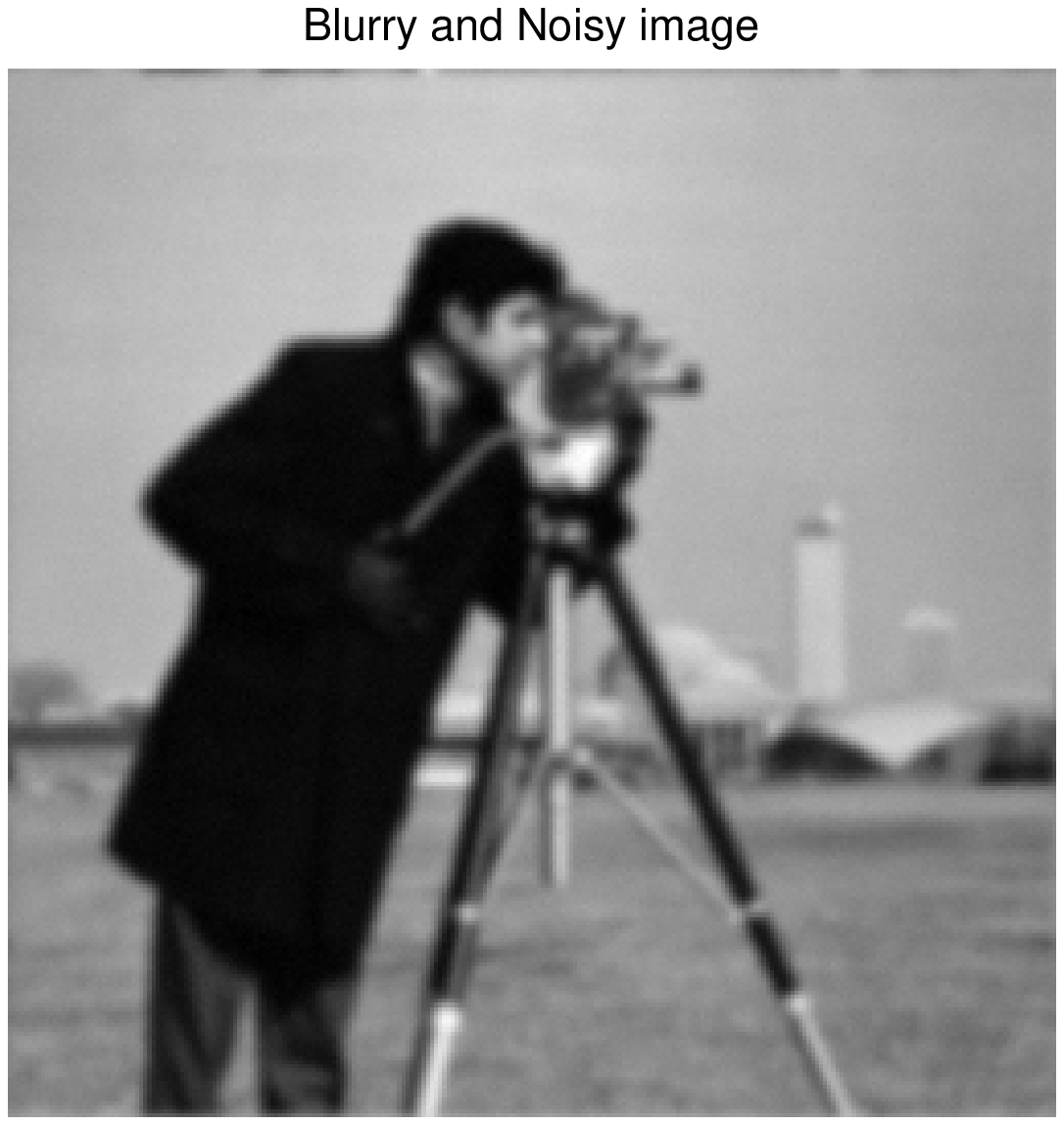}
   \includegraphics[width=0.22\textwidth]{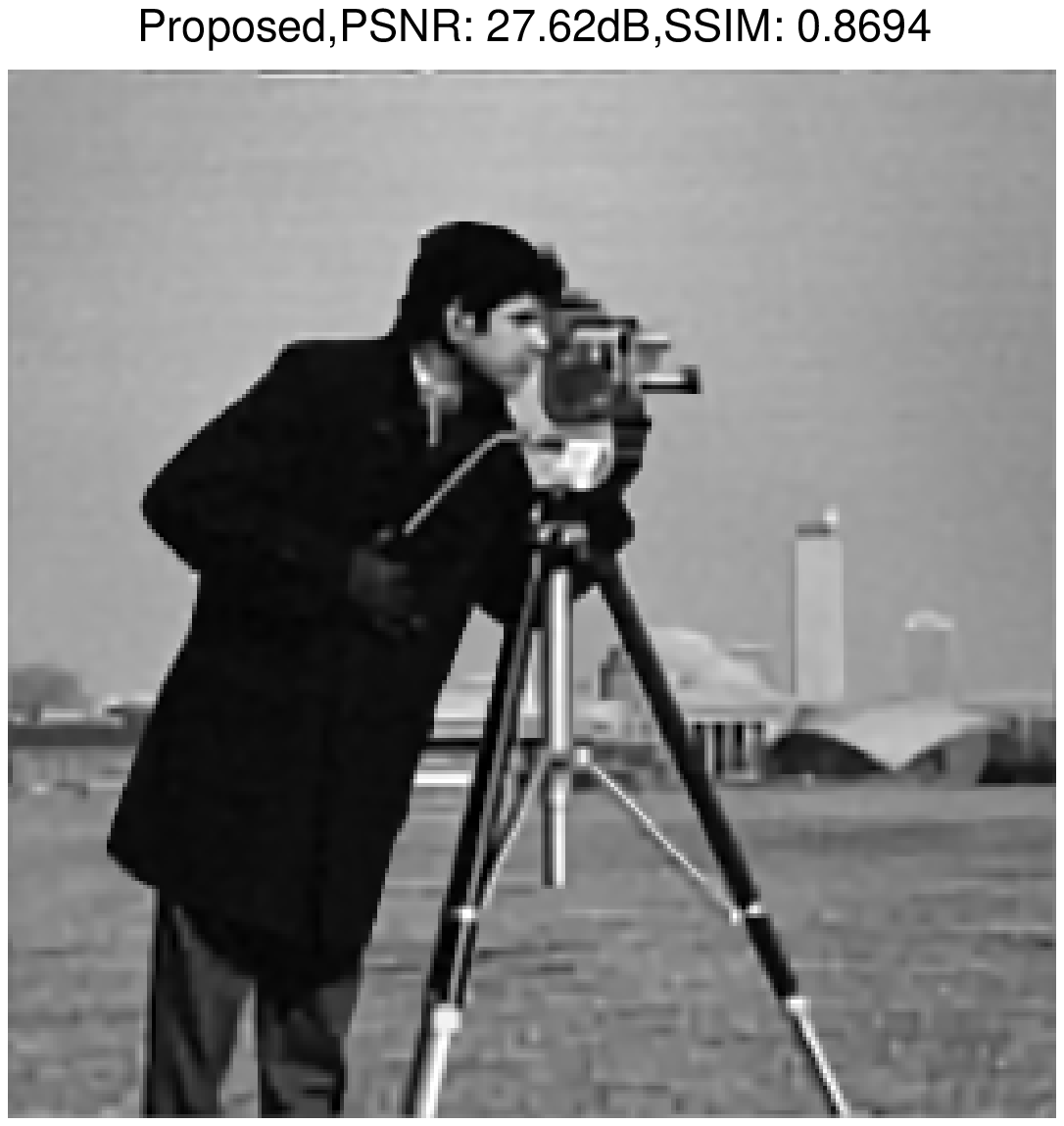}
   \includegraphics[width=0.22\textwidth]{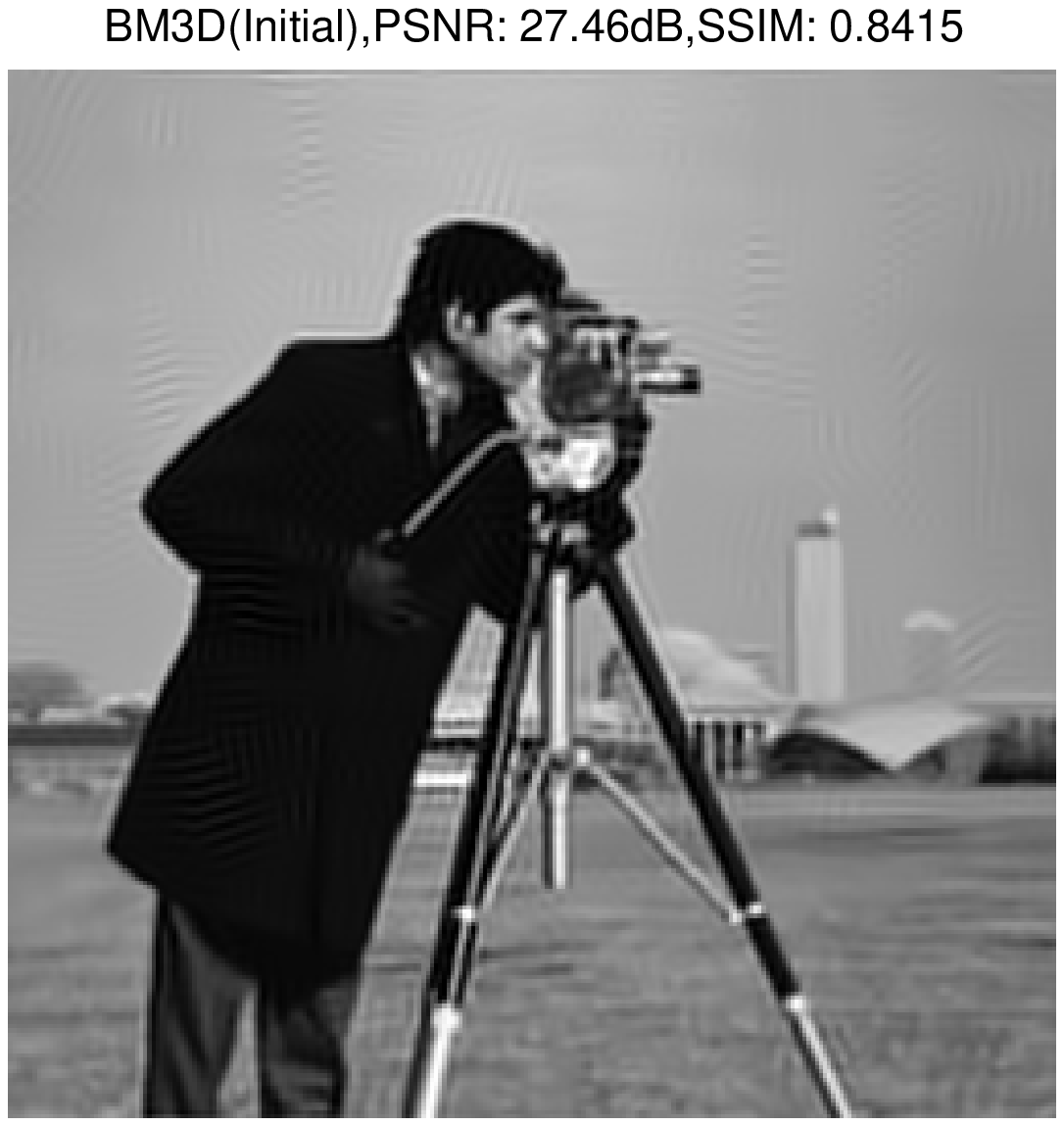}
   \includegraphics[width=0.22\textwidth]{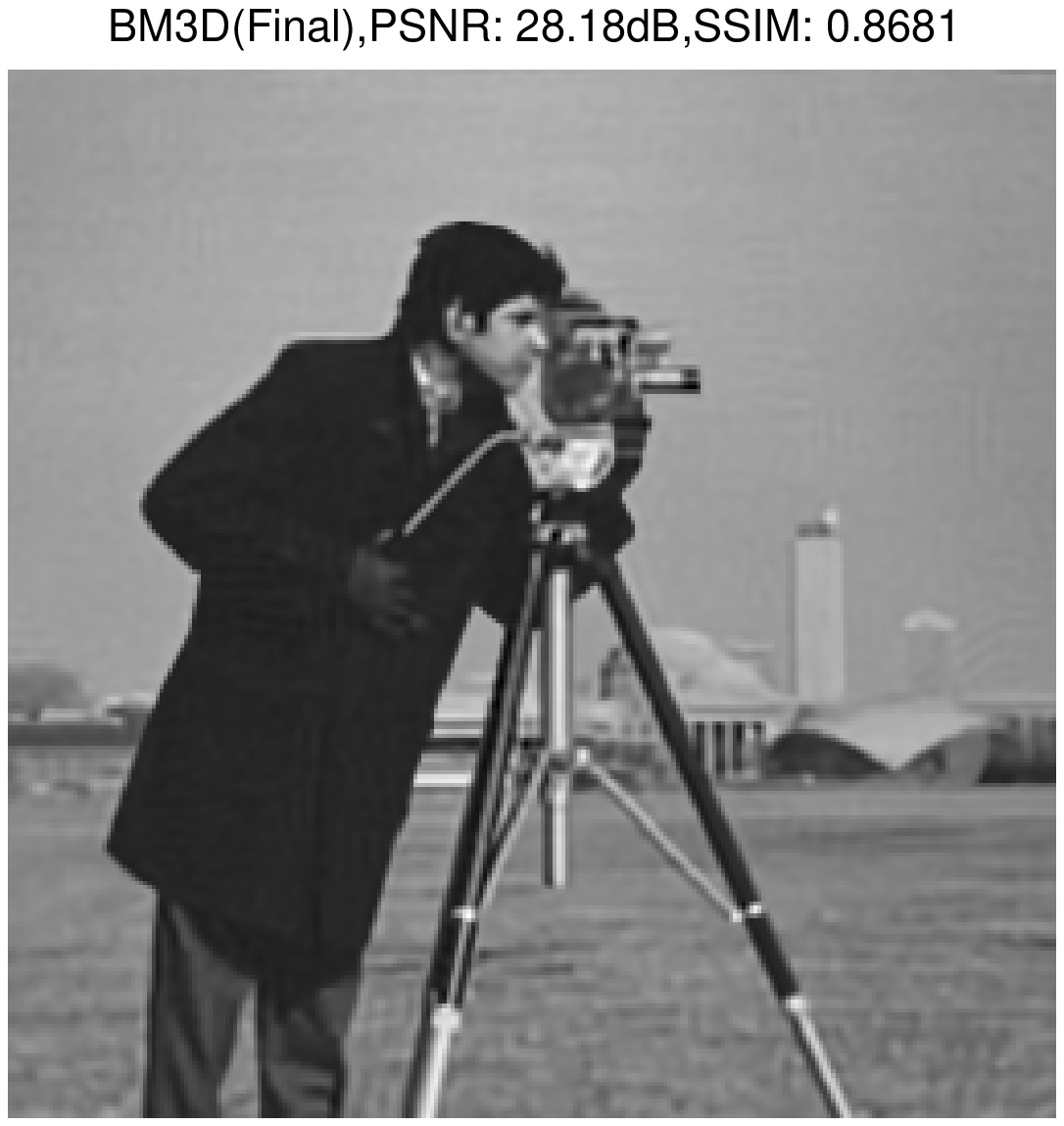} \\
   \includegraphics[width=0.22\textwidth]{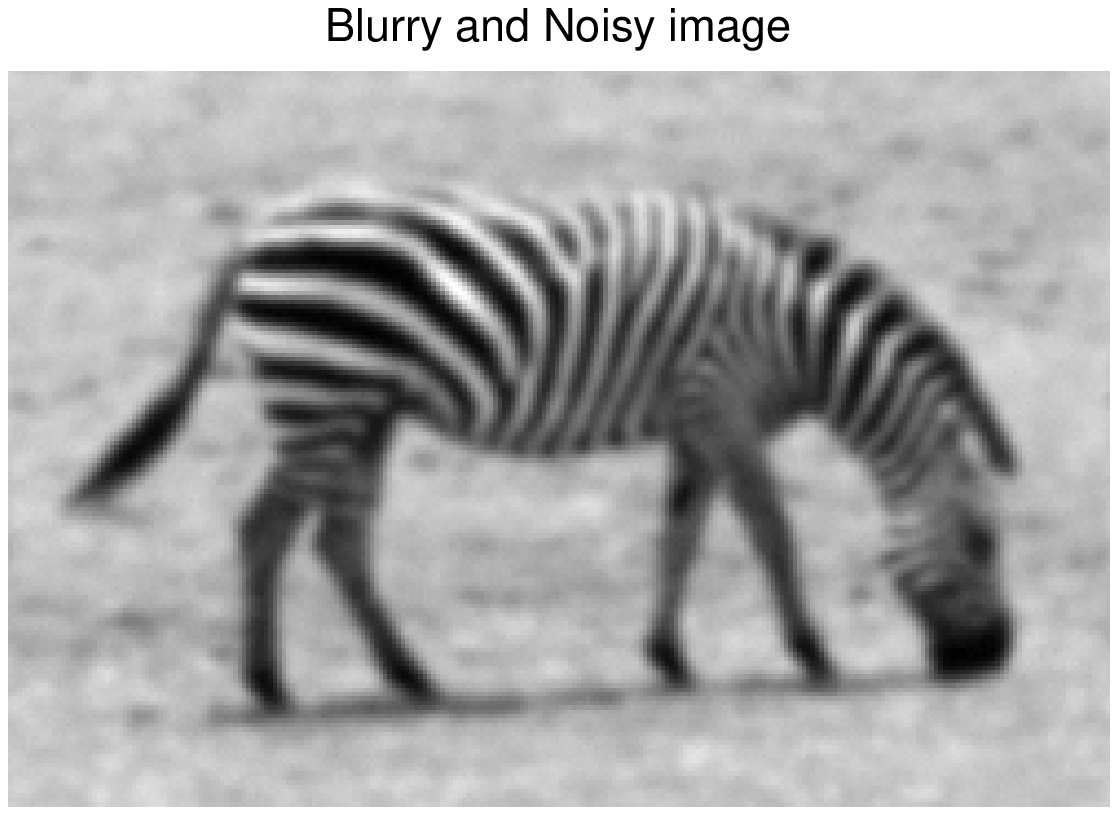}
   \includegraphics[width=0.22\textwidth]{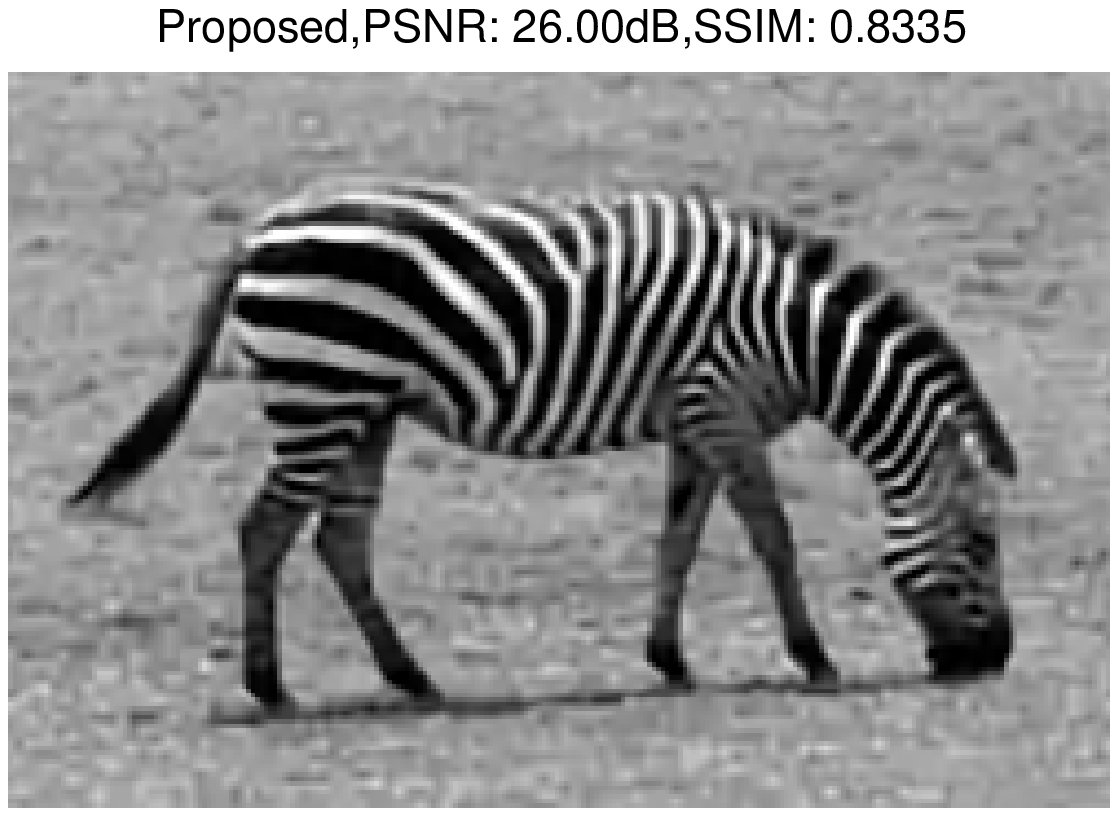}
       \includegraphics[width=0.22\textwidth]{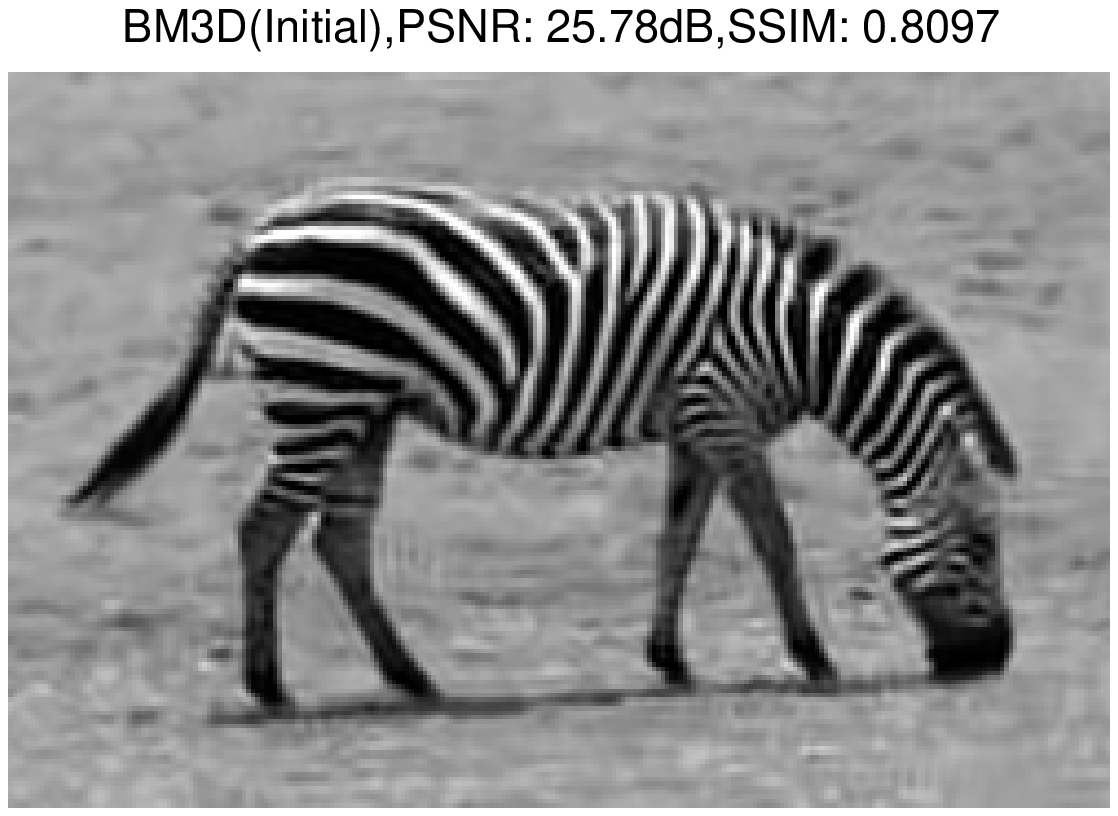}
     \includegraphics[width=0.22\textwidth]{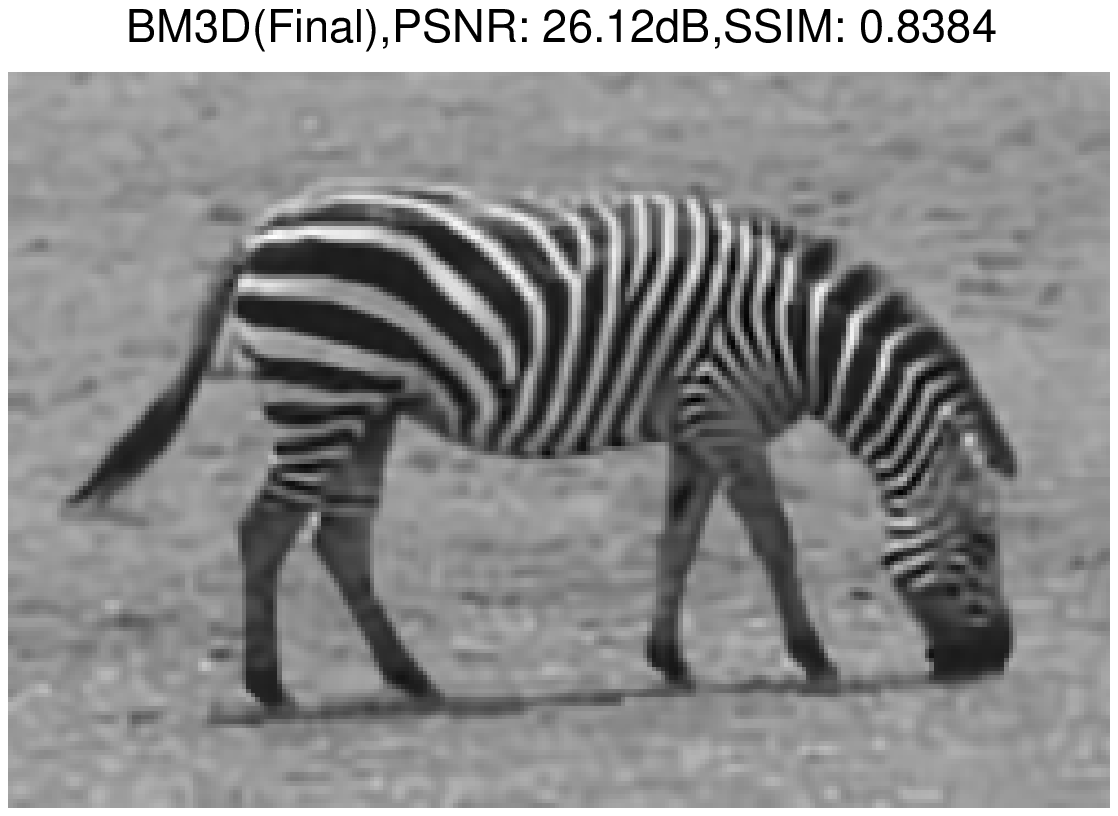} \\

  \caption{\small The visual comparisons between proposed method with BM3D methods \cite{Dabov2008, Danielyan2012}. The first column to fourth column correspond to the blurry and noisy image (blur kernel Type II with noise level $\sigma=\sqrt{2}$), recovered results of Proposed method, IDD-BM3D initial estimate, IDD-BM3D final estimate, respectively. }
\label{fig: BM3D}
\end{figure}

\begin{figure}[h]
  \centering
   \includegraphics[width=0.32\textwidth]{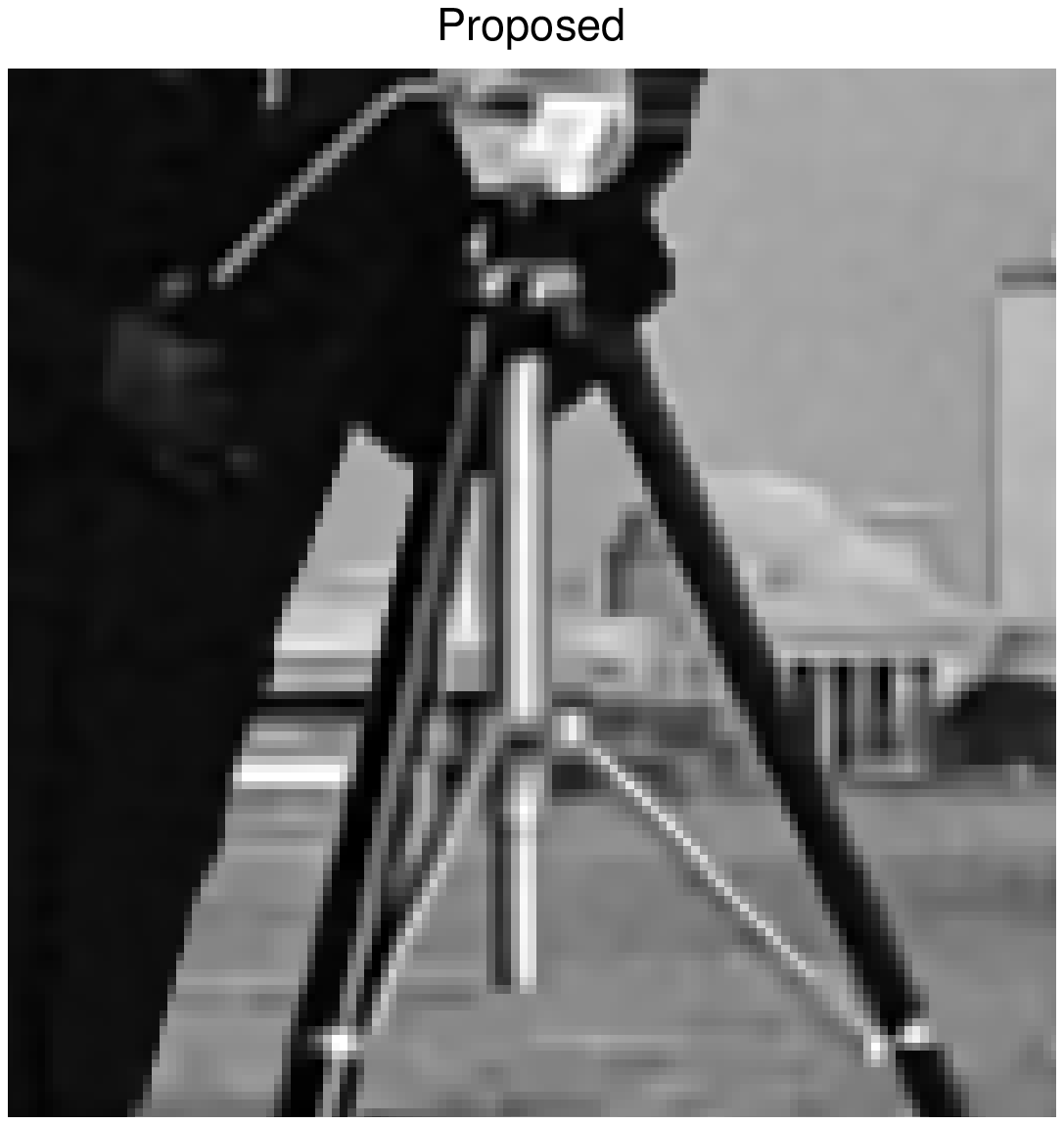}
   \includegraphics[width=0.32\textwidth]{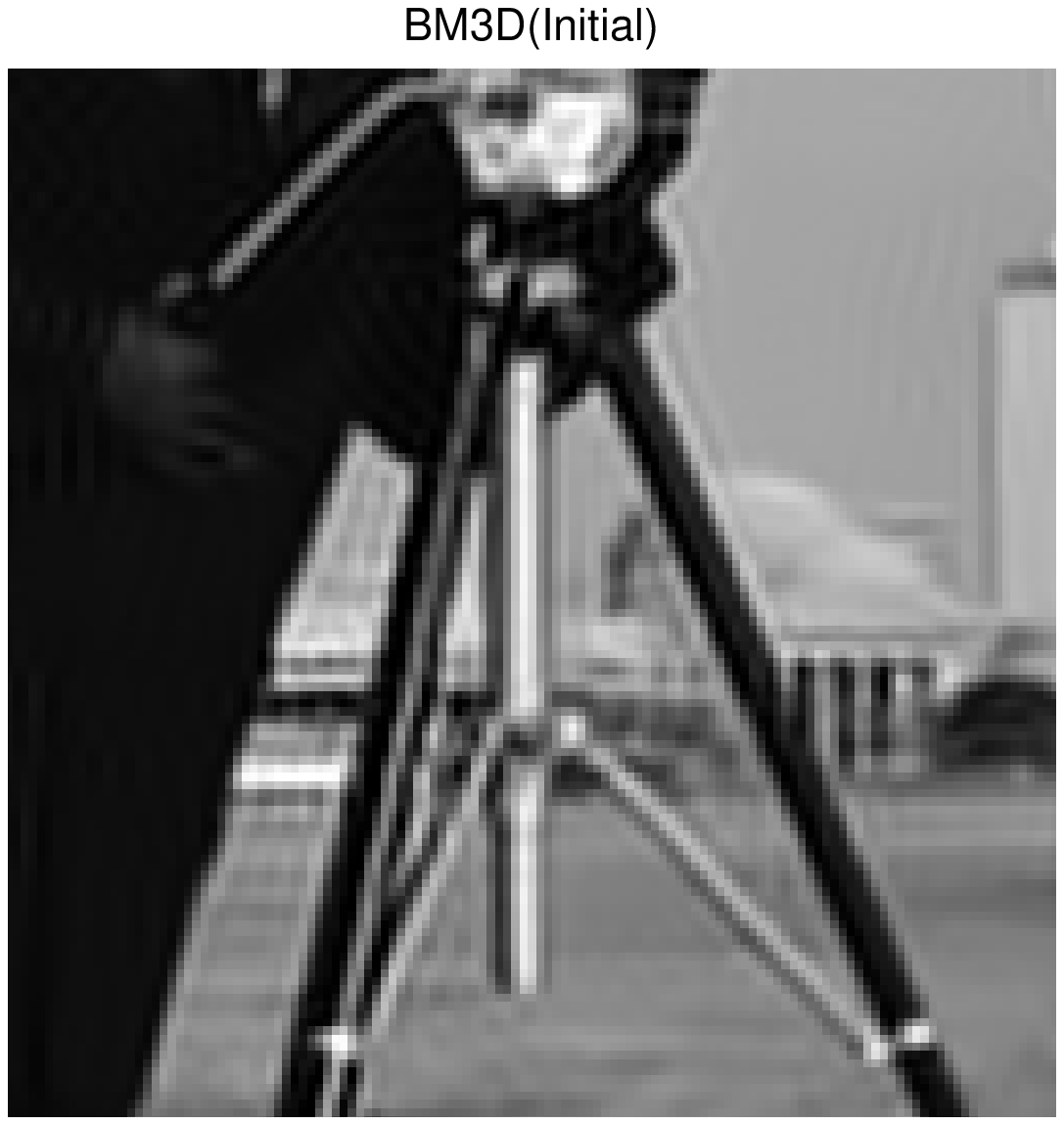}
   \includegraphics[width=0.32\textwidth]{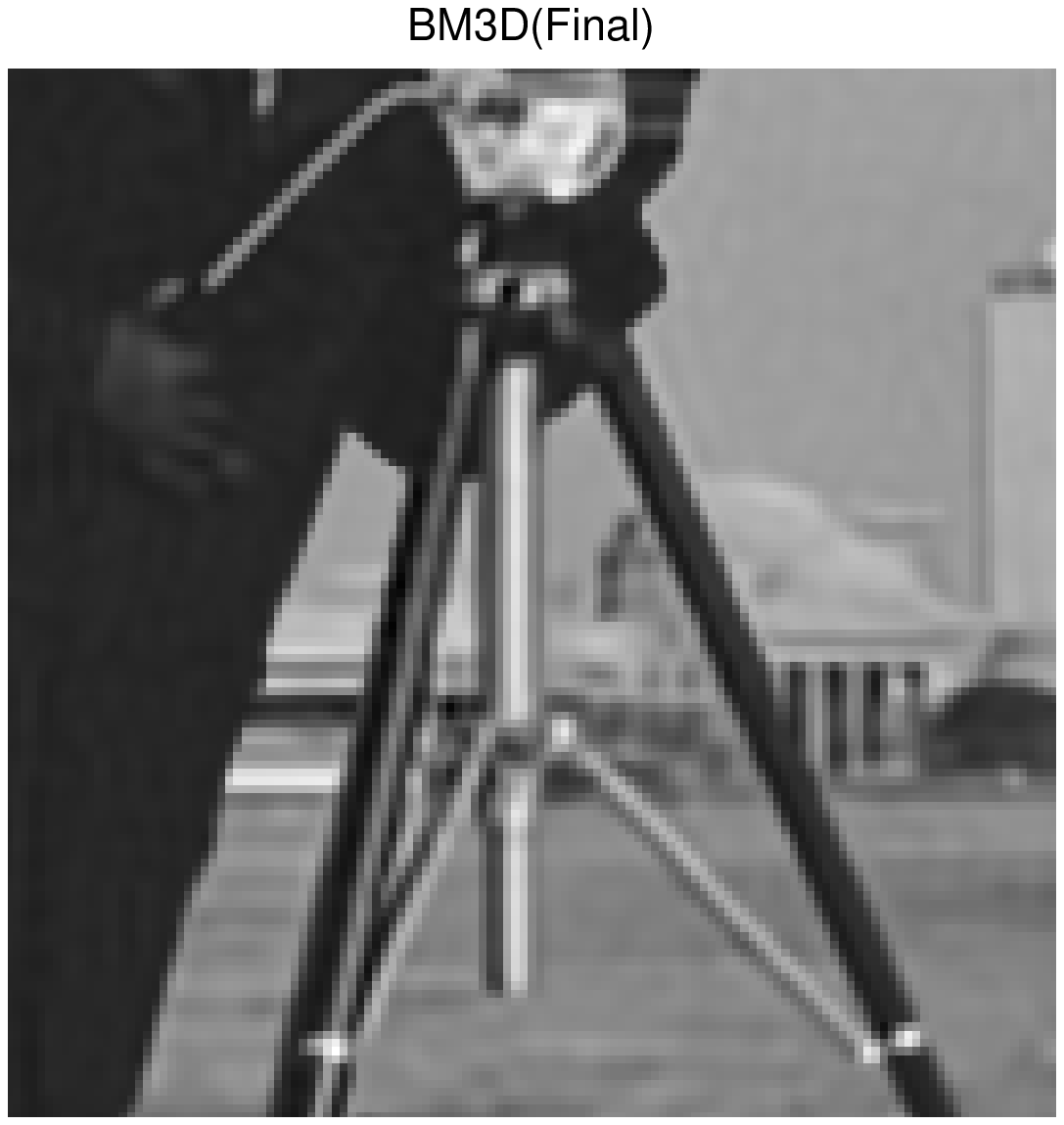} \\
   \includegraphics[width=0.32\textwidth]{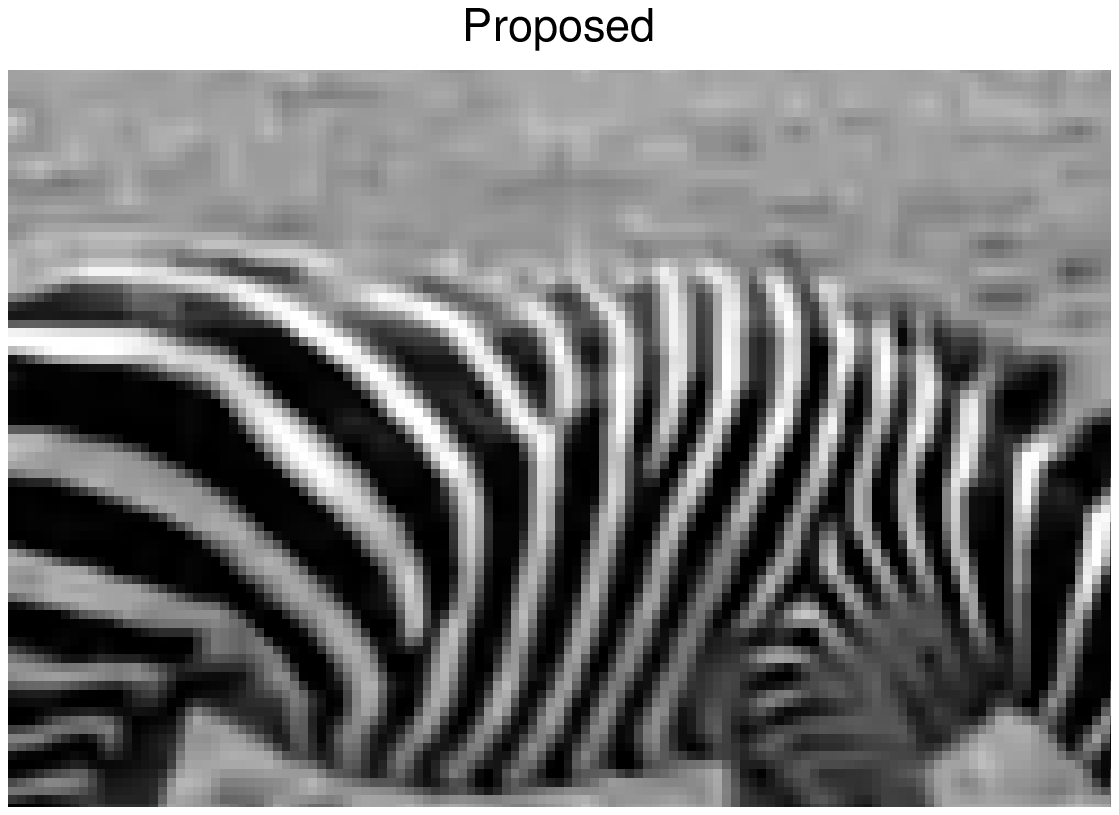}
   \includegraphics[width=0.32\textwidth]{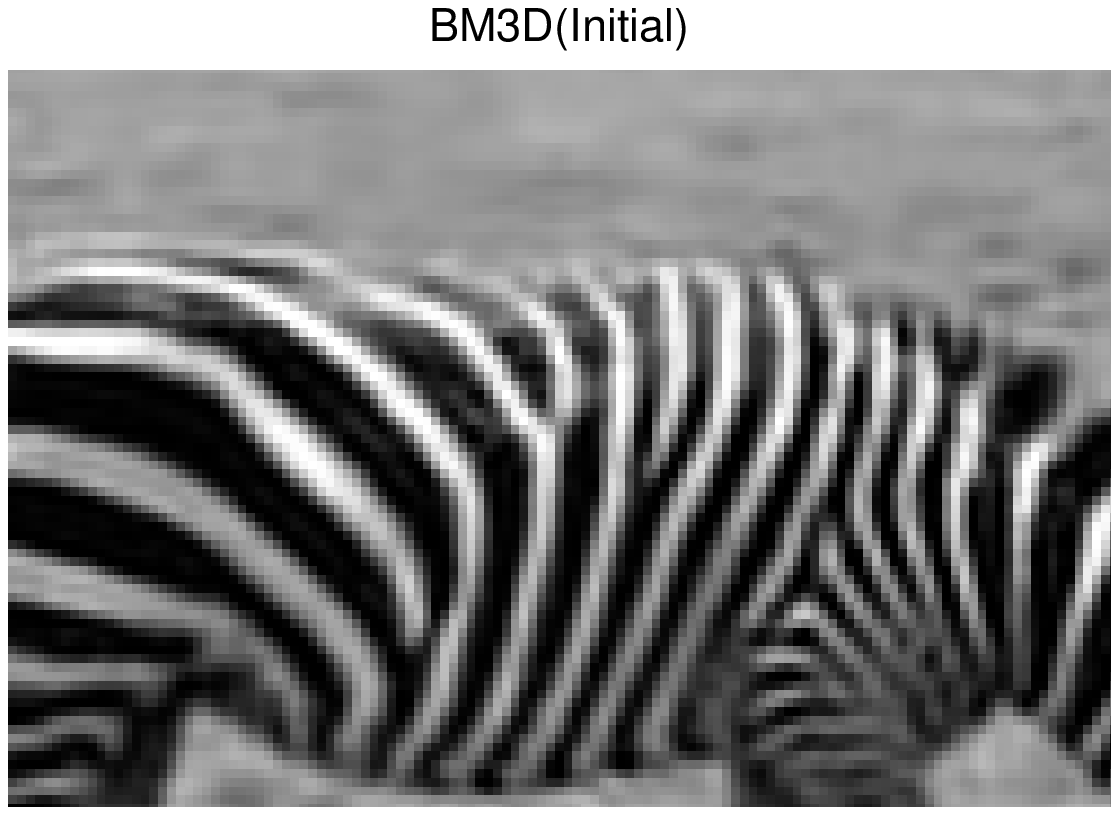}
   \includegraphics[width=0.32\textwidth]{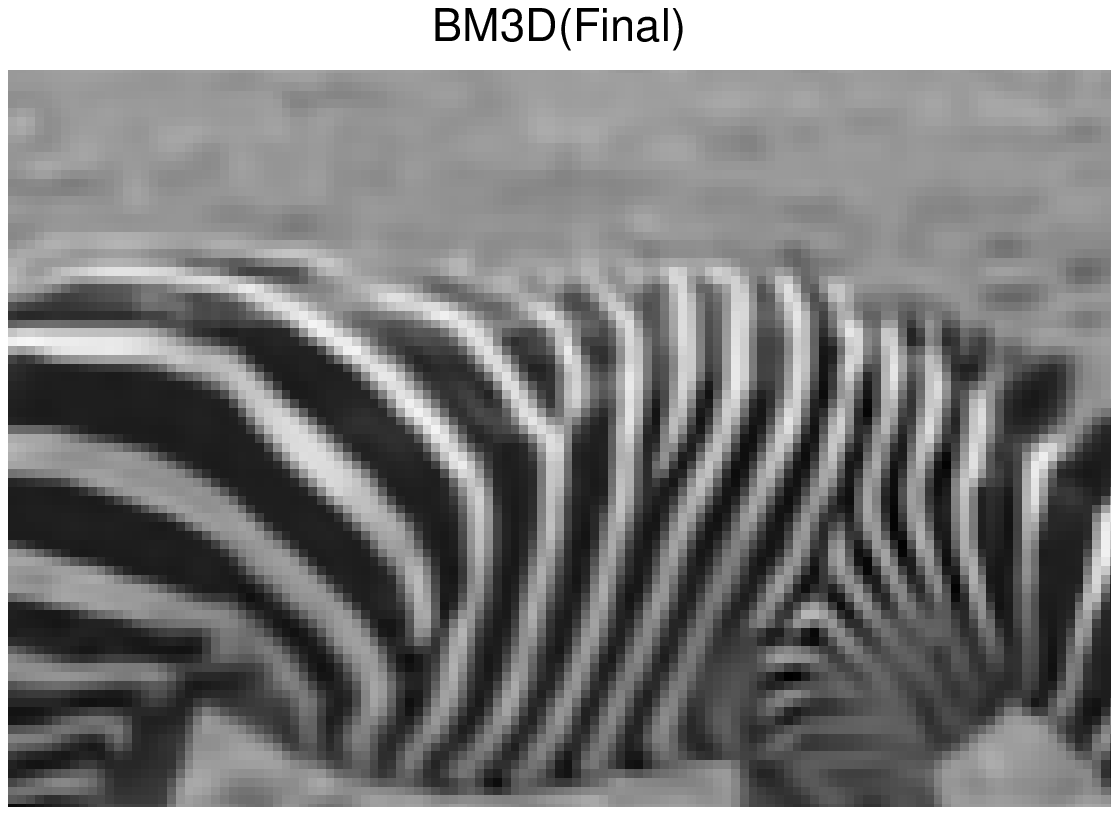} \\

  \caption{\small The zoom in visual detail comparisons between Proposed method with BM3D methods corresponding to Figure \ref{fig: BM3D}.}
\label{fig: zoomin BM3D}
\end{figure}

Finally, we compare the proposed algorithm with the nonlocal patch-based image deblurring IDD-BM3D algorithm \cite{Danielyan2012}, which is one of the state-of-the-art method in this field. To guarantee the performance of IDD-BM3D algorithm, the two-satge approach BM3DDEB \cite{Dabov2008} is used as an initial estimation. Two blur types are consider here. One is the blur type II with noise $\sigma=\sqrt{2}$, and the other is blur type IV with noise $\sigma=3.0, 4.0$, respectively. In Table 2, we list the PSNR and SSIM values of the recovery results of different algorithms. In Figure \ref{fig: BM3D}, we give the visual comparisons between our proposed algorithm and the IDD-BM3D algorithm. As can be seen from both the Table \ref{Table:BM3D} and visual restored results, the recovery quality of our proposed algorithm is approximately comparable with IDD-BM3D algorithm. Specifically, our proposed method outperform the IDD-BM3D algorithm in some cases in terms of better SSIM values, which is more consistent with human eye perception. To make the comparison more clear, in Figure \ref{fig: zoomin BM3D}, we also display the zoom in views corresponding to Figure \ref{fig: BM3D}. we can observe that our proposed method tends to generate more sharper image edges than the IDD-BM3D algorithm. The excellent edge preservation owes to the implement of support prior of the frame coefficients. In addition, as can be seen from the Table \ref{Table:BM3D}, the CPU time of our proposed method is less than the IDD-BM3D algorithm. We emphasize that the code of the IDD-BM3D algorithm is available at http://www.cs.tut.fi/~foi/GCF-BM3D/BM3D.zip. The computationally most intensive parts have been written in C++ due to the high computational complex, while all the parts of our algorithm are written in MATLAB language. Therefore, compared with the IDD-BM3D algorithm, our proposed method is computationally more efficient in this aspect.

\section{Conclusions and possible future work}
In this paper, we propose a new wavelet frame based image restoration model (\ref{eq:new truncatedl0l2}) which exploits the sparsity, nonlocal and support prior of frame coefficients simultaneously. In the proposed model, the penalization of the $l_2$ norm plays the role of preserving the coefficients that contain textures and finer details,  and the penalization of truncated $l_0$ ``norm" is used to recover the smoothness of homogeneous regions and sharp edges.  Better sharpening of the edges in the recovered results is observed compared with the corresponding $l_0$ ``norm" based alternatives.  The corresponding algorithm is  a multi-stage process consisting of solving a series of truncated $l_0$-$l_2$ minimization problem.
The proposed method can bring significant enhancements at the sharp edges of the restored images, since the large wavelet frame coefficients reflect the singularities of the underlying true solution and are not been shrunk. 
The key component of our proposed method is the reliable support detection, and therefore the next step along this line of research is to develop more effective support detection methods.

\textbf{Reference}
\bibliographystyle{model1c-num-names}
\bibliography{<your-bib-database>}

\begin{thebibliography}{l}

\bibitem{Bredies 2010} K. Bredies, K. Kunisch, Total generalized variation, SIAM J.Imaging.Sci. 3(3), (2010) 492-526.

\bibitem{Boyd2010} S. Boyd, N. Parikh, E. Chu, B. Peleato and J. Eckstein, Distributed optimization and statistical learning via the alternating direction method of multipliers, Foundations and Trends in Machine Learning, 3(1), (2010), 1-122.

\bibitem{Buades2005} A. Buades, B. Coll, J. M. Morel, A review of image denoising algorithms, with a new one. Multiscale Model. Simul. 4(2), (2005), 490-530.

\bibitem{Cai2012} J. Cai, B. Dong, S. Osher, et.al, Image restoration: Total variation, wavelet frame and beyond, J.Amer.Math.Soc. 25(4), (2012) 1033-1089.

\bibitem{Chambolle 1997} A. Chambolle, P.-L Lions, Image recovery via total variation minimization and related problems, Numer.Math. 76(2), (1997) 167-188.

\bibitem{Cai2009a} J. Cai, S. Osher, and  Z. Shen, Split Bregman methods and frame based image restoration, Multiscale Modeling and Simulation: A SIAM Interdiscplinary Journal, 8(2), (2009), 337-367.

\bibitem{Chan2003} R. Chan, T. Chan, L. Shen, Wavelet algorithms for high-resolution image reconstruction, SIAM Journal on Scientific Computing, 24(4), (2003), 1408-1432.

\bibitem{Cai2008} J. Cai, R. Chan, L. Shen, and Z. Shen, Convergence analysis of tight framelet approach for missing data recovery, Advances in Computeational Mathematics, (2008), 1-27.

\bibitem{Cai2010} J. Cai, R. Chan, and Z. Shen, Simutaneous cartoon and texture inpainting, Inverse Problems and Imaging, 4(3), (2010), 379-395.

\bibitem{Cai2014} J. Cai, H. Ji, Z. Shen, Data-driven tight frame construction and image denoising. Appl. Comput. Harmonic Anal. 37(1), (2014), 89-105.

\bibitem{Chen2015} D. Chen, Y. Zhou, Wavelet frame based image restoration via combined sparsity and nonlocal prior of coefficients. J.Sci.Computer. (2015).

\bibitem{Cai2009b} J. Cai, R. Chan, L. Shen, and Z. Shen, Convergence analysis of tight framelet approach for missing data recovery, Advances in Computeational Mathematics, 31 (1-3), (2009), 87-113.

\bibitem{Cand¨¨s2008} E. J. Cand¨¨s, M.B. Wakin, An introduction to compressive sampling. IEEE Signal Process.Mag. 25(2), (2008), 21-30.

\bibitem{Chartrand2009} R. Chartrand, Fast algorithms for nonconvex compressive sensing: MRI reconstruction from very few data. In: IEEE International Symposium on Biomedical Imaging (ISBI), (2009), 262-265.

\bibitem{Chartrand2008} R. Chartrand, W. Yin, Iteratively reweighted algorithms for compressive sensing. In: 33rd International Conference on Acoustics, Speech, and Signal Processing (ICASSP), (2008), 3869-3872.

\bibitem{Daubechies2007} I. Daubechies, G. Teschke, and L. vese, Iteratively solving linear inverse problems under general convex constraints, Inverse Problems and Imaging, 1(1), (2007), 29.

\bibitem{Dong2013} B. Dong, Y. Zhang, An efficient algorithm for $l_0$ minimization in wavelet frame based image restoration. J.Sci.Computer. 54(2-3), (2013) 350-368.

\bibitem{Dabov2008} K. Dabov, A. Foi, V. Katkovnik, K. Egiazarian, Image restoration by sparse 3D transform-domain collaborative filtering. International Society for Optics and Photonics, (2008), 2080-2095.

\bibitem{Danielyan2012} A. Danielyan, V. Katkovnik, K. Egiazarian, BM3D frames and variational image deblurring. IEEE Trans. Image Process. 21(4), (2012), 1715-1728.

\bibitem{Deledalle2009} C. A. Deledalle, L. Denis, F. Tupin, Iterative weighted maximum likelihood denoising with probabilistic patch-based weights. IEEE Trans. Image Process. 18(12), (2009), 2661-2672.

\bibitem{Dabov2007} K. Dabov, A. Foi, V. Katkovnik, et al, Image denoising by sparse 3-D transform-domain collaborative filtering. IEEE Trans. Image Process. 16(8), (2007), 2080-2095.

\bibitem{Dongweisheng2011} W. Dong, X. Li, D. Zhang, et al, Sparsity-based image denoising via dictionary learning and structural clustering. IEEE Conference on Computer Vision and Pattern Recognition (CVPR), (2011), 457-464.

\bibitem{Dongweisheng2013a} W. Dong, L. Zhang, G. Shi, et al, Nonlocally centralized sparse representation for image restoration. IEEE Trans. Image Process. 22(4), (2013), 1620-1630.

\bibitem{Dongweisheng2013b} W. Dong, G. Shi, X. Li, Nonlocal image restoration with bilateral variance estimation: a low-rank approach. IEEE Trans. Image Process. 22(2), (2013), 700-711.

\bibitem{Dong2010} B. Dong and Z. Shen, MRA-Based Wavelet Frames and Apllications, IAS Lecture Notes Series, Summer Program on, The Mathematics of Image Processing, Park City Mathematics Institute, (2010).

\bibitem{Daubechies2003} I. Daubechies, B. Han, A. Ron, and Z. Shen, Framelets: Mra-based constructions of wavelet frames, Applied and Computational Harmonic Analysis, 14, (2003), 1-46.

\bibitem{Elad2005} M. Elad, J. Starck, P. Querre, and D. Donoho, Simultaneous cartoon and texture image inpaiting using morphological component analysis (MCA), Applied and Computational Harmonic Analysis, 19(3), (2005), 340-358.

\bibitem{Esser2009} E. Esser, Applications of Lagrangian-based alternating direction methods and connections to split Bragman, CAM Report, 9(31), (2009).

\bibitem{Fadili2005} M. Fadili and J. Starck, Sparse representations and bayesian image inpainting, Proc. SPARS, 5, (2005).

\bibitem{Fadili2009} M. Fadili, J. Starck, and F. Murtagh, Inpainting and zooming using sparse representations, The Computer Journal, 52(1), (2009), 64.

\bibitem{Figueiredo2003} M. Figuriredo and R. Nowak, An EM aogorithm for wavelet-based image restoration,  IEEE Transactiona on Image Processing, 12(8), (2003), 906-916.

\bibitem{Figueiredo2005} M. Figuriredo and R. Nowark, A bound optimization approach to wavelet-based image deconvolution, in  ICIP 2005. IEEE International Conference on, 2, (2005), pp. $\amalg$ -782.

\bibitem{Fan2014} Y. Fan, Y. Wang, T. Huang,  Enhanced joint sparsity via iterative support detection. Eprint Arxiv, 2014.


\bibitem{Goldtein2009} T. Goldtein and S. Osher, The Split Bregman algorithm for L1 regularized problems, SIAM Journal on Imaging Sciences. 2, (2009), 323-343.

\bibitem{He2014} L. He, Y. Wang, Iterative support detection based split bregman method for wavelet frame based image inpainting, IEEE Trans. Image Process. 23(12), (2014), 5470-5485.


\bibitem{Kindermann2005} S. Kindermann, S. Osher, P. W. Jones, Deblurring and denoising of images by nonlocal functionals. Multiscale Model. Simul. 4(4), (2005), 1091-1115.


\bibitem{Mairal2009} J. Mairal, F. Bach, J. Ponce, et al, Non-local sparse models for image restoration. IEEE International Conference on Computer Vision (ICCV), (2009), 2272-2279.

\bibitem{Quan2014} Y. Quan, H. Ji, Z. Shen, Data-Driven Multi-scale Non-local Wavelet Frame Construction and Image Recovery. J. Sci. Comput. 63(2), (2014), 307-329.

\bibitem{Ron1997} A. Ron and Z. Shen, Affine Systerms in $L_2(\mathbb{R}^d)$: The Analysis of the Analysis Operator, Journal of Functional Analysis, 148(2), (1997), 408-447.

\bibitem{ROF 1992} L. I. Rudin, S. Osher, and E. Fatemi, Nonlinear total variation based noise removal algorithms, phys. D, Nonliear Phenomena. 60(1-4) (1992), 259-268.


\bibitem{Rockafellar1976} R. T. Rockafellar, Augmented Lagrangians and applications of the proximal point algorithm in convex programming. Math. Oper. Res. 1(2),  (1976), 97-116.

\bibitem{Shen2010} Z. Shen, Wavelet frames and image restorations, in Proceedings of the International Congress of Mathematicians, 4 (2010), 2834-2863.

\bibitem{Starck2005} J. Starck, M. Elad, and D. Donoho, Image decomposition via the combination of sparse representations and a variational approach,
IEEE Trans. Image Process. 14(10), (2005), 1570-1582.


\bibitem{Shen2011} Z. Shen, K. C. Toh, S. Yun, Anaccelerated proximal gradient algorithm for frame-based image restoration via the balanced approach. SIAM J. Imaging Sci. 4(2), (2011), 573-596.


\bibitem{Wang 2008} Y. Wang, J. Yang, W. Yin, and Y. Zhang, A new alternating direction minimization algorithm for total variation image reconstruction, SIAM J.Imaging.Sci. 1(3), (2008) 248-272.

\bibitem{Wang2010} Y. Wang,  W. Yin, Sparse signal reconstruction via iteration support detection, SIAM J.Imaging.Sci. 3(3), (2010), 462-491.

\bibitem{Wang2004} Z. Wang, A. C. Bovik, H. R. Sheikh, et al, Image quality assessment: from error visibility to structural similarity. IEEE Trans. Image Process. 13(4), (2004), 600-612.

\bibitem{Zhangxiaoqun2010} X. Zhang, M. Burger, X. Bression, et.al, Bregmanized nonlocal regularization for deconvolution and sparse reconstruction, SIAM J.Imaging.Sci. 3(3), (2010) 253-276.

\bibitem{Zhangxiaoqun2011} X. Zhang, M. Burger, S. Osher, A unified primal-dual algorithm framework based on Bregman iteration. J. Sci. Comput. 46(1),  (2011), 20-46.

\bibitem{ZhangY2013} Y. Zhang, B. Dong, Z. Lu, $l_0$ minimization for wavelet frame based image restoration. Math.Comput. 82(282), (2013) 995-1015.






\end{thebibliography}


\end{document}